\documentclass{sfuthesis}

\title{A Review of Deep Learning-Powered Mesh Reconstruction Methods}
\author{Zhiqin Chen}
\previousdegrees{Simon Fraser University}
\copyrightyear{2023}
\semester{Spring 2023}


\committee{
    \member{Hao (Richard) Zhang}{Supervisor \\ Professor, Computing Science}
    \member{Yasutaka Furukawa}{Committee Member \\ Associate Professor, Computing Science}
    \member{Andrea Tagliasacchi}{Examiner \\ Associate Professor, Computing Science}
}

\usepackage{amsmath}
\usepackage{amssymb}
\usepackage{amsthm} 
\usepackage{graphicx}
\usepackage[pdfborder={0 0 0}]{hyperref}

\begin{document}

\frontmatter
\maketitle{}

\begin{abstract}

With the recent advances in hardware and rendering techniques, 3D models have emerged everywhere in our life. Yet creating 3D shapes is arduous and requires significant professional knowledge. Meanwhile, Deep learning has enabled high-quality 3D shape reconstruction from various sources, making it a viable approach to acquiring 3D shapes with minimal effort. Importantly, to be used in common 3D applications, the reconstructed shapes need to be represented as polygonal meshes, which is a challenge for neural networks due to the irregularity of mesh tessellations.

In this survey, we provide a comprehensive review of mesh reconstruction methods that are powered by machine learning. We first describe various representations for 3D shapes in the deep learning context. Then we review the development of 3D mesh reconstruction methods from voxels, point clouds, single images, and multi-view images. Finally, we identify several challenges in this field and propose potential future directions.

\end{abstract}

\addtoToC{Table of Contents}%
\tableofcontents%
\clearpage




\mainmatter%

\chapter{Introduction}
\label{sec:intro}

With the recent advances in hardware and rendering technologies, 3D models have emerged everywhere in our life: movies, video games, AR/VR, autonomous driving, and even on websites\cite{sketchfab}.
In contrast to images and videos where people can easily create them with cameras or smartphones, modeling 3D content requires professional knowledge.
Much like programming, only ones who have gone through tedious training and practices are able to create 3D models and scenes in 3D modeling software.

\begin{figure}[b!]
\begin{center}
\includegraphics[width=1.0\linewidth]{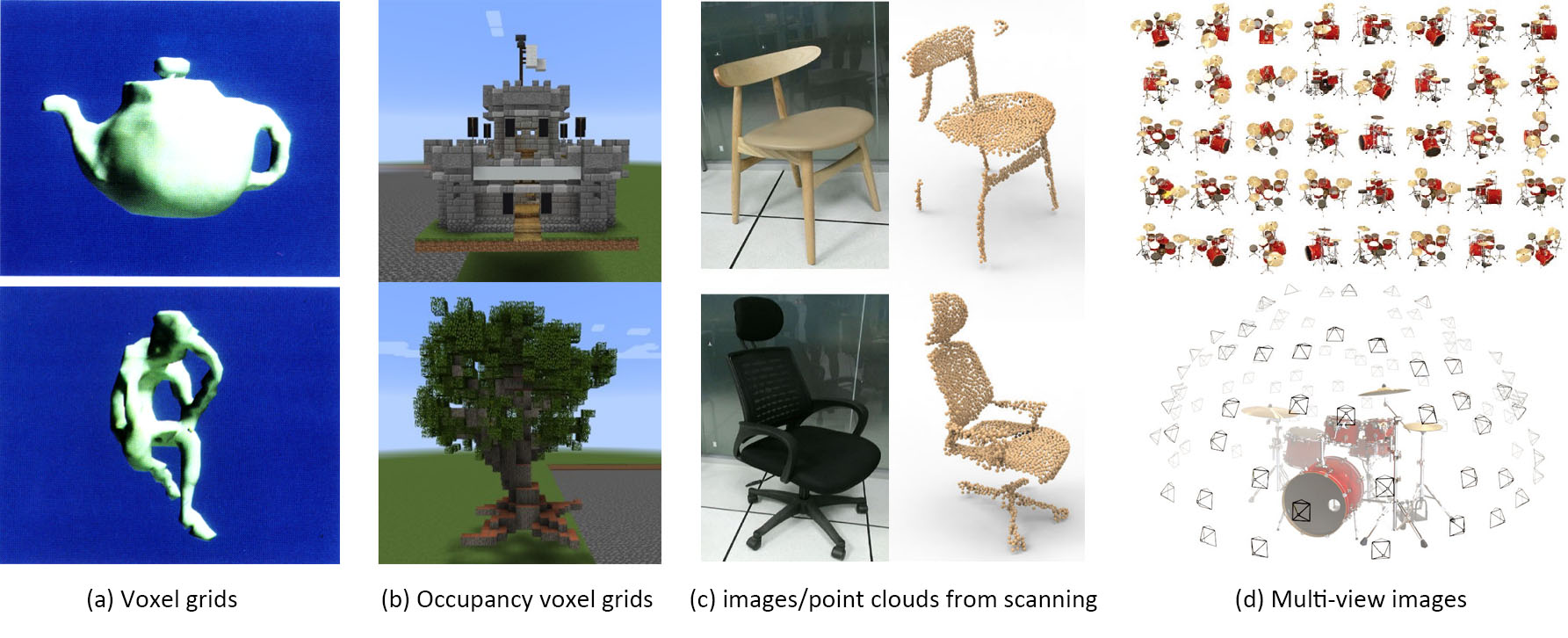}
\end{center}
\caption{
Examples of easy-to-acquire data that can be used to reconstruct 3D shapes. (a) is in fact represented by 3D grids of float point numbers; they are hard to visualize so only an iso-surface is shown. (b) contains 3D grids of binary occupancies; each occupied voxel is represented as a textured cube. (c) shows single-view RGB images of the objects and their corresponding partial point clouds obtained from a depth sensor. (d) has multi-view images on top and the camera poses visualized at the bottom; note that point clouds can also be aggregated from multiple views to provide a more complete coverage of the object surface. These figures are taken from their corresponding publications (a)\cite{galyean1991sculpting}, (b)\cite{sudhakaran2021growing}, (c)\cite{yin2018p2p}, (d)\cite{ mildenhall2021nerf}.
} 
\label{fig:modeling_inputs}
\end{figure}

Nonetheless, it is possible to create 3D models in a simpler fashion. Ideally, the users can provide some easy-to-acquire data as input, and rely on a sophisticated 3D reconstruction algorithm to obtain the 3D model. Some example inputs are shown in Figure~\ref{fig:modeling_inputs}.
First, instead of modeling polygons as in most 3D modeling software, one can model 3D shapes by using a volumetric brush to add or remove volumes in a 3D grid to sculpt the shape, akin to painting with a brush on a 2D pixel grid in most 2D painting software, see Figure~\ref{fig:modeling_inputs} (a).
Similarly, Minecraft (video game) players can create coarse 3D models with textured voxels, by moving their avatars in the voxelized game world and putting together the shapes using voxels as bricks, see Figure~\ref{fig:modeling_inputs} (b).
On the other hand, single or multi-view images taken from cameras, and point clouds obtained from scanners in modern phones or other depth sensors can also be used for 3D reconstruction, see Figure~\ref{fig:modeling_inputs} (c,d).

Therefore, having robust tools to perform 3D shape reconstruction from these inputs are of great importance. They can be highly useful in people's daily lives and benefit the entertainment industry.
Great amount of research and effort have already been invested in the related fields. Many classic reconstruction methods such as Marching Cubes\cite{lorensen1987marching} for iso-surface reconstruction, Screened Poisson\cite{kazhdan2013screened} for point cloud reconstruction, and COLMAP\cite{schoenberger2016sfm,schoenberger2016mvs} for multi-view reconstruction are very robust algorithms and have been used even today.
However, it is the deep learning revolution that has brought tools for people to achieve many tasks that can never be achieved before. Deep learning-powered methods have not only relaxed the input constraints, allowing us to reconstruct shapes from sparse or noisy data, but also improved the reconstruction quality to a whole new level.

Still, many challenges remain, and most methods are far from being able to be applied in real products. One of the big challenges is that almost all software and hardware can only support triangle meshes, which should be the ideal output of these reconstruction methods. However, the non-uniformity and irregularity of triangle tessellations do not naturally support conventional convolution operations, so new network architectures and new representations have been devised to represent 3D shapes in compatible ways with respect to neural networks.

This survey provides a comprehensive review of those mesh reconstruction methods powered by machine learning.
In Section~\ref{sec:representation}, we describe various representations for 3D shapes in the deep learning context.
Then in Section~\ref{sec:voxel}, \ref{sec:pointcloud}, \ref{sec:singleimage}, and \ref{sec:multiview}, we review 3D reconstruction methods that reconstruct surfaces from voxels, point clouds, single images, and multi-view images, respectively.
Finally, in Section~\ref{sec:conclusion}, we identify several challenges in this field and propose potential future directions.

\chapter{Representations}
\label{sec:representation}

The foundation for any algorithm is the data representation. For images, pixel is the representation used by both academia and industry. Unfortunately, there is no such unified representation for 3D models. In fact, researchers have proposed a wide range of representations for 3D generative tasks.

In this survey, we focus on representations that are essentially triangle meshes. We also consider representations that can be easily converted into triangle meshes, such as CSG (Constructive Solid Geometry) trees and parametric surfaces. We will also discuss implicit representations such as voxel grids and neural implicit, since they are the most popular representations, although they can face issues when being converted into triangle meshes, such as creating excessive numbers of vertices and triangles.

\begin{table}[b!]
\begin{center}
\resizebox{1.0\linewidth}{!}{
\begin{tabular}{l|c|l|c}
\hline
Representation & General representation & Section & Example \\
\hline\hline
Deform one template & Mesh & \ref{sec:representation_1_deform_template} & {\small Human from single image} \cite{pavlakos2018learning} \\
Retrieve and deform template & Mesh & \ref{sec:representation_2_retrieve_and_deform_template} & ShapeFlow \cite{jiang2020shapeflow} \\
Deform one primitive & Mesh & \ref{sec:representation_3_deform_one_primitive} & Pixel2Mesh \cite{wang2018pixel2mesh} \\
Deform multiple primitives & Mesh & \ref{sec:representation_4_deform_multi_primitive} & AtlasNet \cite{groueix2018papier} \\
Set of primitives & Mesh & \ref{sec:representation_5_set_of_primitives} & BSP-Net \cite{chen2020bsp} \\
Primitive detection & Parametric surface & \ref{sec:representation_6_primitive_detection} & ParSeNet \cite{sharma2020parsenet} \\
\hline
Grid mesh & Mesh & \ref{sec:representation_7_grid_mesh} & Neural Marching Cubes \cite{chen2021nmc} \\
Grid polygon soup & Mesh & \ref{sec:representation_8_grid_polygon_soup} & Adaptive O-CNN \cite{wang2018adaptive} \\
\hline
Grid voxels & Implicit & \ref{sec:representation_9_grid_voxel_SDF} & 3D-R2N2 \cite{choy20163d} \\
Neural implicit & Implicit & \ref{sec:representation_10_neural_implicit} & IM-Net\cite{chen2019learning}, DeepSDF \cite{park2019deepsdf} \\
\hline
Primitive CSG & CSG tree & \ref{sec:representation_11_CSG} & CAPRI-Net \cite{yu2022capri} \\
Sketch and extrude & CSG tree & \ref{sec:representation_12_sketch_extrude} & ExtrudeNet \cite{ren2022extrudenet} \\
\hline
Connect given vertices & Mesh & \ref{sec:representation_13_connect_V} & PointTriNet \cite{sharp2020pointtrinet} \\
Generate and connect vertices & Mesh & \ref{sec:representation_14_generate_VT} & PolyGen \cite{nash2020polygen} \\
Sequence of edits & Mesh & \ref{sec:representation_15_sequence_of_edits} & {\small Modeling 3D Shapes by RL} \cite{lin2020modeling} \\
\hline
\end{tabular}
}
\end{center}
\caption{
The representations covered in this section and where to find them. The representations are arbitrarily named and may not be consistent with other papers. A representative work is included for each representation to give readers a rough idea of the representation.
}
\label{table:representations}
\end{table}

Table~\ref{table:representations} lists a summary of representations used in deep-learning mesh reconstruction methods. Note that although methods outputting point clouds are not considered, some of them use points to represent an implicit field, thus a 3D shape can be easily extracted. For example, ``Shape As Points''\cite{peng2021shape} proposes a point-to-mesh layer using a differentiable formulation of Poisson Surface Reconstruction \cite{kazhdan2006poisson,kazhdan2013screened} to represent a shape as an implicit field from a set of points with normals. On the other hand, with works such as ``Analytic Marching''\cite{lei2020analytic} that can extract the exact polygonal mesh from neural implicit representations, the gap between neural implicit and explicit meshes may diminish in the future.

In the following, we will introduce each representation and its related works in detail.

\section{Deform one template}
\label{sec:representation_1_deform_template}

\begin{figure}[b!]
\begin{center}
\includegraphics[width=0.8\linewidth]{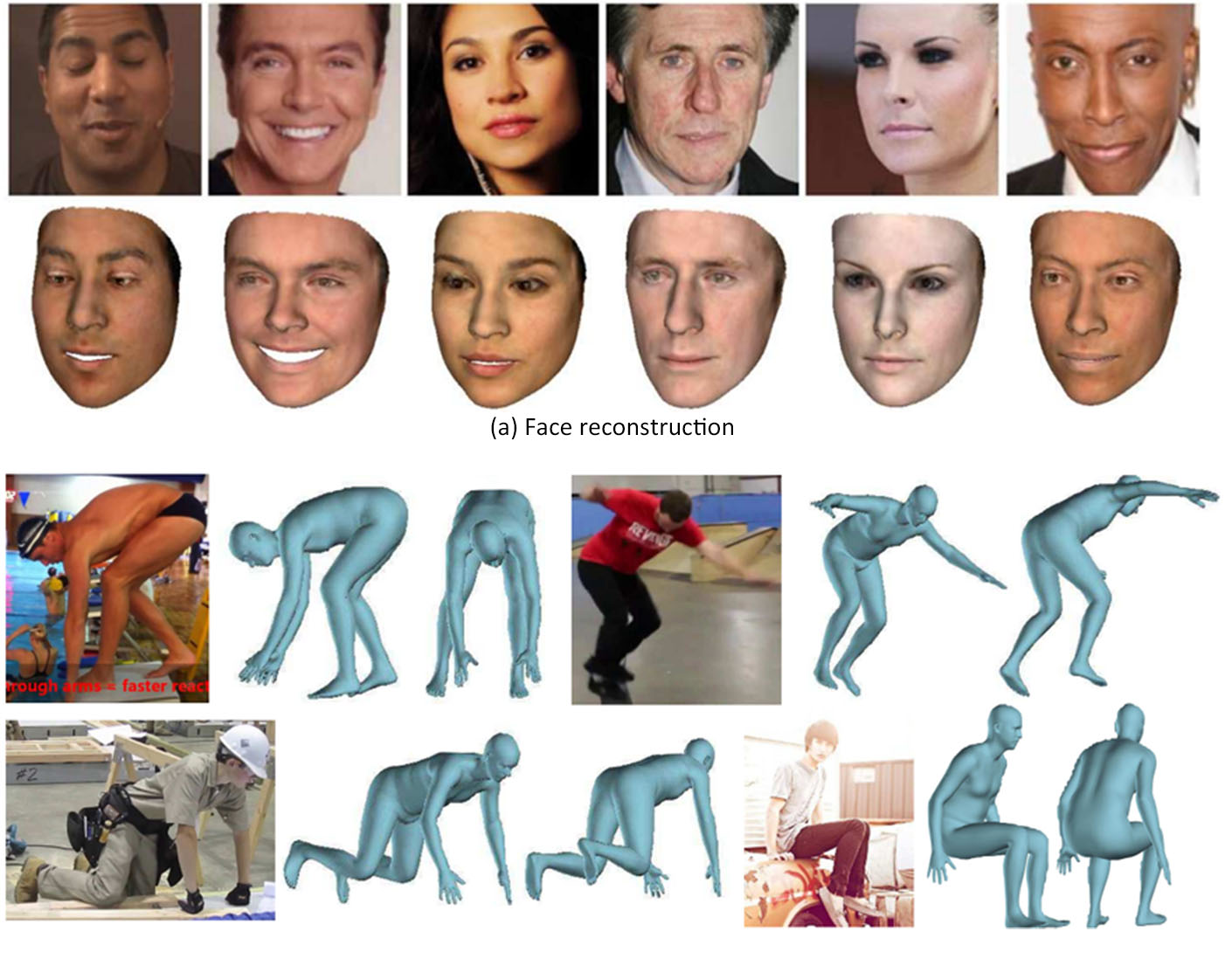}
\end{center}
\vspace{-8mm}
\caption{
Example inputs and results of face and human shape reconstruction methods. These figures are taken from their corresponding publications (a)\cite{deng2019accurate}, (b)\cite{pavlakos2018learning}.
} 
\label{fig:1_deform_template}
\end{figure}

Deforming a single template mesh to create shapes of different poses has been widely used for face \cite{deng2019accurate} and human body \cite{kolotouros2019convolutional,bogo2016keep,pavlakos2018learning} shape reconstruction, see Figure~\ref{fig:1_deform_template}. However, due to the availability of a single template mesh, such representation is not suitable for reconstructing general 3D objects, thus we do not provide more discussion here. Reconstructing face and human body meshes from images and videos are active research domains and we encourage the readers to search for related surveys if interested.

\clearpage
\section{Retrieve and deform template}
\label{sec:representation_2_retrieve_and_deform_template}

\begin{figure}[b!]
\begin{center}
\includegraphics[width=1.0\linewidth]{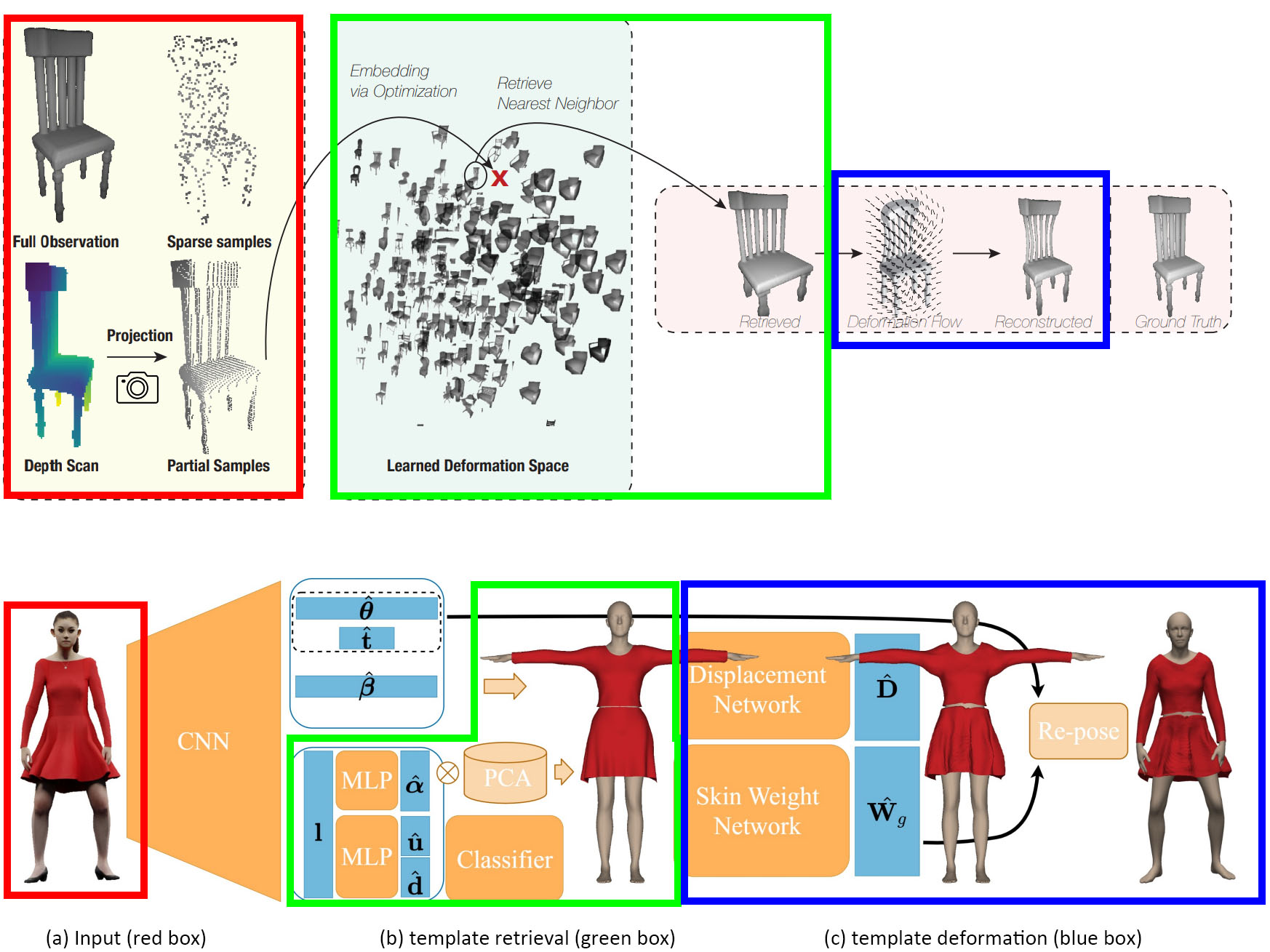}
\end{center}
\caption{
Overview of two retrieve-and-deform shape reconstruction methods: ShapeFlow \cite{jiang2020shapeflow} (top) and BCNet \cite{jiang2020bcnet} (bottom). These figures are taken from their corresponding publications.
} 
\label{fig:2_retrieve_and_deform_template}
\end{figure}

Extending the idea of deforming a single shape, one can first retrieve a most suitable template for the target object, and then deform the template, to achieve optimal performance. Examples are shown in Figure~\ref{fig:2_retrieve_and_deform_template}. The methods usually process the inputs with neural networks to either classify which template is most suitable, such as BCNet \cite{jiang2020bcnet} and Multi-Garment Net (MGN) \cite{bhatnagar2019multi}, as shown in Figure~\ref{fig:2_retrieve_and_deform_template} (b) bottom; or embed the input into a deformation-aware latent space to retrieve the nearest-neighbor template, such as 3D Deformation Network (3DN) \cite{wang20193dn}, Deformation-Aware 3D Model Embedding and Retrieval \cite{uy2020deformation}, and ShapeFlow \cite{jiang2020shapeflow}, as shown in Figure~\ref{fig:2_retrieve_and_deform_template} (b) top. Then, a deformation network is applied to deform the template into the target shape, as shown in Figure~\ref{fig:2_retrieve_and_deform_template} (c).

\clearpage

\section{Deform one primitive}
\label{sec:representation_3_deform_one_primitive}

``Deform one template'' and ``retrieve and deform template'' assume high-quality shape templates are given, and they are most used for highly specialized reconstruction tasks. For general shape reconstruction tasks, especially when reconstructing 3D shapes with only 2D image supervision, a primitive shape could be used as the initial shape, and it can be deformed to approximate the target shape.

\begin{figure}[b!]
\begin{center}
\includegraphics[width=1.0\linewidth]{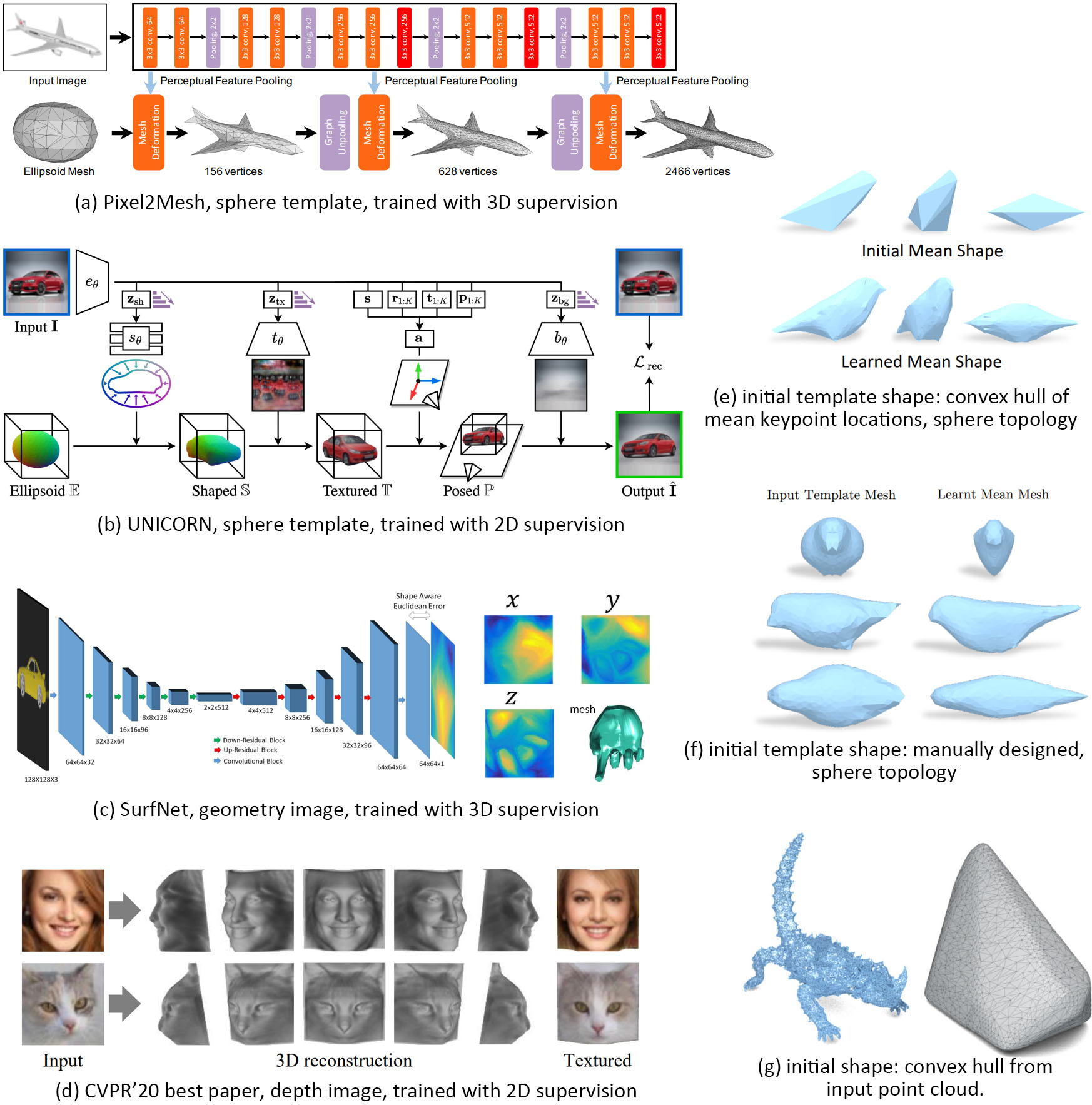}
\end{center}
\caption{
Overview of some methods that deform a simple primitive shape. (a) shows a method that deforms and subdivides a sphere to reconstruct a 3D shape from a single image input, trained with ground truth 3D supervision on the output mesh. (b) is similar to (a) but it only has 2D supervision, thus it needs to predict geometry, texture, camera pose, and background image, and render them with a differentiable renderer, and finally compare the rendered image with the ground truth image for supervision. (c) predicts a geometry image from the input image, therefore a mesh can be recovered from the geometry image. (d) predicts depth map and texture map from an input image, therefore geometry can be recovered from the depth map. (e) and (f) show two different initial templates that can be used in shape reconstruction tasks with only 2D supervision similar to (b). Note that in (e)(f) the template meshes are optimizable, and the optimized (learned) templates after training are also includes in the figures. (g) shows the initial mesh for mesh reconstruction from point clouds. These figures are taken from their corresponding publications (a)\cite{wang2018pixel2mesh}, (b)\cite{monnier2022unicorn}, (c)\cite{sinha2017surfnet}, (d)\cite{wu2020unsupervised}, (e)\cite{kanazawa2018learning}, (f)\cite{goel2020shape}, (g)\cite{hanocka2020point2mesh}.
} 
\label{fig:3_deform_one_primitive}
\end{figure}

The most commonly used primitive is a simple sphere, as shown in Figure~\ref{fig:3_deform_one_primitive} (a)(b). Representative works that are trained with ground truth 3D supervision include AtlasNet (sphere version) \cite{groueix2018papier}, Pixel2Mesh \cite{wang2018pixel2mesh}, Pixel2Mesh++ \cite{wen2019pixel2mesh++}, Neural mesh flow \cite{gupta2020neural}, and \cite{pan2019deep}. It may not be the most popular representation for models trained with 3D supervision, but it indeed has been the dominant representation in models trained with only 2D image supervision, such as N3MR (Neural 3d Mesh Renderer) \cite{kato2018neural}, Soft Rasterizer \cite{liu2019soft}, DIB-R \cite{chen2019learning}, DIB-R++ \cite{chen2021dib}, Image GANs meet Differentiable Rendering \cite{zhang2020image}, UNICORN \cite{monnier2022unicorn}, NeRS \cite{zhang2021ners}, and \cite{henderson2020leveraging,pavllo2020convolutional,li2020self}. There are multiple ways to perform deformation on the sphere: it can be an MLP (MultiLayer perceptron) that takes a 3D point on the sphere surface and outputs its deformed coordinates \cite{groueix2018papier,pan2019deep,monnier2022unicorn,zhang2021ners}; or using graph convolutional networks on the sphere meshes to predict vertex positions \cite{wang2018pixel2mesh,wen2019pixel2mesh++}; or using a CNN decoder to predict a displacement map on the sphere \cite{pavllo2020convolutional}; the majority of the methods simply use an MLP to predict the offsets of all vertices in the template mesh, outputting a $V \times 3$ vector where $V$ is the number of vertices in the template.

Some works on 3D reconstruction with 2D supervision adopt a better initialization than a simple sphere. CMR \cite{kanazawa2018learning} initializes the template shape as the convex hull of the mean keypoint locations obtained after running SFM (structure from motion) on the annotated keypoints, as shown in Figure~\ref{fig:3_deform_one_primitive} (e). \cite{kar2015category} initializes the template shape using the visual hull of the annotated object masks on the training images. \cite{goel2020shape} initializes the template shape using a very simple but manually designed mesh, as shown in Figure~\ref{fig:3_deform_one_primitive} (f). Note that for these works, the template meshes are optimizable during training; the optimized (learned) templates after training are also includes in Figure~\ref{fig:3_deform_one_primitive} (e)(f). 

One can also consider deforming a uniform 2D grid as deforming one primitive, with the primitive being a square. Therefore, many works that predict depth images from a single image can be put into this representation category, e.g., ``Unsupervised learning of probably symmetric deformable 3d objects from images in the wild'' \cite{wu2020unsupervised}, as shown in Figure~\ref{fig:3_deform_one_primitive} (d). Similarly, there are models that predict geometry images \cite{gu2002geometry}, such as SurfNet \cite{sinha2017surfnet}, as shown in Figure~\ref{fig:3_deform_one_primitive} (c).

There are also works that fit a primitive into a point cloud to reconstruct a mesh from the point cloud, such as Point2Mesh \cite{hanocka2020point2mesh}, where the initial mesh is the convex hull of the point cloud, as shown in Figure~\ref{fig:3_deform_one_primitive} (g). However, a single sphere-topology primitive can only fit a single sphere-topology object, thus limiting the representation ability of this representation.

\clearpage

\section{Deform multiple primitives}
\label{sec:representation_4_deform_multi_primitive}

Deforming a single primitive shape limits the topology and the representation ability of the reconstructed shapes, therefore a nature solution is to fit multiple primitive shapes. However, this representation is mostly used when direct 3D supervision is available; it might be too complex and uncontrollable for reconstructing 3D shapes from 2D image supervision.

\begin{figure}[b!]
\begin{center}
\includegraphics[width=1.0\linewidth]{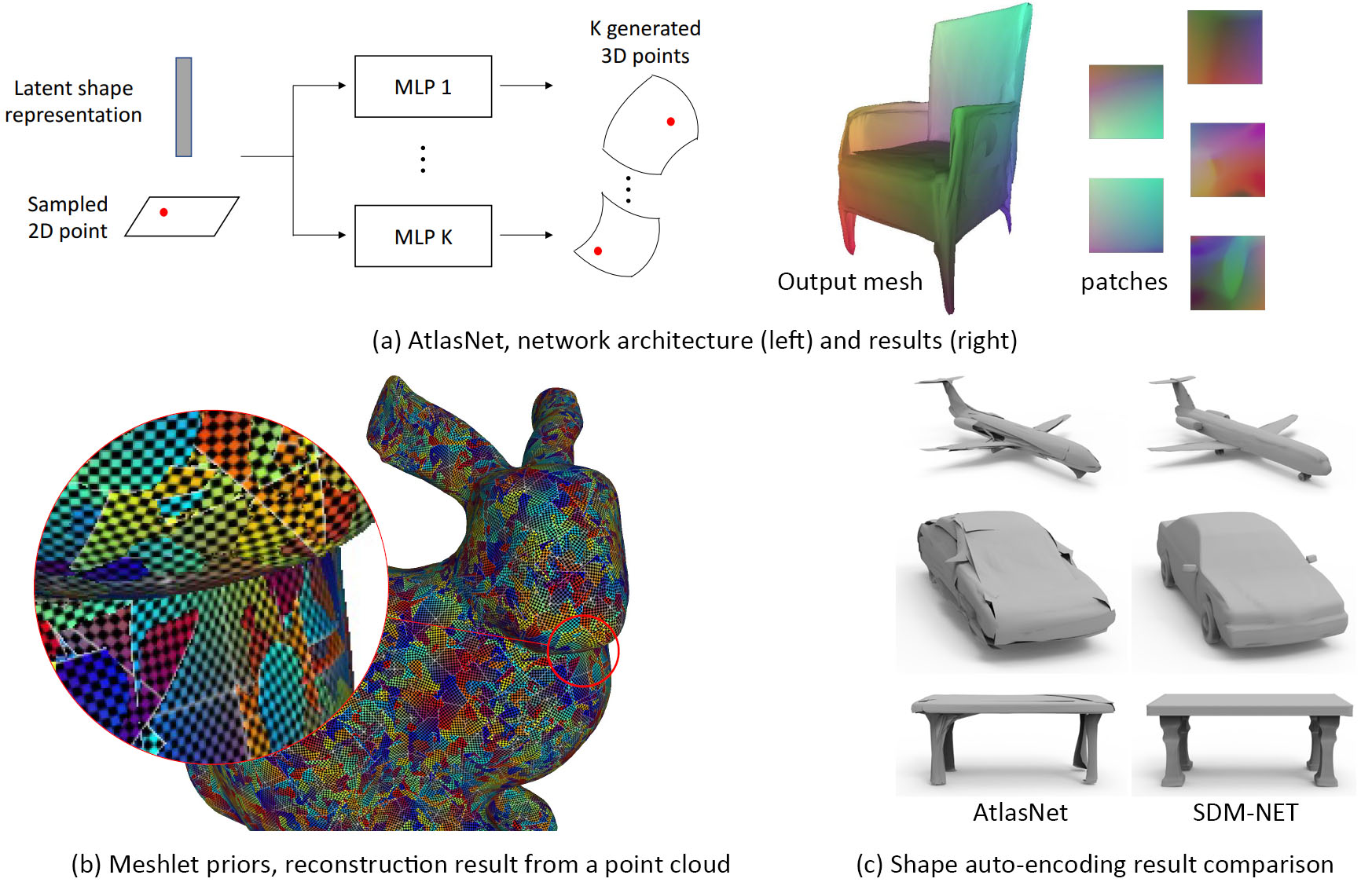}
\end{center}
\caption{
Overview of some methods that deform multiple primitive shapes. These figures are taken from their corresponding publications (a)\cite{groueix2018papier}, (b)\cite{badki2020meshlet}, (c)\cite{gao2019sdm}.
} 
\label{fig:4_deform_multi_primitive}
\end{figure}

Representative works include AtlasNet (patch version) \cite{groueix2018papier}, where an output shape is represented as a collection of square meshes, each deformed via an independent MLP, as shown in Figure~\ref{fig:4_deform_multi_primitive} (a).
Photometric Mesh Optimization \cite{lin2019photometric} adopts the 25-patch version of AtlasNet to reconstruct an object mesh from multi-view images.
Follow-up works fit square meshes into local patches of a shape to perform mesh reconstruction from point clouds, such as Deep geometric prior \cite{williams2019deep} and Meshlet priors \cite{badki2020meshlet}, as shown in Figure~\ref{fig:4_deform_multi_primitive} (b).

The square patches in AtlasNet lead to many issues, including overlapping patches, self-intersections, and conspicuous artifacts on the surface. SDM-NET \cite{gao2019sdm} has output shapes made of deformed unit cube meshes; it addresses the above issues via strong part-wise 3D supervision. A comparison between AtlasNet and SDM-NET can be found in Figure~\ref{fig:4_deform_multi_primitive} (c).
Another work \cite{smirnov2020learning} applies a deformable parametric template composed of Coons patches \cite{coons1967surfaces} to generate manifold and piecewise-smooth shapes made of parametric surfaces.

\clearpage

\section{Set of primitives }
\label{sec:representation_5_set_of_primitives}

This representation is very similar to the representation introduced in the previous section (``deform multiple primitives''). They are both shapes made of primitive meshes. The key difference is that in ``deform multiple primitives'', the primitive meshes have been deformed via a neural network. In ``set of primitives'', each primitive is just a primitive defined by a set of parameters, therefore there is no deformation network.

\begin{figure}[b!]
\begin{center}
\includegraphics[width=1.0\linewidth]{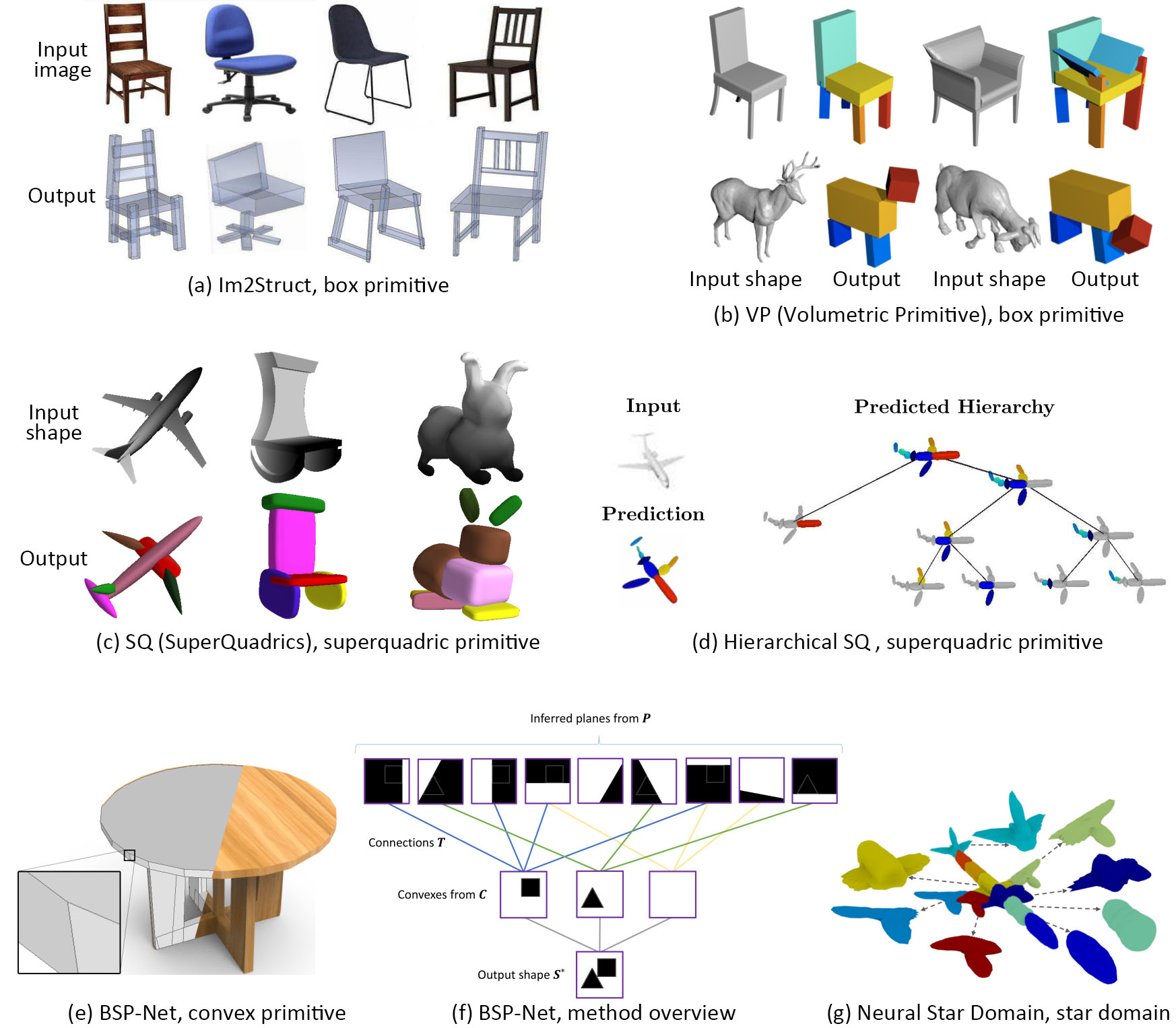}
\end{center}
\caption{
Overview of some methods that output shapes made of primitive meshes. These figures are taken from their corresponding publications (a)\cite{niu2018im2struct}, (b)\cite{tulsiani2017learning}, (c)\cite{paschalidou2019superquadrics}, (d)\cite{paschalidou2020learning}, (e)\cite{chen2020bsp}, (f)\cite{chen2020bsp}, (g)\cite{kawana2020neural}.
} 
\label{fig:5_set_of_primitives}
\end{figure}

A commonly used primitive type is bounding boxes or boxes that approximate the shape, such as those in Im2struct \cite{niu2018im2struct} and VP \cite{tulsiani2017learning}, as shown in Figure~\ref{fig:5_set_of_primitives} (a)(b). Boxes are easy to define: the network only needs to predict the size, the translation, the rotation, and the existence probability of each primitive box. An explicit mesh can also be easily obtained by a union of the box meshes.

Follow-up works such as SQ \cite{paschalidou2019superquadrics} and Hierarchical SQ \cite{paschalidou2020learning} use superquadrics instead of boxes for the primitives. Superquadrics \cite{barr1981superquadrics} are a parametric family of surfaces that can be used to describe cubes, cylinders, spheres, octahedra, ellipsoids, and other simple primitives, with just two shape parameters. Therefore, using superquadrics can improve the representation ability to better approximate the target shape, as shown in Figure~\ref{fig:5_set_of_primitives} (c)(d). To generate superquadrics, the network needs to predict the shape parameters, the size, the translation, the rotation, and the existence probability of each superquadric primitive. Meshes also can be easily obtained from superquadrics.

Convex primitives proposed in BSP-Net \cite{chen2020bsp} and CvxNet \cite{deng2020cvxnet} are more compact than superquadrics, and probably also more flexible as their primitive can model any convex shape. One example output of BSP-Net is shown in Figure~\ref{fig:5_set_of_primitives} (e). Unlike boxes or superquadrics, convex primitives cannot be described by a few parameters. The methods require intricate neural network design, as shown in Figure~\ref{fig:5_set_of_primitives} (f), where the network first predicts a set of planes or half-spaces, and then the network selects those half-spaces with learned binary weights and intersects them to form convex primitives, and finally the convex primitives are united to form the output shape. Polygonal meshes can be directly extracted from the tree structure.

To make the primitives more flexible, Neural Star Domain \cite{kawana2020neural} represents each primitive as a star domain, as shown in Figure~\ref{fig:5_set_of_primitives} (g). One can consider a star domain as a ``height'' function defined on the surface of a unit sphere (like a height map of earth's surface), therefore a star domain can be approximately represented as the coefficients of spherical harmonics. So to generate those primitives, the network needs to predict a few spherical harmonics coefficients for each primitive, plus the translation vector. Meshes can be extracted by deforming a unit-sphere template mesh with respect to the ``height'' function.

\clearpage

\section{Primitive detection}
\label{sec:representation_6_primitive_detection}

In previous primitive-based representations, the output shapes are usually generated by a shape decoder from a global shape latent code, therefore they are leaning towards generative models, and the learned shape latent space can be applied in different tasks such as auto-encoding pre-training.

\begin{figure}[b!]
\begin{center}
\includegraphics[width=1.0\linewidth]{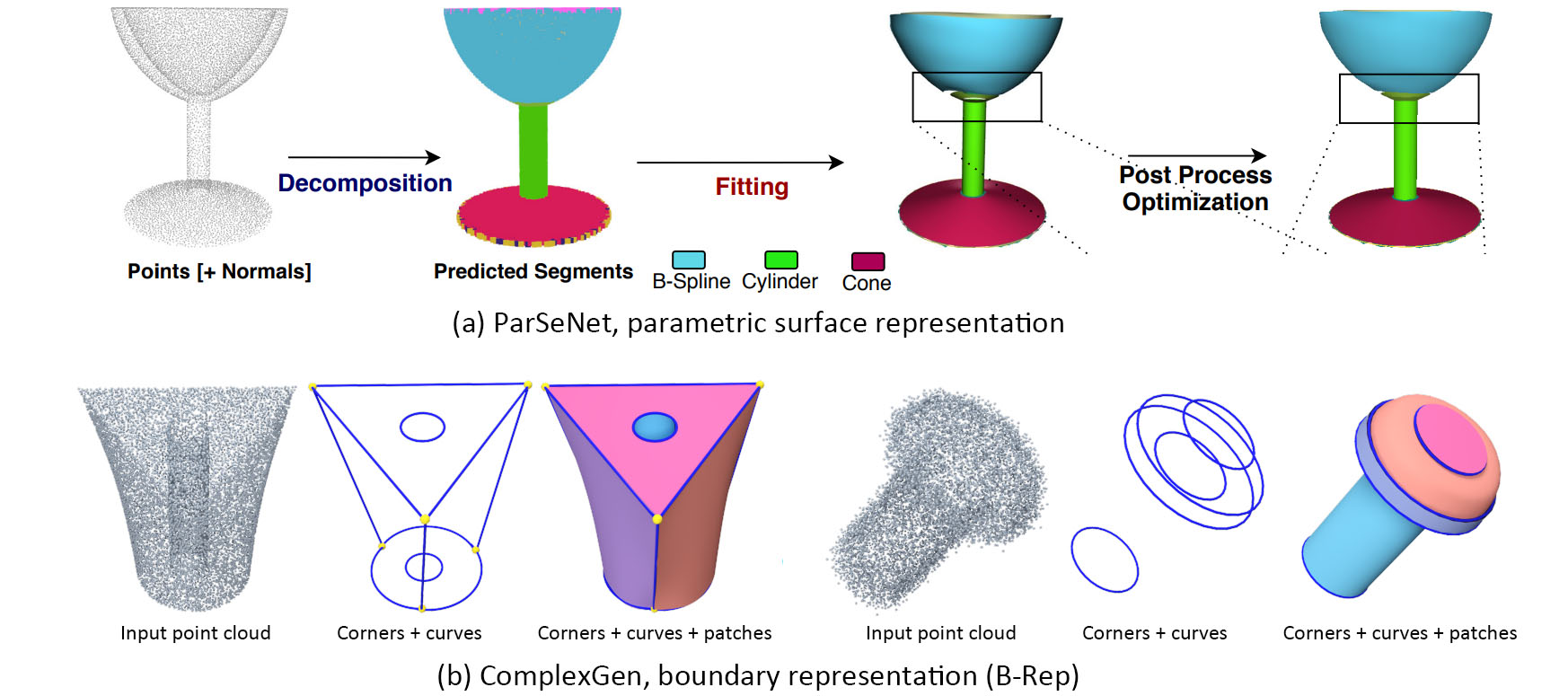}
\end{center}
\caption{
Overview of some methods that perform primitive (corner, edge, patch) detection which eventually leads to shape reconstruction. These figures are taken from their corresponding publications (a)\cite{sharma2020parsenet}, (b)\cite{guo2022complexgen}.
} 
\label{fig:6_primitive_detection}
\end{figure}

The representation in this section, ``primitive detection'', does not have a shape latent space. It is more like an object detection model in computer vision, where the primitive types and parameters are directly predicted from global and local shape features by the neural networks. Therefore, such representation is usually for reconstructing CAD models of mechanical parts. Specifically, SPFN \cite{li2019supervised}, CPFN \cite{le2021cpfn}, ParSeNet \cite{sharma2020parsenet}, HPNet \cite{yan2021hpnet}, and Point2Cyl \cite{uy2022point2cyl} all reconstruct parametric surfaces from point clouds. They first use a neural network to extract per-point features from the input point clouds, and then apply a clustering module to segment the point cloud into patches belonging to different primitives, and finally classify the primitive type and regress the primitive parameters for each patch, as shown in Figure~\ref{fig:6_primitive_detection} (a). SPFN \cite{li2019supervised} and CPFN \cite{le2021cpfn} can reconstruct planes, spheres, cylinders, and cones; ParSeNet \cite{sharma2020parsenet} and HPNet \cite{yan2021hpnet} in addition can reconstruct open and closed B-spline patches. ComplexGen \cite{guo2022complexgen} adopts a more structured representation, boundary
representation (B-Rep), to recover corners, curves (line, circle, B-spline, ellipse) and patches (plane, cylinder, torus, B-spline, cone, sphere) simultaneously along with their mutual topology constraints, as shown in Figure~\ref{fig:6_primitive_detection} (b). Point2Cyl \cite{uy2022point2cyl} utilizes a ``sketch and extrude'' representation, which will be discussed in Scetion \ref{sec:representation_12_sketch_extrude}

\clearpage

\section{Grid mesh}
\label{sec:representation_7_grid_mesh}

Inspired by classic iso-surfacing algorithms such as Marching Cubes (MC) \cite{lorensen1987marching}, Dual Contouring (DC) \cite{ju2002dual}, and Marching Tetrahedra (MT) \cite{doi1991efficient}, which operate on a regular grid structure, several methods have been proposed to also generate a regular grid of parameters, so that surfaces can be extracted cell by cell. Compared to other representations, the advantage of this representation is that a 3D convolutional neural network (CNN) can be applied to produce the grid outputs, and CNNs have been thoroughly studies and heavily optimized to be very efficient and effective, compared to the network architectures (usually MLPs or graph convolutions) used in other representations.

\begin{figure}[b!]
\begin{center}
\includegraphics[width=1.0\linewidth]{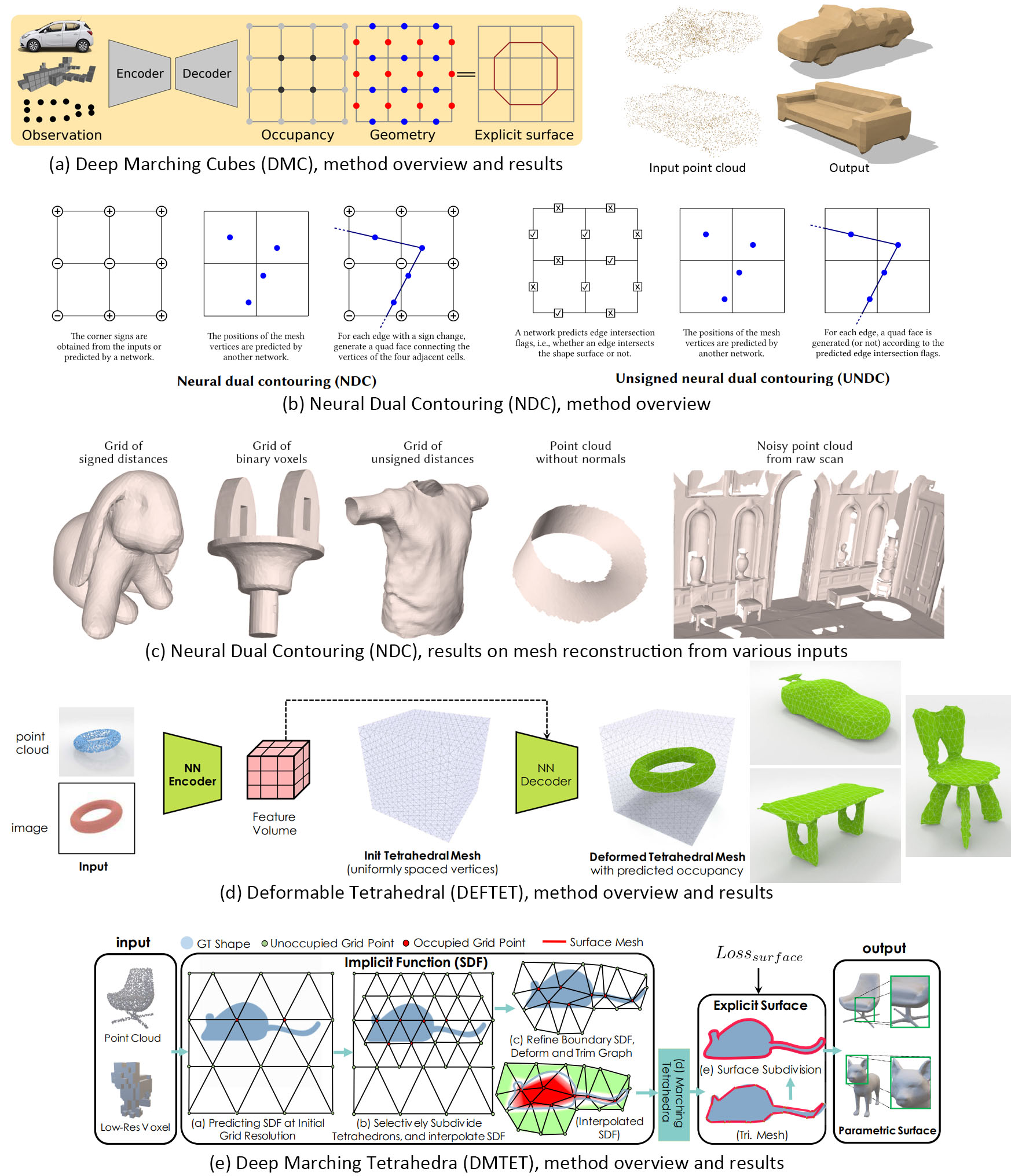}
\end{center}
\caption{
Overview of some methods that output regular grids of parameters that can be parsed into meshes. These figures are taken from their corresponding publications (a)\cite{liao2018deep}, (b)\cite{chen2022ndc}, (c)\cite{chen2022ndc}, (d)\cite{gao2020learning}, (e)\cite{shen2021deep}.
} 
\label{fig:7_grid_mesh}
\end{figure}

One earliest work on this representation is Deep Marching Cubes (DMC) \cite{liao2018deep}, as shown in Figure~\ref{fig:7_grid_mesh} (a). DMC has an encoder-decoder structure, where the encoder encodes the inputs into shape latent codes, and the decoder is a 3D CNN to generate a grid of inside-outside signs and a grid of vertex positions. An explicit triangle mesh can be extracted by applying MC algorithm on the predicted sign grid to determine the topology, i.e., mesh tessellation, in each cell, and the positions of the mesh vertices are given by the predicted grid of vertex positions. Neural Marching Cubes (NMC) \cite{chen2021nmc} has similar idea in spirit, but it targets iso-surfacing tasks such as mesh reconstruction from grids of signed distances or occupancies. NMC has ``local'' backbone networks (fully convolutional without bottleneck layer) with small receptive fields for better generalizability, and it has enriched the MC cube tessellation cases to better reconstruct detailed geometry such as sharp features.

Neural Dual Contouring (NDC) \cite{chen2022ndc} is a follow-up work of NMC, but it adopts DC, a much simpler algorithm compared to MC, as the meshing algorithm. NDC has very simple methods: a ``local'' network generates a grid of signs or intersection flags, another ``local'' network generates a grid of interior vertex positions, and finally DC is applied to connect the generated vertices to form mesh surfaces, as shown in Figure~\ref{fig:7_grid_mesh} (b). NDC can take any input that can be converted into a grid, such as grids of signed or unsigned distances, binary occupancies, and point clouds, so that the ``local'' networks can process them. Some results of NDC are shown in Figure~\ref{fig:7_grid_mesh} (c).

DEFTET \cite{gao2020learning} also predicts a grid structure - a tetrahedron grid, as shown in Figure~\ref{fig:7_grid_mesh} (d). The method predicts the occupancy for each tetrahedron, and the offset for each vertex relative to their initial positions in the regular tetrahedron grid. The output of DEFTET is a tetrahedral mesh, which is one of the dominant representations for volumetric solids in graphics, and it can be directly use in simulation. Note that DEFTET predicts the occupancy for each tetrahedron, unlike DMC, NMC, and NDC where they predict signs on grid vertices. DMTET \cite{shen2021deep} also uses tetrahedron grid, but it does predict signs (signed distance on each grid vertex), so it requires a differentiable iso-surfacing step (differentiable marching tetrahedra), as shown in Figure~\ref{fig:7_grid_mesh} (e). Follow-up work Nvdiffrec \cite{munkberg2022extracting} combines DMTET and differentiable rendering to reconstruct meshes from multi-view images.

\clearpage

\section{Grid polygon soup}
\label{sec:representation_8_grid_polygon_soup}

\begin{figure}[b!]
\begin{center}
\includegraphics[width=1.0\linewidth]{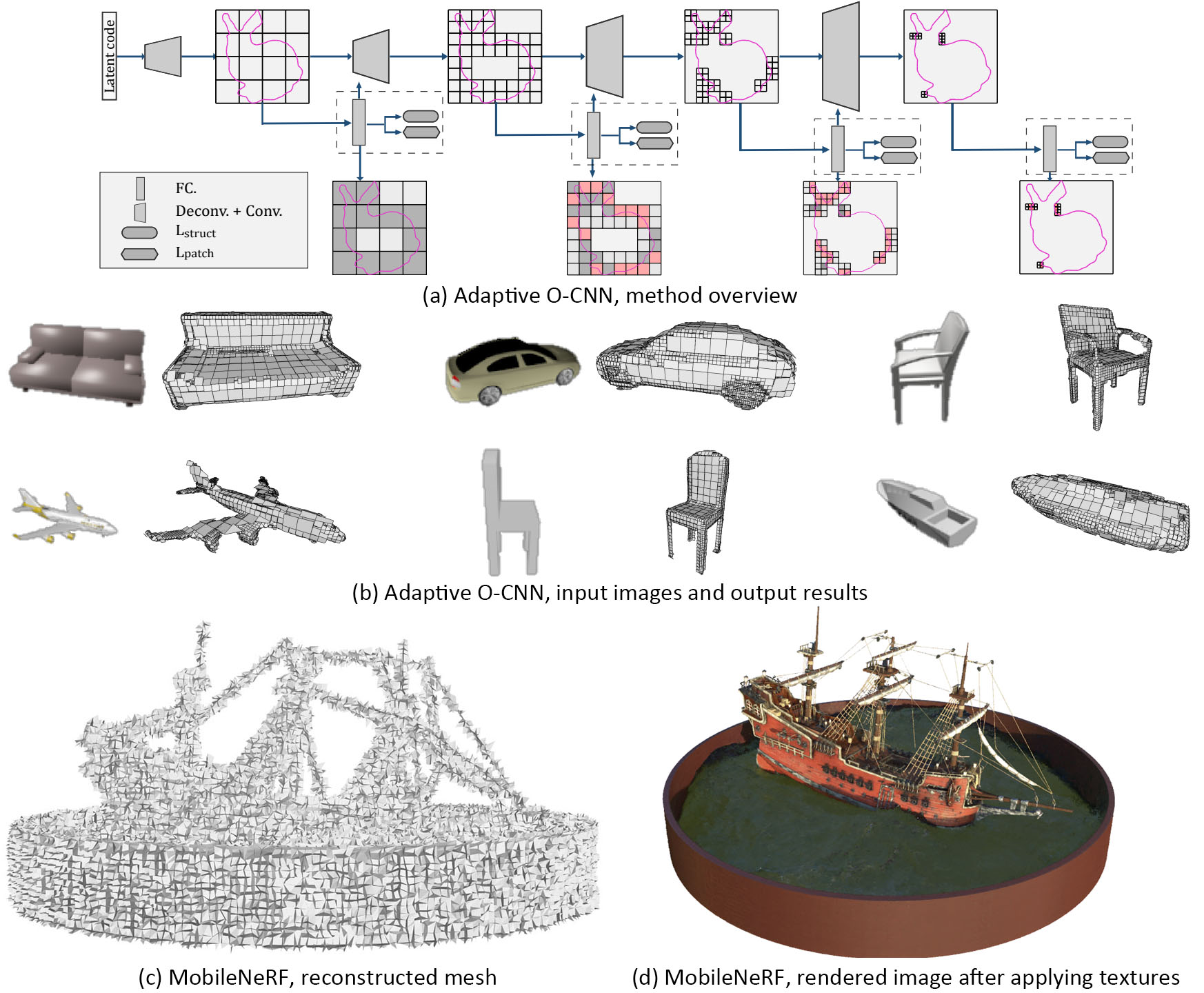}
\end{center}
\caption{
Overview of some methods that output polygon soups. These figures are taken from their corresponding publications (a)(b)\cite{wang2018adaptive}, (c)(d)\cite{chen2022mobilenerf}.
} 
\label{fig:8_grid_polygon_soup}
\end{figure}

This is not a common representation as it tends to produce messy outputs - polygon soups, as the name suggests. Representative works include Adaptive O-CNN \cite{wang2018adaptive} and MobileNeRF \cite{chen2022mobilenerf}.
Adaptive O-CNN \cite{wang2018adaptive} uses an octree-like network structure that subdivides nodes according to the expected complexity of surface details, as shown in Figure~\ref{fig:8_grid_polygon_soup} (a). Each leaf node predicts the parameters of a plane ($\mathbf{n}$ and $b$ in $\mathbf{n} \cdot \mathbf{p}+b=0$). Some results are shown in Figure~\ref{fig:8_grid_polygon_soup} (b).
MobileNeRF \cite{chen2022mobilenerf} can reconstruct geometry from multi-view images. It uses a regular grid mesh as the initial mesh, and optimizes its vertex positions and face occupancies during training. However, the output mesh is coarse and has poor quality, as shown in Figure~\ref{fig:8_grid_polygon_soup} (c). The output mesh must be coupled with the learned texture maps to produce high-quality renderings, as shown in Figure~\ref{fig:8_grid_polygon_soup} (d).

\clearpage

\section{Grid voxels}
\label{sec:representation_9_grid_voxel_SDF}

Similar to ``Grid mesh'' where grid structures are used to store local mesh properties, it is in fact easier for the neural networks to directly predict a grid of voxels carrying implicit field values, such as signed distances or occupancies, and then apply meshing or iso-surfacing algorithms to extract the mesh. Occupancy or signed distance grids have been very popular representations, and the reason is similar to that of ``Grid mesh'', that a mature technique, convolutional neural networks (CNN), can be applied to produce the grid outputs. Due to the popularity and wide usage of voxels, there are thousands of publications that adopt this representation. Therefore, this section will only cover a few representative works to explain the various use cases.

\begin{figure}[b!]
\begin{center}
\includegraphics[width=1.0\linewidth]{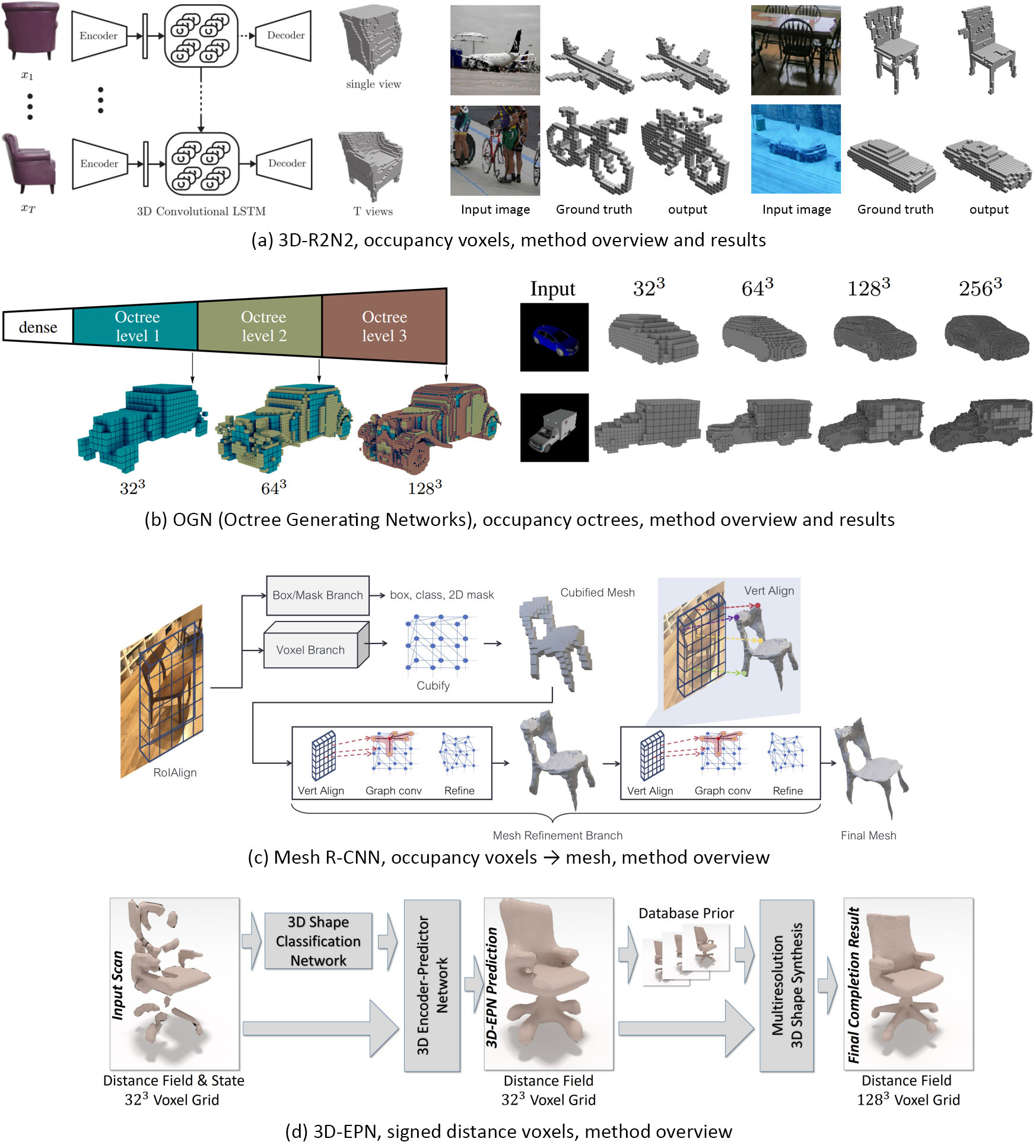}
\end{center}
\caption{
Overview of some methods that output grids of occupancies or signed distances. Note that most methods that output occupancy voxels directly create meshes from the cubic voxels, such as in (a)(b)(c). Methods that output signed distance voxels typically apply marching cubes to extract the mesh, as in (d). These figures are taken from their corresponding publications (a)\cite{choy20163d}, (b)\cite{tatarchenko2017octree}, (c)\cite{gkioxari2019mesh}, (d)\cite{dai2017shape}.
} 
\label{fig:9_grid_voxel_SDF}
\end{figure}

The simplest way to generate voxels is to use a 3D CNN network to predict a 3D grid of occupancies. 3D-R2N2 \cite{choy20163d} is one of the most famous and highly cited papers in single-image shape reconstruction, because it provided a dataset of rendered ShapeNet \cite{chang2015shapenet} images that everyone uses in their benchmark tests. 3D-R2N2 has a simple network structure, as shown in Figure~\ref{fig:9_grid_voxel_SDF} (a). If one focus on the upper half of the network, it is an encoder-decoder structure where a 2D CNN encoder encodes the input image into a latent code, and a 3D CNN decoder decodes the latent code into a voxel grid. To take multiple images from different views as input, 3D-R2N2 utilizes Long Short-Term Memory (LSTM) \cite{hochreiter1997long} to aggregate the sequence of latent codes before feeding them to the shape decoder.

Note that 3D-R2N2 has a bottleneck in the network architecture to produce a global latent code that is not capable of representing detailed shape geometries and is unlikely to generalize to inputs dissimilar to those in the training set. Some works that adopt voxels use a local network or U-Net \cite{ronneberger2015u} to take into account local features, such as DECOR-GAN \cite{chen2021decor} and GenRe \cite{zhang2018learning}.

However, voxel grids are of $O(N^3)$ space complexity. It is hard to generate a voxel grid of sufficiently high resolution due to hardware memory constraints. Therefore, octree representation which adaptively subdivides voxels has been applied in many works, such as HSP (Hierarchical Surface Prediction) \cite{hane2017hierarchical}, 3D-CFCN \cite{cao2018learning}, OctNetFusion \cite{riegler2017octnetfusion}, OGN \cite{tatarchenko2017octree}, and Dual OCNN\cite{Wang2022Dual}. An example is shown in Figure~\ref{fig:9_grid_voxel_SDF} (b); one can observe that voxels are subdivided where details need to be represented.

One can also apply a refinement neural network on the extracted meshes to improve the mesh quality, such as in Mesh R-CNN \cite{gkioxari2019mesh}, as shown in Figure~\ref{fig:9_grid_voxel_SDF} (c).
Although the result looks worse in my honest opinion.

Finally, predicting grids of signed distances rather than occupancies can better model smooth surfaces, as signed distances contain more information about the surface than binary occupancies. Example works include 3D-EPN \cite{dai2017shape} and Deep level sets \cite{michalkiewicz2019deep}, as shown in Figure~\ref{fig:9_grid_voxel_SDF} (d).

\clearpage
\section{Neural implicit}
\label{sec:representation_10_neural_implicit}

Neural implicit representation is an extremely popular representation nowadays. It was proposed in CVPR 2019 by three concurrent works IM-Net \cite{chen2019imnet}, OccNet \cite{mescheder2019occupancy}, and DeepSDF \cite{park2019deepsdf}. Since then there has been an explosion of neural implicit papers. This section will only list a few representative works. We refer the readers to the survey ``Neural Fields in Visual Computing and Beyond'' \cite{xie2022neural} for more related works.

Neural implicit representation is essentially an MLP (MultiLayer perceptron) that takes a point's coordinates as input and outputs the inside-outside sign \cite{chen2019imnet,mescheder2019occupancy} or signed distance \cite{park2019deepsdf} of that input point. The MLP itself represents the implicit function of a 3D shape. To generate different output shapes based on the input, the MLP can be conditioned via concatenating a shape latent code with the input point coordinates before feeding the point to the MLP \cite{chen2019imnet,park2019deepsdf}, or modulating the MLP network weights using the latent code \cite{mescheder2019occupancy,sitzmann2020implicit}. The main difference between neural implicit and ``grid voxels'' is that ``grid voxels'' uses 3D CNNs to directly predict a grid of implicit field values, therefore it outputs the entire shape with one network forward pass, whereas neural implicit processes each single input point individually with the same MLP. To output the entire shape, a grid of points need to be sampled in space and the MLP needs to run on each of the points to produce a grid of implicit field values. Marching Cubes (MC) \cite{lorensen1987marching} is usually applied to the grid to extract an explicit mesh. Theoretically, neural implicit can represent shapes with infinitely fine resolution without increasing the model (MLP) size, which is in contrast to ``grid voxels'' which has an $O(N^3)$ space complexity, therefore neural implicit has been considered as a compact representation.

Structured Implicit Functions \cite{genova2019learning} propose to use scaled axis-aligned anisotropic 3D Gaussians (i.e., a set of Gaussian balls) to represent the implicit field, instead of an MLP. It can speed up training and inference of neural implicit and has been adopted in follow-ups that have global shape latent codes.

However, global shape latent codes cannot represent detailed shape geometries and they often lead to overfitting on a specific dataset or category of shapes, therefore methods have been proposed to include local features to condition the MLP so that the network can have better accuracy on local details and generalize better. PIFu \cite{saito2019pifu}, PIFuHD \cite{saito2020pifuhd}, DISN \cite{xu2019disn}, and D$^2$IM-Net \cite{li2021d2im} employ local features from 2D image encoders for single-view 3D reconstruction tasks. LIG \cite{jiang2020local}, ConvONet (Convolutional occupancy networks) \cite{peng2020convolutional}, SA-ConvONet \cite{tang2021sa}, POCO \cite{boulch2022poco}, and IF-Nets \cite{chibane2020implicit} employ local features extracted from point clouds or voxels to perform object or scene reconstructions.

Differentiable rendering has also been developed for neural implicit. \cite{liu2019learning}, DVR \cite{niemeyer2020differentiable}, IDR \cite{yariv2020multiview}, and DIST \cite{liu2020dist} can reconstruct objects from multi-view images; they assume the object mask is given for each input image, and each ray intersects the surface at most once (only one intersection point per ray for the gradient to propagate). NeuS \cite{wang2021neus}, HF-NeuS \cite{wanghf}, UNISURF \cite{oechsle2021unisurf}, VolSDF \cite{yariv2021volume}, and \cite{azinovic2022neural} also reconstruct objects or scenes from multi-view images. However, they are based on the ray marching volume rendering formula of NeRF \cite{mildenhall2021nerf}. They do not need object masks, and they sample numerous points along each ray to perform volume rendering.

Other interesting works include:
SIREN \cite{sitzmann2020implicit} and Fourier Feature Networks \cite{tancik2020fourier} for improving the representation ability of neural implicit by applying sine activation functions and Fourier feature mapping, respectively;
InstantNGP \cite{mueller2022instant} for greatly improving the training speed by applying multi-resolution Hash encoding;
\cite{liu20222d} for introducing a method to reconstruct 3D shapes from single-view 2D images with the help of pseudo multi-view images generated by StyleGAN \cite{karras2019style};
MeshSDF \cite{remelli2020meshsdf} for designing differentiable iso-surface extraction algorithm (Marching Cubes) on neural implicit;
SAL (Sign Agnostic Learning) \cite{remelli2020meshsdf} for proposing a method to learn neural implicit from unoriented point clouds;
NeuralUDF \cite{chibane2020neural} and MeshUDF \cite{guillard2022meshudf} for proposing Neural unsigned distance fields and how to extract explicit meshes from them, respectively;
GIFS \cite{ye2022gifs} for proposing a neural field to represent general shapes including non-watertight shapes and
shapes with multi-layer surfaces, by predicting whether two points are separated by any surface, instead of the inside-outside of each point.

\begin{figure}[b!]
\begin{center}
\includegraphics[width=1.0\linewidth]{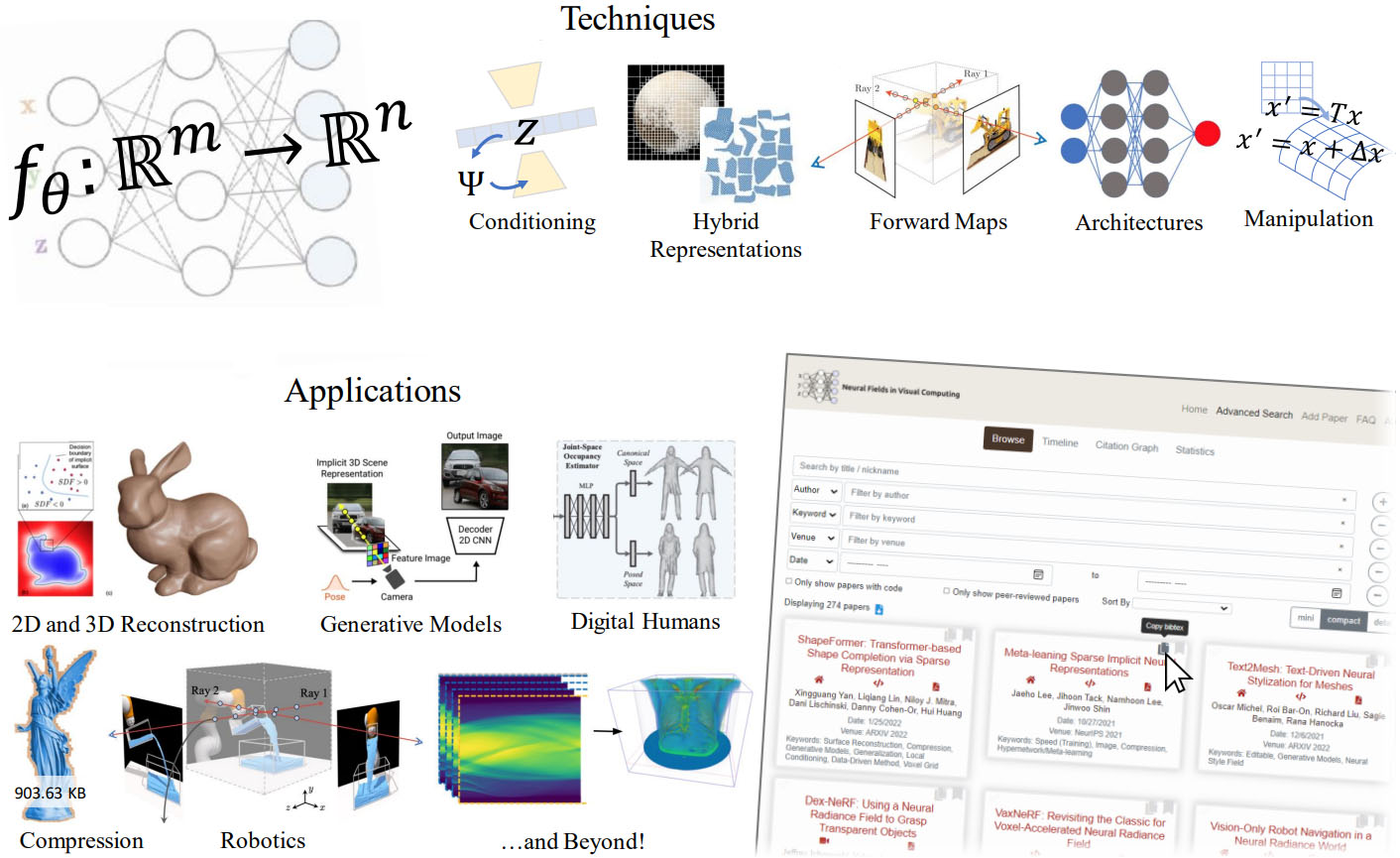}
\end{center}
\caption{
Overview of neural implicit. This entire figure is taken from the survey \cite{xie2022neural}, which contains figures from various neural implicit papers.
} 
\label{fig:10_neural_implicit}
\end{figure}

\clearpage

\section{Primitive CSG}
\label{sec:representation_11_CSG}

\begin{figure}[b!]
\begin{center}
\includegraphics[width=1.0\linewidth]{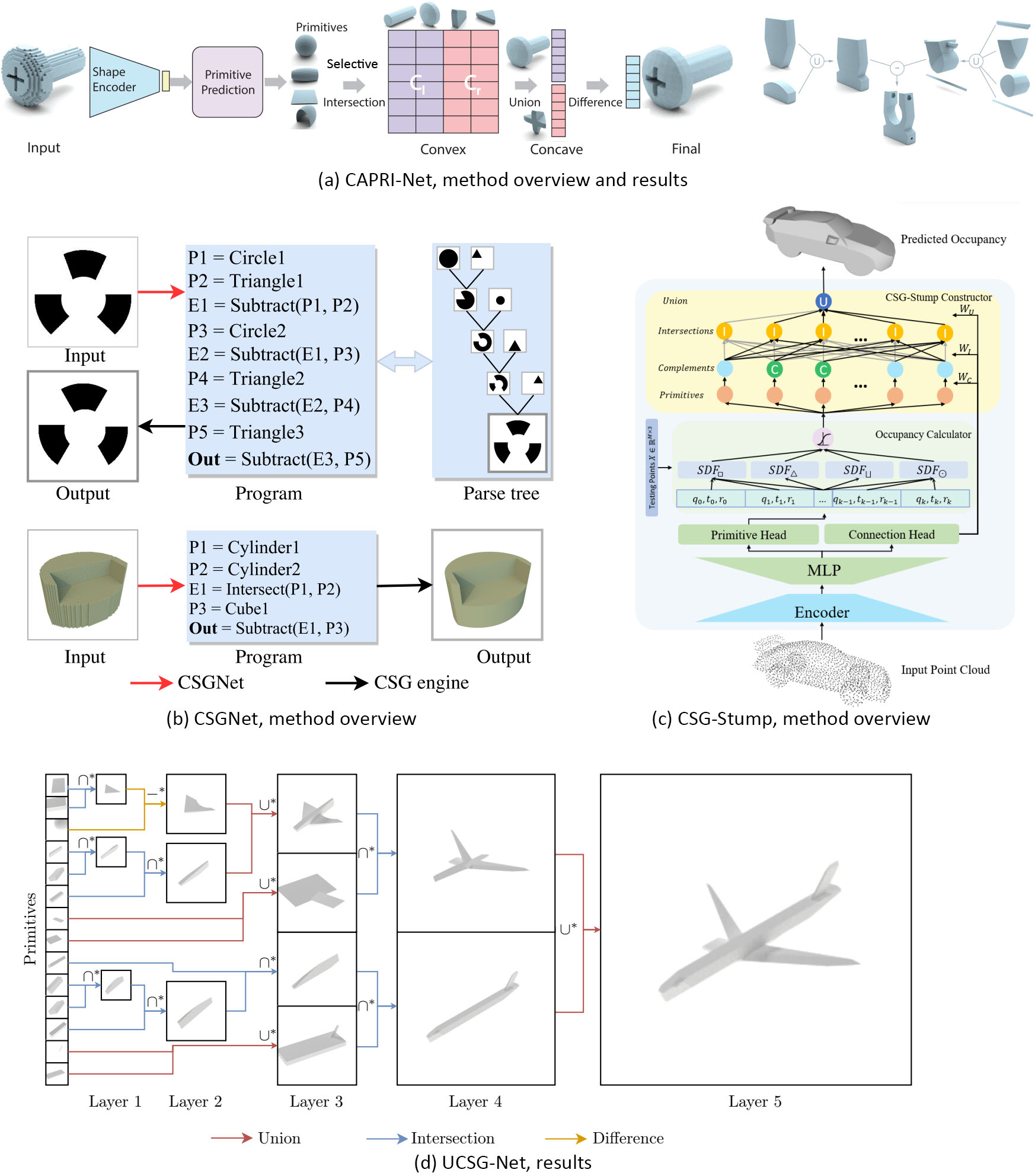}
\end{center}
\caption{
Overview of some methods that output CSG-trees. These figures are taken from their corresponding publications (a)\cite{yu2022capri}, (b)\cite{sharma2018csgnet}, (c)\cite{ren2021csg}, (d)\cite{kania2020ucsg}.
}
\label{fig:11_CSG}
\end{figure}

CSG is short for ``Constructive Solid Geometry''. It is a common representation for CAD (computer aided design) models, where primitive shapes such as polyhedrons, ellipsoids, and cylinders are merged via boolean operators such as union, intersection, and difference. A simple CSG tree is shown in Figure~\ref{fig:11_CSG} (a) right, and a complex CSG tree is shown in Figure~\ref{fig:11_CSG} (d). CSG representation is a compact representation that is able to preserve sharp and smooth geometric features.

When ground truth CSG-trees are available, a network can be trained with fully-supervised learning to produce a sequence of operations and their operands, where the sequence is equivalent to a CSG-tree. A representative work is CSGNet \cite{sharma2018csgnet}, as shown in Figure~\ref{fig:11_CSG} (b).

Yet when ground truth is unavailalbe, learning to represent a shape as a CSG-tree with neural networks is a very challenge task, because CSG trees include selecting suitable primitives and performing CSG operations. Both steps are discrete and non-differentiable. BSP-Net \cite{chen2020bsp} from section \ref{sec:representation_5_set_of_primitives} ``Set of primitives'' could be considered as using CSG representation, since it intersects half-spaces to form convex shapes and then unions convex shapes to form the output shape. BSP-Net has two training stages: in the continuous stage, the primitive selection and the CSG operations are approximated by continuous linear operations (weighted sum) and ReLU (rectified linear unit); in the second stage, the continuous weights are binarized via naive thresholding, and the continuous operations (sum) are replaced with discrete operations (max). Therefore, BSP-Net first learns a continuous approximation of the CSG-tree, and then discretize it to obtain a real CSG-tree.

BSP-Net only adopts planes (half-spaces) as primitives, therefore it only approximates piece-wise planar shapes. Its follow-up work CAPRI-Net \cite{yu2022capri} relaxed the primitive types to axis-aligned quadratic surfaces, therefore it is able to represent smooth surfaces with a single primitive. In addition, CAPRI-Net also introduced a differentiable difference operator. The predicted quadratic surface primitives are first selected and intersected to form convex shapes; then the convex shapes are unioned to form two potentially concave shapes; finally the difference of the two concave shapes forms the output shape. These steps are illustrated in Figure~\ref{fig:11_CSG} (a) left.

While CAPRI-Net uses quadratic surfaces as primitives, other works use primitive shapes as primitives. UCSG-Net \cite{kania2020ucsg} predicts parameters of boxes and spheres and use them as primitives. Similarly CSG-Stump \cite{ren2021csg} predicts boxes, spheres, cylinders and cones. Both works use modified softmax functions to approximate discrete operations (min,max). While UCSG-Net has several CSG layers where each layer supports multiple types of CSG operations, as shown in Figure~\ref{fig:11_CSG} (d); CSG-Stump only has three layers with fixed ordering (complement-intersection-union), similar to CAPRI-Net (intersection-union-difference), as shown in Figure~\ref{fig:11_CSG} (c).

\clearpage

\section{Sketch and extrude}
\label{sec:representation_12_sketch_extrude}

\begin{figure}[b!]
\begin{center}
\includegraphics[width=1.0\linewidth]{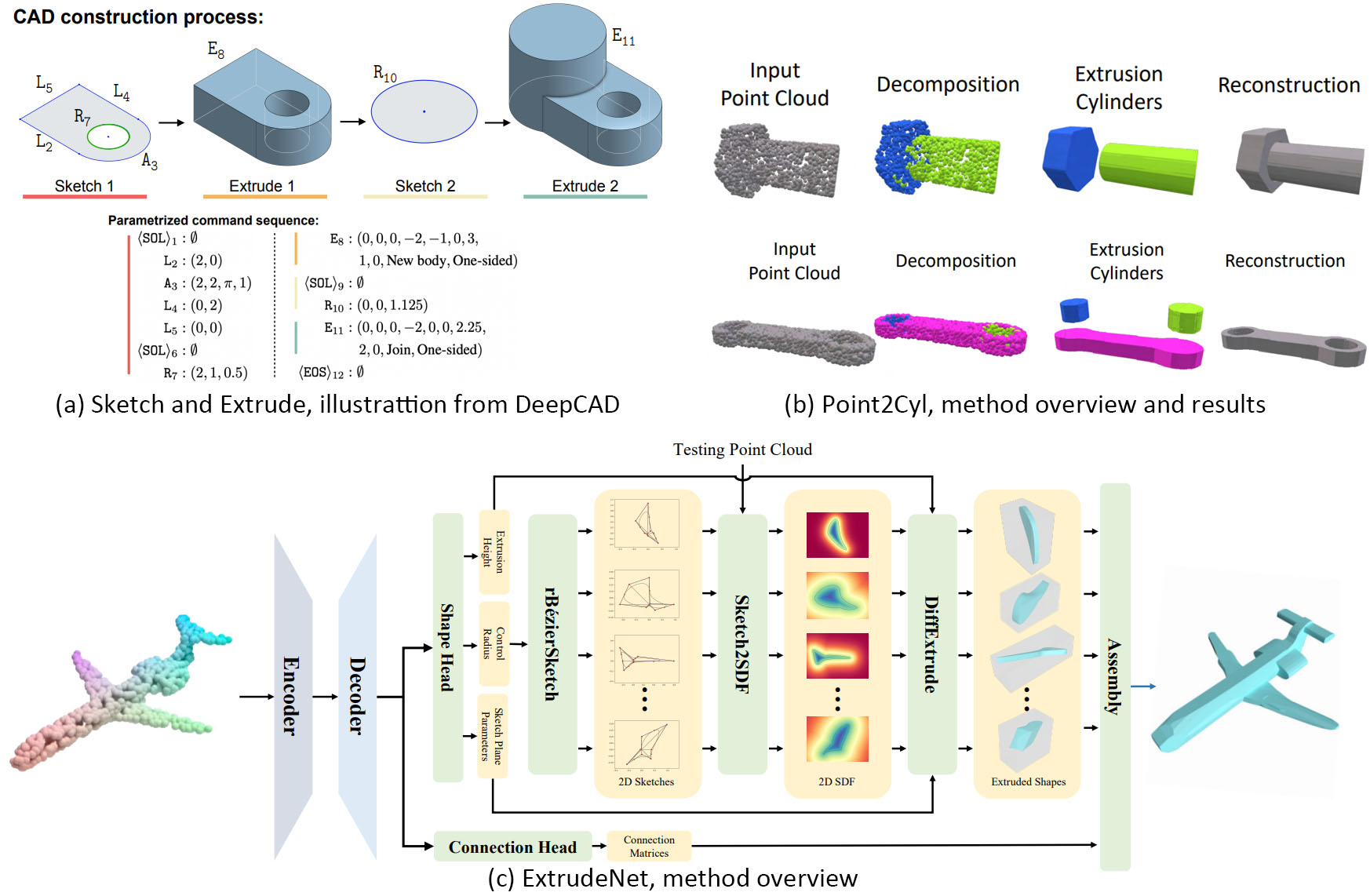}
\end{center}
\caption{
Overview of some methods that output sketch and extrude sequences. These figures are taken from their corresponding publications (a)\cite{wu2021deepcad}, (b)\cite{uy2022point2cyl}, (c)\cite{ren2022extrudenet}.
}
\label{fig:12_sketch_extrude}
\end{figure}

``Sketch and extrude'' is also a CSG representation. However, different from ``Primitive CSG'' where only primitive 3D shapes are used for CSG operations, this representation takes an approach more similar to the workflow of creating CAD models, by repeatedly sketching a 2D profile and then extruding that profile into a 3D body. DeepCAD \cite{wu2021deepcad} is a pioneer in generating CAD models with such a representation, by applying a Transformer network \cite{vaswani2017attention} to predict the command sequences for creating the output shape, as shown in Figure~\ref{fig:12_sketch_extrude} (a). Follow-up work \cite{lambourne2022reconstructing} reconstructs CAD shapes from rounded voxel models; it represents profiles as 2D occupancy images. Point2Cyl \cite{uy2022point2cyl} reconstructs CAD shapes from point clouds; it represents profiles as 2D neural implicit. Point2Cyl is also considered a representative work in ``primitive detection'', since it first segments the point cloud into patches, and then reconstruct extrusion cylinders from those patches, as shown in Figure~\ref{fig:12_sketch_extrude} (b).

The works mentioned above all apply supervised training with ground truth command sequences. ExtrudeNet \cite{ren2022extrudenet} can learn such sequences unsupervisedly. It uses closed cubic Bezier curves as profiles and proposed a differentiable sketch-to-SDF module and a differentiable extrusion module to construct 3D parts, as shown in Figure~\ref{fig:12_sketch_extrude} (c). Those 3D parts are combined using CSG-Stump \cite{ren2021csg} to form the output shape.

\clearpage
\section{Connect given vertices}
\label{sec:representation_13_connect_V}

This representation can only be used for reconstructing a mesh from a point cloud, because it only connects given vertices from the point cloud. The methods can be classified by whether it infers the inside-outside regions of the shape and thus generates a closed mesh.

\begin{figure}[b!]
\begin{center}
\includegraphics[width=0.95\linewidth]{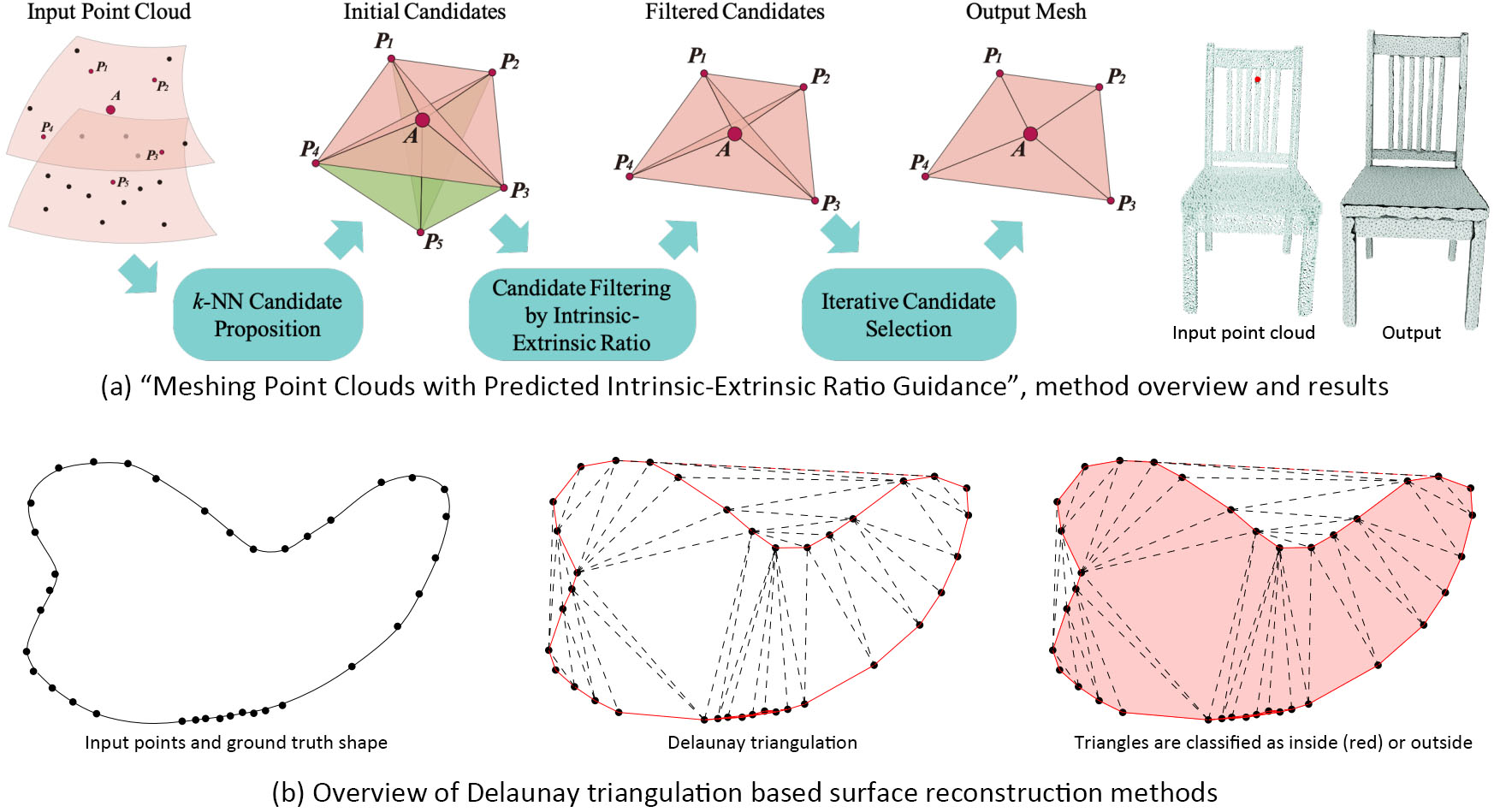}
\end{center}
\vspace{-8mm}
\caption{
Overview of some methods that output sketch and extrude sequences. Figure (a) is taken from \cite{wu2021deepcad}, (b) is modified from a figure in the survey \cite{cazals2006delaunay}.
}
\label{fig:13_connect_V}
\end{figure}

Most methods do not generate a closed mesh. \cite{liu2020meshing} generates a collection of triangles by proposing candidate triangles, classifying the candidate triangles with a neural network to determine which triangles should exist in the output mesh, and repeating this process, as shown in Figure~\ref{fig:13_connect_V} (a). PointTriNet \cite{sharp2020pointtrinet} has a classification network to classify whether a given candidate triangle should exist in the output mesh, and a proposal network to suggest likely neighbor triangles for a given existing triangle. PointTriNet alternates between these networks (classifying candidate triangles and proposing new candidates) to generate a collection of triangles as the output mesh. \cite{rakotosaona2021learning} first estimates local neighborhoods around each point, and then perform a 2D projection of these neighborhoods so that a 2D Delaunay triangulation is computed to provide candidate triangles, and finally those candidate triangles are aggregated to maximize the manifoldness of the reconstructed mesh.

Delaunay triangulation based surface reconstruction methods can guarantee to generate a closed mesh because they first perform 3D Delaunay triangulation on the input points to obtain a tessellation of the 3D space with tetrahedrons, and then classify which tetrahedrons are inside the shape and which are outside. After the inside-outside labels are assigned, a surface can be reconstructed by extracting triangle faces between tetrahedrons of different labels, as shown in Figure~\ref{fig:13_connect_V} (b). DeepDT \cite{luo2021deepdt} uses a graph neural network on the dual graph of the Delaunay triangulation to predict the label of each tetrahedron.

\vspace{-10mm}

\clearpage
\section{Generate and connect vertices}
\label{sec:representation_14_generate_VT}

This representation first generates a set of mesh vertices with a neural network, and then selectively connects those vertices to form mesh faces with another neural network.
The representation can directly generate a 3D mesh as an indexed face set, however, it is rarely used due to its extremely high complexity.

\begin{figure}[b!]
\begin{center}
\includegraphics[width=1.0\linewidth]{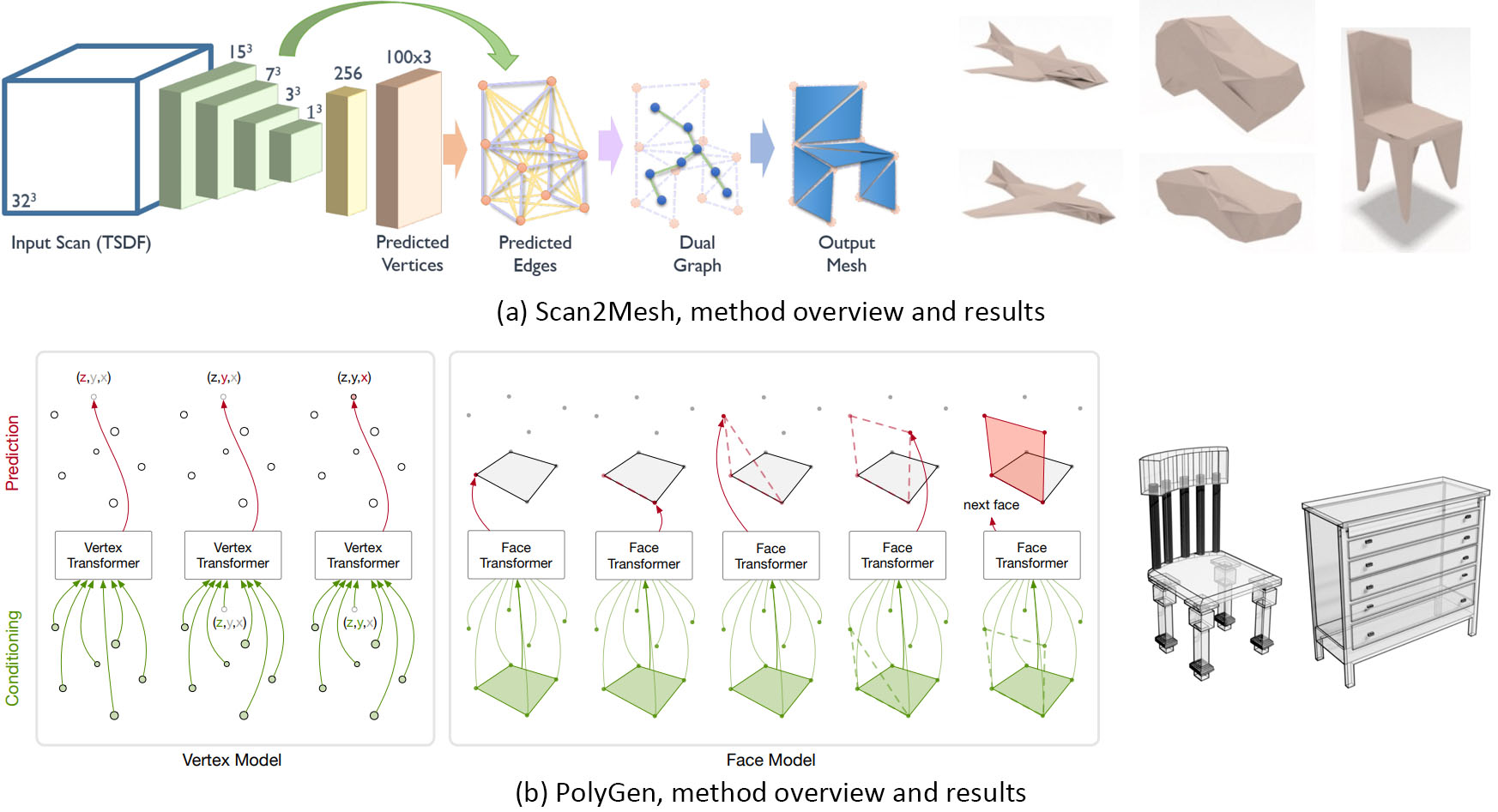}
\end{center}
\vspace{-8mm}
\caption{
Overview of some methods that first generate vertices and then connect those vertices to form faces. These figures are taken from their publications (a)\cite{dai2019scan2mesh}, (b)\cite{nash2020polygen}.
}
\label{fig:14_generate_VT}
\end{figure}

Scan2Mesh \cite{dai2019scan2mesh} uses a point cloud generator (MLP) to generate a set of points. Then it constructs a fully connected graph on these points, and uses a graph neural network to predict which mesh edges should exist in the output mesh. Finally it considers all possible triangle faces that can be formed from the predicted edges, constructs a dual graph on the faces, and uses a graph neural network to predict which mesh faces should exist in the output mesh. The method overview is shown in Figure~\ref{fig:14_generate_VT} (a). Due to the construction of a complete graph on the predicted points and the subsequent graph neural networks, this method can only predict a limited number of vertices (100 in the experiments).

PolyGen \cite{nash2020polygen} first generates mesh vertices sequentially from lowest to highest on the vertical axis. The continuous vertex positions are quantized to form discrete bins for likelihood calculation. The next vertex is generated by a vertex Transformer, which takes the current sequence of vertex positions as input, and outputs a distribution over discretized vertex positions. Then it generates polygon faces, also sequentially from lowest to highest on vertex indices. The next face is generated by a face Transformer, which takes the generated vertices and the current sequence of face indices as input, and outputs a distribution over vertex indices. The method overview is shown in Figure~\ref{fig:14_generate_VT} (b).

\vspace{-10mm}

\clearpage
\section{Sequence of edits}
\label{sec:representation_15_sequence_of_edits}

\begin{figure}[b!]
\begin{center}
\includegraphics[width=1.0\linewidth]{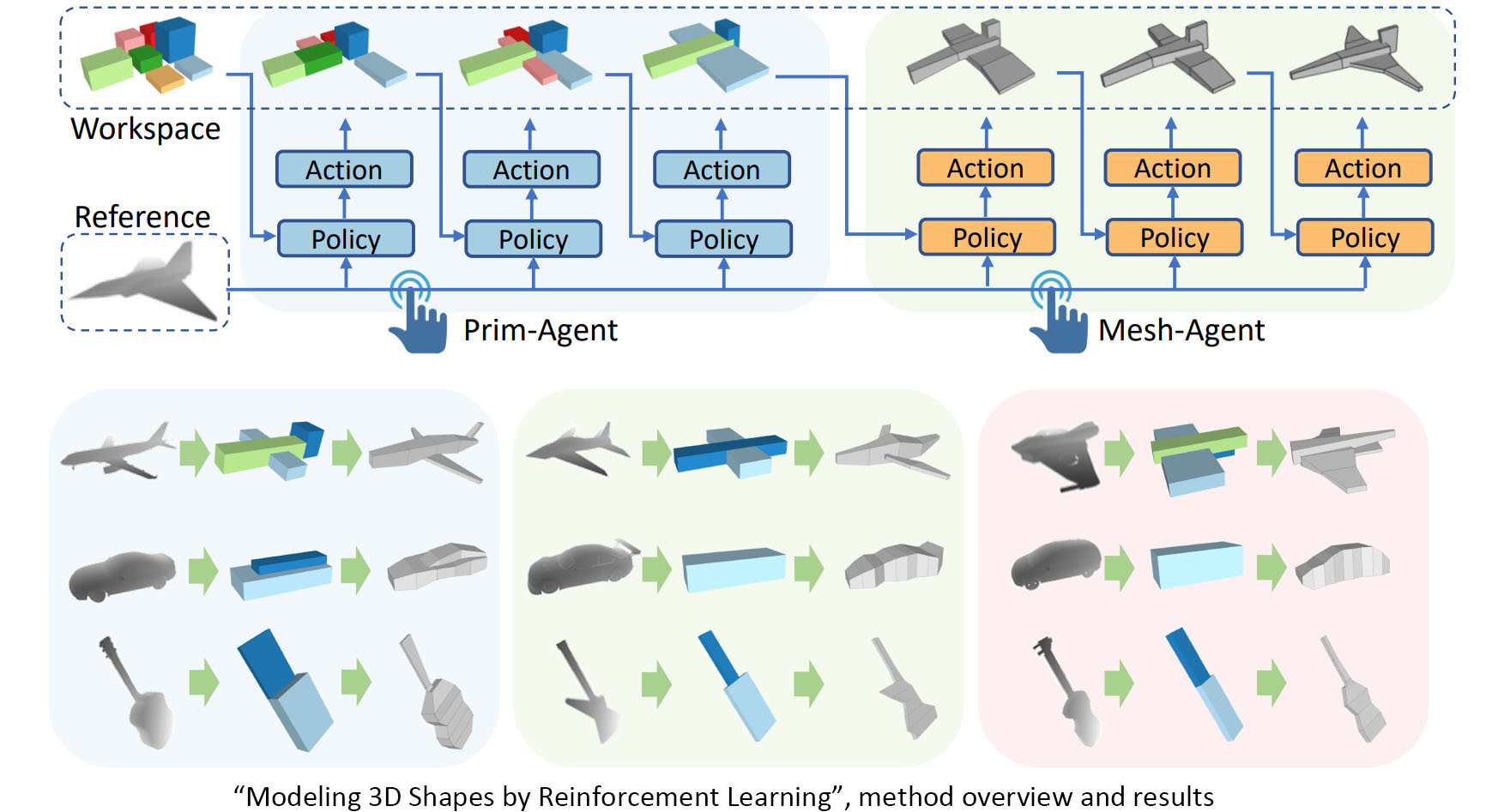}
\end{center}
\caption{
Overview of \cite{lin2020modeling} that learns a sequence of editing operations via reinforcement learning for shape reconstruction. The figures is taken from \cite{lin2020modeling}.
}
\label{fig:15_sequence_of_edits}
\end{figure}

This section was planned to cover works that model 3D shapes using reinforcement
learning to generate a sequence of editing operations. However, such works are very scarce and now it seems this section is dedicated for a single paper, ``Modeling 3D Shapes by Reinforcement Learning'' \cite{lin2020modeling}. The method contains two neural networks, a Prim-Agent that approximates the shape using primitives, and a Mesh-Agent that edits the mesh to create detailed geometry, as shown in Figure~\ref{fig:15_sequence_of_edits}. Given a depth image as shape reference and a set of pre-defined primitives, the Prim-Agent predicts a sequence of
actions on primitives (drag primitive corner points or delete primitive) to approximate the target shape. Then the edge loops are added to the output primitives to subdivide each primitive into segments for finer editing control. Finally, the Mesh-Agent takes as input the shape reference and the primitive-based representation, and predicts actions on edge loops (drag edge loop corner points) to create detailed geometry.

\clearpage

\chapter{Reconstruction from Voxels}
\label{sec:voxel}

In this section, we review works that reconstruct shapes from a grid of occupancies or signed distances. Based on the motivations, we divide the collection of works into two categories: shape super-resolution and shape parsing, where shape super-resolution reconstructs a more detailed and visually pleasing shape from the input voxels, and shape parsing decomposes the input voxels into primitives and CSG sequences for reverse engineering a CAD shape. Note that reconstruct complete shapes from partial voxels, i.e., shape completion, is also an active research field, which we will not discuss due to the scope of this survey.

\section{Shape super-resolution}
\label{sec:voxel_1_superresolution}

Shape super-resolution strives to recover and even enhance geometric features from voxel inputs. Methods such as OccNet \cite{mescheder2019occupancy} and IM-Net \cite{chen2019imnet} can convert input voxels into neural implicit representation, which can be sampled at arbitrarily high resolution. Although one can argue that the outputs of those methods indeed have higher resolution compared to the input voxels, they do not properly recover shape details; in fact, the reconstructed outputs often lose details presented in the input voxels, due to the usage of a global shape latent code. Therefore, works that perform voxel super-resolution mostly adopt local neural networks which take into account both local and global shape features.

\begin{figure}[b!]
\begin{center}
\includegraphics[width=1.0\linewidth]{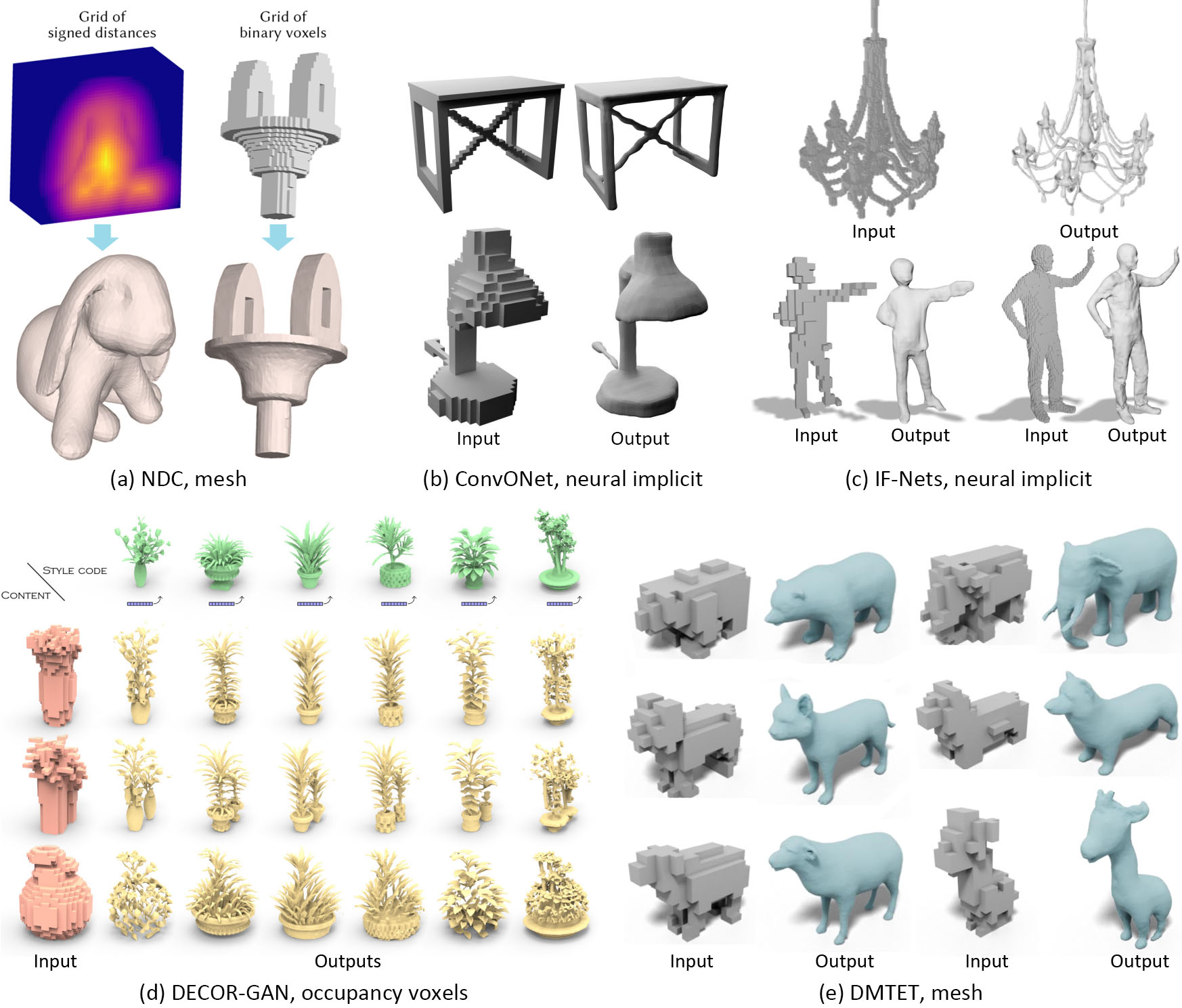}
\end{center}
\vspace{-8mm}
\caption{
Overview of some single image super-resolution works. These figures are taken from their corresponding publications (a)\cite{chen2022ndc}, (b)\cite{peng2020convolutional}, (c)\cite{chibane2020implicit}, (d)\cite{shen2021deep}, (e)\cite{chen2021decor}.
}
\label{fig:voxel_1_superresolution}
\end{figure}

Neural Marching Cubes (NMC) \cite{chen2021nmc} and Neural Dual Contouring (NDC) \cite{chen2022ndc} are data-driven iso-surfacing algorithms. They reconstruct meshes from input voxels and they are able to recover geometric features such as sharp corners and edges, as shown in Figure~\ref{fig:voxel_1_superresolution} (a).
They adopt and modify the mesh tessellations in classic iso-surfacing algorithms Marching Cubes (MC) \cite{lorensen1987marching} and Dual Contouring (DC) \cite{ju2002dual}, and use neural networks to predict the tessellation case and the vertex positions in each voxel to directly output a polygonal mesh. Their neural network backbones have limited receptive fields, meaning that they can only infer geometric features from local regions and they do not have access to global shape information. Thus, the methods are considered focusing more on the mesh reconstruction side than the shape super-resolution side.

Similarly, methods such as Convolutional occupancy networks (ConvONet) \cite{peng2020convolutional} and ``Implicit Functions in feature space'' (IF-Nets) \cite{chibane2020implicit} can reconstruct shapes as neural implicit from input voxels.
ConvONet adopts convolutional encoders to process the input voxels, and predict either a 3D grid of deep features (grid setting) or three 2D grids of deep features on three orthogonal planes (tri-plane setting). Then for each query point, the deep features are retrieved from the grid or the planes via trilinear or bilinear interpolation, and the deep features are concatenated with the query point coordinates to be fed into an MLP to predict the inside/outside status of the query point.
IF-Nets also adopts convolutional encoders to compute multi-scale 3D grids encoding global and local features. Then for each query point, the deep features are retrieved from those grids via trilinear interpolation, and then the deep features (without query point coordinates) are fed into an MLP to predict the inside/outside status of the query point. Note that the backbone networks in these works are sufficiently large to produce global shape features, yet they also utilize local features.
They are adept at recovering geometric features from reasonably dense input voxels, as shown in Figure~\ref{fig:voxel_1_superresolution} (b)(c). However, they do not perform well on coarse inputs, since they only recover details and do not create more details on the coarse shape, as shown in Figure~\ref{fig:voxel_1_superresolution} (c) bottom.

Therefore, methods have been proposed to generate new details on the coarse shape, mostly with the help of a local patch discriminator\cite{goodfellow2020generative} (PatchGAN\cite{isola2017image}).
DECOR-GAN \cite{chen2021decor} performs shape detailization by treating the problem as that of geometric detail transfer. Given a low-resolution coarse voxel shape as input, DECOR-GAN upsamples it into a higher-resolution voxel shape enriched with geometric details while preserving the overall structure (or content) of the input. The geometric details are learned from a set of detailed exemplar shapes, and the style of the output detail can be controlled via a style latent code encoding the local style of the corresponding exemplar shape. DECOR-GAN utilizes a 3D CNN generator for upsampling coarse voxels and a 3D PatchGAN discriminator to enforce local patches of the generated shape to be similar to those in the training detailed shapes.
DECOR-GAN is able to refine a coarse shape into a variety of detailed shapes with different styles, as shown in Figure~\ref{fig:voxel_1_superresolution} (d).
Another work, DMTET \cite{shen2021deep}, predicts a deformable tetrahedral grid that carries a signed distance at each grid vertex. Coupled with a differentiable marching tetrahedra\cite{doi1991efficient} layer that converts the signed distances to explicit meshes, it is able to directly output a polygon mesh. It also applies a 3D PatchGAN discriminator on the signed distance field computed from the predicted mesh to improve the local details. DMTET has been discussed in Section \ref{sec:representation_7_grid_mesh}; its method overview can be found in Figure~\ref{fig:7_grid_mesh} (e), and some of its voxel super-resolution results are in Figure~\ref{fig:voxel_1_superresolution} (e).

Another approach to generate new details on the coarse shape is to retrieve high-resolution local patches from a database and combine them into a new shape with respect to the structure and context of the input coarse voxels. RetrievalFuse \cite{siddiqui2021retrievalfuse} creates a shared embedding space between coarse voxel chunks and a database of high-quality voxel chunks from indoor scene data. For a given coarse voxel input, multiple approximate reconstructions are created with retrieved chunks from the database, and the reconstructed scenes are then fused together with an attention-based blending to produce the final reconstruction.

\section{Shape parsing}
\label{sec:voxel_2_parsing}

Parsing an input voxel grid into primitives and (optionally) a sequences of operations requires specific output representations, namely, ``set of primitives'' (Section \ref{sec:representation_5_set_of_primitives}), ``primitive CSG'' (Section \ref{sec:representation_11_CSG}), ``sketch and extrude'' (Section \ref{sec:representation_12_sketch_extrude}), and ``sequence of edits'' (Section \ref{sec:representation_15_sequence_of_edits}). One can refer to those sections for related works. Note that most methods in those sections use global shape latent codes, meaning that the methods can convert any input (point clouds, voxels, images, etc) into parsed shapes, as long as the input can be encoded into global latent codes.

\chapter{Reconstruction from Point Clouds}
\label{sec:pointcloud}

In this section, we review works that reconstruct objects and scenes from point clouds, with or without point normals. We divide the methods into two categories: one based on explicit representations, and the other on implicit representations. Methods with explicit representations can directly output a mesh, but it usually does not guarantee the surface quality, for example, they may not be watertight and may contain no-manifoldness and self-intersections. Methods with implicit representations do guarantee to produce a watertight, manifold mesh without self-intersections, but they require an iso-surfacing algorithm to extract the mesh from the implicit field.

Note that a significant amount of works in 3D deep learning use a global shape latent code to encode the shape, such as ShapeFlow \cite{jiang2020shapeflow}, AtlasNet \cite{groueix2018papier}, OccNet \cite{mescheder2019occupancy}, DeepSDF\cite{park2019deepsdf}, Structured Implicit Functions \cite{genova2019learning}, OctField \cite{tang2021octfield}, CSG-Stump \cite{ren2021csg}, and DeepCAD \cite{wu2021deepcad}. As mentioned in previous sections, global shape latent codes cannot represent detailed shape geometries and they often lead to overfitting on a specific dataset or category of shapes. Therefore, in this section, we will only discuss works that take local features into account.

\section{Explicit representation}
\label{sec:pointcloud_1_explicit}

\begin{figure}[b!]
\begin{center}
\includegraphics[width=1.0\linewidth]{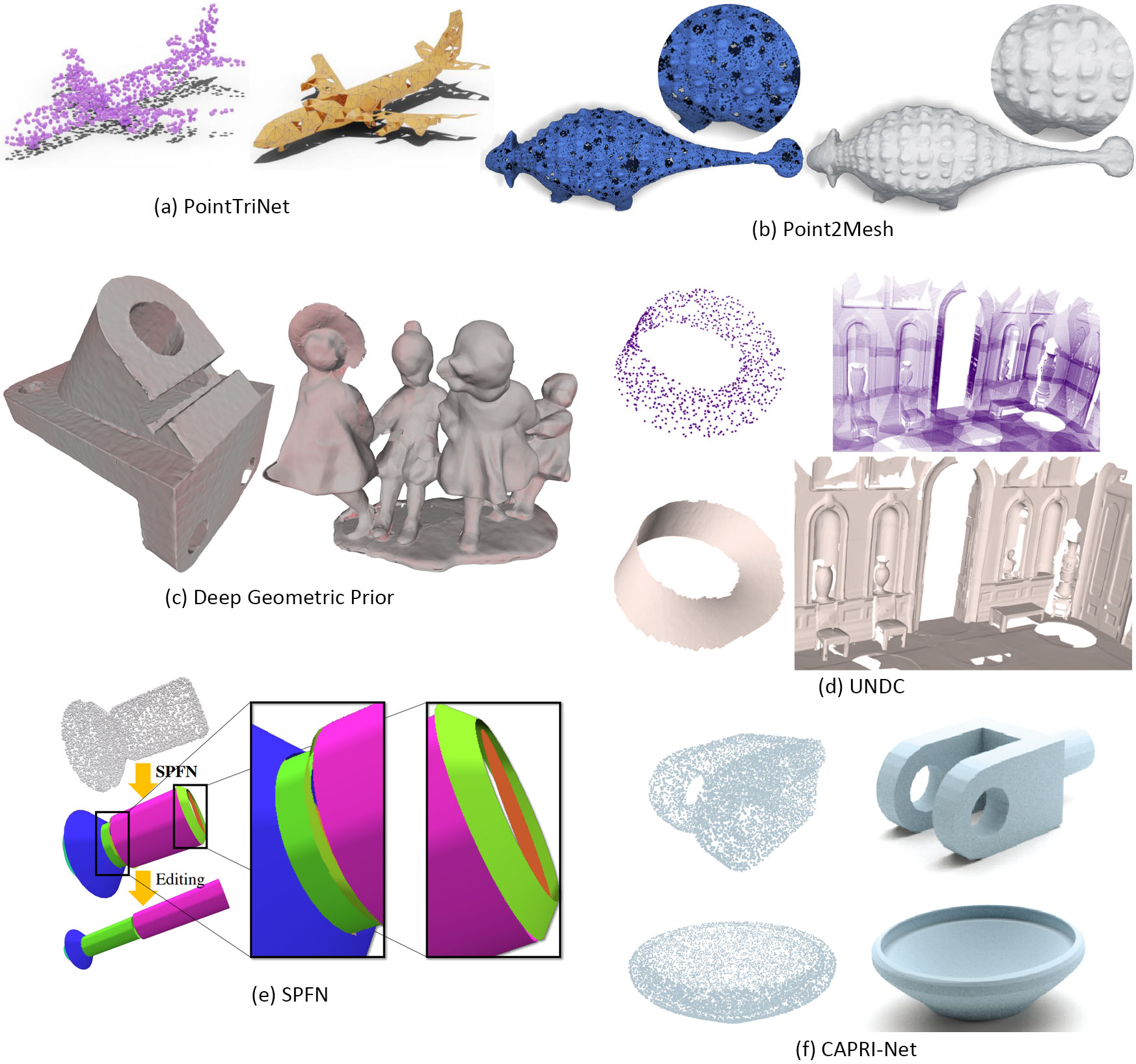}
\end{center}
\caption{
Inputs and results of some point cloud surface reconstruction works using explicit representations. These figures are taken from their corresponding publications (a)\cite{sharp2020pointtrinet}, (b)\cite{hanocka2020point2mesh}, (c)\cite{williams2019deep}, (d)\cite{chen2022ndc}, (e)\cite{li2019supervised}, (f)\cite{yu2022capri}.
}
\label{fig:pointcloud_1_explicit}
\end{figure}

Given a clean and mostly uniform point cloud, a straightforward solution to reconstruct a mesh is to create triangles from the given points. Example classic algorithms include Delaunay triangulation based surface reconstruction methods and ball-pivoting \cite{bernardini1999ball}. Various methods employ deep learning to improve the performance of such approaches. They are detailed in Section \ref{sec:representation_13_connect_V} ``connect given vertices''. An example is shown in Figure~\ref{fig:pointcloud_1_explicit} (a).

If the shape is a single object and has simple topology, it is possible to deform a coarse initial mesh (e.g., from the convex hull of the input point cloud) to fit the point cloud in order to reconstruct a mesh. Specifically, Point2Mesh \cite{hanocka2020point2mesh} deforms the initial mesh to shrink-wrap a single input
point cloud. The initial mesh is represented as a graph and its vertex positions as the predictions of a graph neural network defined on the graph. During training, the initial mesh is deformed by optimizing the weights of the graph neural network. The graph neural network automatically learns certain self-priors that encapsulate reoccurring geometric patterns from the input shape, which encourages local-scale geometric self-similarity across the surface of the reconstructed shape to improve the reconstruction quality, similar to Deep Image Prior \cite{ulyanov2018deep}. An example result is shown in Figure~\ref{fig:pointcloud_1_explicit} (b).

For more complex shapes, one could split the input point cloud into overlapping patches, and fit each patch with a deformable 2D square surface. Deep Geometric Prior \cite{williams2019deep} took inspiration from AtlasNet \cite{groueix2018papier} and Deep Image Prior \cite{ulyanov2018deep}, to overfit each local patch from the input point cloud with a neural network (MLP) deforming a square patch by minimizing their Wasserstein distance. The method applies loss terms and training procedures to encourage consistency between overlapping patches. The reconstructed collection of patches can be further sampled and reconstructed to produce a manifold mesh. The entire procedure does not require any training data or explicit regularization; it relies on the priors reside in the neural networks (MLPs). An example result is shown in Figure~\ref{fig:pointcloud_1_explicit} (c). Meshlet Priors \cite{badki2020meshlet} has a similar idea, with the key difference being that it requires training a Variational Autoencoder (VAE) \cite{VAE} on the square patches (meshlets) to learn a low dimensional manifold of plausible (or ``real'') meshlets, so that the local patches from the input point cloud are fitted with only plausible meshlets when the meshlets are decoded with the VAE from optimizable latent codes. The result meshlets also can be further sampled and reconstructed to produce a manifold mesh. An example result is shown in Figure~\ref{fig:4_deform_multi_primitive} (b).

The Unsigned version of Neural Dual Contouring (UNDC) \cite{chen2022ndc} takes a unique approach. It discretizes the space into a 3D grid, and from the input point cloud it predicts for each grid edge whether the edge will be intersected by the output mesh or not. It also predicts an interior vertex for each grid cell, so that if an edge is predicted to have intersected by the mesh, a quad face will be created to connect the four vertices of the four adjacent cells of that edge; a 2D example is shown in Figure~\ref{fig:7_grid_mesh} (b) right. Therefore, UNDC can directly reconstruct a quad mesh from input points without the need of point normals. Some example results are shown in Figure~\ref{fig:pointcloud_1_explicit} (d).

If one's goal is to produce a more structured output, e.g., a collection of primitives represented as parametric surfaces, then methods introduced in Section \ref{sec:representation_6_primitive_detection} ``primitive detection'' can produce exactly those outputs. Those methods first use a neural network to extract per-point features from the input point clouds, and then apply a clustering module to segment the point cloud into patches belonging to different primitives, and finally classify the primitive type and regress the primitive parameters for each patch. Some results are shown in Figure~\ref{fig:pointcloud_1_explicit} (e).

Methods in \ref{sec:representation_11_CSG} ``primitive CSG'' can also reconstruct structured representations (CSG-trees) from point clouds. Unfortunately most of them use global shape latent codes and cannot generalize. One exception is CAPRI-Net \cite{yu2022capri}. Although CAPRI-Net is pre-trained on a category-specific dataset, it applies a fine-tuning step after training, i.e., input-specific optimization, which allows it to better generalize to new shapes. Some results are shown in Figure~\ref{fig:pointcloud_1_explicit} (f).

\clearpage

\section{Implicit representation}
\label{sec:pointcloud_2_implicit}

The majority of the deep learning methods adopt implicit representations for shape reconstruction from point clouds. However, the boundary between the two representations ``grid voxels'' and ``neural implicit'' has become blurry since a great number of works use 3D CNNs to predict a grid structure and then use interpolation techniques to obtain the features of continuous query points, which will be used to regress the implicit field values of the query points. In the works listed below, only two works Dual OCNN \cite{Wang2022Dual} and RetrievalFuse \cite{siddiqui2021retrievalfuse} use pure occupancy voxel representation, and the rest involve neural implicit more or less.

\subsection{Overfit a single shape}
\label{sec:pointcloud_2_implicit_1}

``Sign Agnostic Learning'' (SAL) \cite{atzmon2020sal}, ``Implicit Geometric Regularization'' (IGR) \cite{gropp2020implicit}, and ``SAL with Derivatives'' (SALD) \cite{atzmon2020sald} all overfit an MLP representing neural implicit from a point cloud without normals. SIREN \cite{sitzmann2020implicit} mainly improves the representation capability of MLPs; it can overfit a neural implicit from a point cloud with normals.

SAL \cite{atzmon2020sal} only assumes the input point cloud is sufficiently dense to produce a good unsigned distance field, so that SAL can learn a neural signed distance field from the unsigned distance field. It proposed an unsigned similarity function, e.g., $(|f(p)|-u(p))^2$, where $f$ is an MLP, $f(p)$ is the predicted signed distance of point $p$ by network $f$, and $u(p)$ is the ground truth unsigned distance of $p$ by computing the shortest distance between $p$ and the input points. What truly makes SAL work is a geometric network initialization, so that the MLP is initialized to approximately represent the signed distance function of sphere centered at the origin. Intuitively, the sphere can shrink-wrap the input points with guidance from the unsigned similarity function.

IGR \cite{gropp2020implicit} proposed to supervise the learned implicit field with three terms. First, an L1 norm on the given input points so that those points should lie on the 0-isosurface. Second, an Eikonal term on any sampled point in the space. The Eikonal term is the squared error between the L2 norm of the point's gradient and one. It encourages any point's gradient in the implicit field to be a unit vector, so that the field is a valid signed distance field. Third, if the normal of an input point is available, then a supervision term is applied to the point, which is the L2 norm of the difference between the point's gradient and the point's ground truth normal.

SALD \cite{atzmon2020sald} improved SAL \cite{atzmon2020sal} by including derivatives - a regression loss between the gradient of a query point in the predicted signed distance field and the gradient in the ground truth unsigned distance field.

SIREN \cite{sitzmann2020implicit} proposed to use periodic activation functions in MLPs for implicit neural representations. It has demonstrated that such modified MLPs are capable of representing complex signals that naive MLPs cannot. Using the three loss terms proposed in IGR \cite{gropp2020implicit}, SIREN can quickly reconstruct a high-quality shape from an oriented point cloud.

Needrop \cite{boulch2021needrop} is not an overfitting method; it uses global shape latent codes to learn to reconstruct shapes as occupancy field from sparse, noisy, and normal-less point clouds. It uses an unique approach inspired by Buffon’s needle problem. Unlike previous approaches where the loss terms are defined on single points, Needrop queries two points at a time, and the loss is defined to encourage the two points to be on the same side of the shape or on different sides, according to where the two points are sampled.

\subsection{Divide space into local cube patches}
\label{sec:pointcloud_2_implicit_2}

``Local Implicit Grid'' (LIG) \cite{jiang2020local} and ``Deep Local Shapes'' (DeepLS) \cite{chabra2020deep} are conceptually very similar. They both realized that global shape latent codes in early neural implicit methods are not generalizable, therefore they divide the space into a grid of overlapping cube patches, and fit an MLP for each patch. The MLP is pre-trained on a collection of shape patches to learn a low dimensional latent space of plausible (or ``real'') shape patches. During inference, the MLP is fixed, but a latent code is optimized independently for each patch of the input point cloud to match the MLP's isosurface to the input points in that patch, therefore reconstructing a plausible patch according to the input points. They both require point normals to help this optimization. Finally, the reconstructed patches are stitched together to give the reconstructed shape or scene.

SAIL-S3 \cite{zhao2021sign} can reconstruct a shape from point clouds without normals, like SAL \cite{atzmon2020sal}. It also divides the space into a grid of overlapping cube patches and optimizes the latent code of an MLP to fit the isosurface with respect to the input points in each patch, but it does not need a pre-trained MLP. It learns the implicit field of all the patches at the same time to leverage the surface self-similarities by improving correlations among the optimized latent codes of those patches. It can reconstruct more complex shapes than SAL \cite{atzmon2020sal} thanks to the fitting of local patches rather than an entire shape.

RetrievalFuse \cite{siddiqui2021retrievalfuse} creates a shared embedding space between patches of sparse point clouds and a database of high-quality voxel chunks from indoor scene data. For a given point cloud input, multiple approximate reconstructions are created with retrieved chunks from the database, and the reconstructed scenes are then fused together with an attention-based blending to produce the final reconstruction. It does not require ground truth point normals as input.

\subsection{3D CNN then local neural implicit}
\label{sec:pointcloud_2_implicit_3}

ConvONet \cite{peng2020convolutional} and IF-Nets \cite{chibane2020implicit} are also covered in Section \ref{sec:voxel_1_superresolution} for voxel super-resolution. They can also reconstruct shapes from point clouds without normals.

ConvONet \cite{peng2020convolutional} first uses a point cloud encoder to process the input points, and the per-point features are pooled into grids to be processed by CNNs for further feature extraction. The grids can be a 3D grid of deep features (grid setting), or three 2D grids of deep features on three orthogonal planes (tri-plane setting), or a combination of both. Then for each query point, the deep features are retrieved from the grid or the planes via trilinear or bilinear interpolation, and the deep features are concatenated with the query point coordinates to be fed into an MLP to predict the inside/outside status of the query point.

IF-Nets \cite{chibane2020implicit} adopts CNN encoders on an input point cloud (discretized into a 3D grid) to compute multi-scale 3D grids encoding global and local features. Then for each query point, the deep features are retrieved from those grids via trilinear interpolation, and then the deep features (without query point coordinates) are fed into an MLP to predict the inside/outside status of the query point.

SA-ConvONet \cite{tang2021sa} is a follow-up of SAL \cite{atzmon2020sal} and ConvONet \cite{peng2020convolutional}. Its idea is to apply sign agnostic learning proposed in SAL as a post-processing step to improve the reconstruction quality of ConvONet. It does not require ground truth point normals as input.

GIFS \cite{ye2022gifs} has the same backbone network with IF-Nets \cite{chibane2020implicit}. However, it does not predict the inside/outside status of each query point, but rather predict whether two query points are separated by any surface. The concept is similar to Unsigned Neural Dual Contouring (UNDC) \cite{chen2022ndc}, but UNDC's meshing is based on Dual Contouring \cite{ju2002dual}, and this work on a modified version of Marching Cubes \cite{lorensen1987marching}.
It can represent general shapes including non-watertight shapes and shapes with multi-layer surfaces. It does not require ground truth point normals as input.

\subsection{Point cloud encoder then local neural implicit}
\label{sec:pointcloud_2_implicit_4}

Points2Surf \cite{erler2020points2surf} can reconstruct a shape from a noisy, normal-less point cloud. It is purely based on point cloud encoders without any CNN. For an query point in space, it adopts one PointNet \cite{qi2017pointnet} to encode points sampled at the neighborhood of the query point into a local feature code, and another PointNet to encode the points sampled at the entire input point cloud into a global feature code. The decoder takes the query point and the local and global features as input and outputs the sign probability and unsigned distance of the query point, which can be combined into the signed distance of the query point.

POCO \cite{boulch2022poco} claims that methods in the previous subsections \ref{sec:pointcloud_2_implicit_2} and \ref{sec:pointcloud_2_implicit_3} only infer latent vectors on a coarse regular 3D grid, therefore they lose the direct connection with the input points sampled on the surface of objects, and they attach information uniformly in space rather than where it matters the most, i.e., near the surface. To address these issues, POCO proposed to use point cloud convolutions and compute latent vectors at each input point. Then for each query point, it performs a learning-based interpolation on nearest neighbors in input points to retrieve an weighted-averaged feature vector, and the feature vector is processed by an MLP to predict the occupancy of the query point. It does not require ground truth point normals as input.

\subsection{Implicit field defined by points}
\label{sec:pointcloud_2_implicit_5}

The methods in this sections use points (either input points or predicted points) and their properties (such as normals) to directly compute the implicit field value of any query point, similar to classic Radial Basis Function (RBF) surface reconstruction methods. The inference from these points to an implicit field does not involve any deep learning or neural networks. Therefore, the backbone networks that produce these points are the trainable parts in those methods, and they possess learned priors from the training datasets.

Neural Splines \cite{williams2021neural} does not involve any neural networks. It can reconstruct an implicit field from a set of points and their normals, It is a kernel method for surface reconstruction based on kernels arising from infinitely-wide shallow ReLU networks, that is, the reconstructed implicit field is the same as the fitted result of an infinitely-wide, one hidden layer, ReLU network, when the first layer weights are fixed at random values according to some specified distribution, and when it is supervised to have its 0-isosurface cross all input points and its gradients at input points agree with their normals. It shows that the inductive bias in MLPs favors smoothness and it can lead to smooth reconstructed shape surfaces. This is a very theoretical paper and its idea cannot be explained in simple words. Please take a look if interested.

``Neural Kernel Fields'' (NKF) \cite{williams2022neural} is a follow-up of Neural Splines \cite{williams2021neural}. It proposed to replace the fixed point properties (normals) with a learned feature vector, so as to have data dependent kernels. Those features are predicted by a network similar to the backbone of ConvONet \cite{peng2020convolutional}, taking the points and their normals as input. It still uses the same formulation in Neural Splines to process those point properties into a implicit field. It has showed better reconstruction quality than Neural Splines, and it is able to perform shape completion, which Neural Splines cannot do because it does not involve learning.

``Shape As Points'' (SAP) \cite{peng2021shape} proposed a differentiable point-to-mesh layer using a differentiable formulation of Poisson Surface Reconstruction (PSR) \cite{kazhdan2006poisson,kazhdan2013screened}, so that a shape can be represented as a set of points with normals. SAP uses the backbone network of ConvONet \cite{peng2020convolutional} to process an input noisy un-oriented point cloud and output a clean oriented point cloud by predicting the offset and normal for each input point. Afterwards, the shape can be reconstructed from the clean oriented point cloud by Poisson Surface Reconstruction. Note that the training loss is not defined on predicted points and normals, but on the occupancy values on query points like other neural implicit works, because the differentiable Poisson Surface Reconstruction layer will propagate the error of the occupancy to the points and normals and then to the weights of the backbone network.

``Deep Implicit Moving Least-Squares'' (Deep IMLS) \cite{liu2021deep} takes a sparse and un-oriented point cloud as input, and use a U-Net-like O-CNN autoencoder \cite{wang2017cnn} to predict an octree structure where each octree node contains a fixed number of predicted points with normals. Those predicted points with normals are then used to construct an implicit field by implicit moving least-squares (IMLS) surface formulation \cite{kolluri2008provably}. Similar to SAP \cite{peng2021shape}, IMLS is also differentiable, so the reconstruction loss is defined on implicit field values, specifically, on signed distances.

\subsection{Octrees}
\label{sec:pointcloud_2_implicit_6}

The $O(N^3)$ space complexity of regular 3D grids makes methods based on regular grids hard to scale up. Therefore, some methods have been proposed to include adaptive spatial structures such as octrees in the neural networks to improve both efficiency and quality. 

Deep IMLS \cite{liu2021deep} in the previous subsection uses an octree structure to predict points and their normals in each octree node for reconstructing an implicit field.

AdaConv \cite{ummenhofer2021adaptive} proposed multiscale convolutional kernels that can be applied to adaptive grids as generated with octrees. The kernels span multiple resolutions, which allows applying convolutional networks to adaptive grids for large problem sizes where the input data is sparse but the entire domain needs to be processed. It takes a point cloud with normals as input, and then build the octree and process the input using the proposed convolutional kernels. Finally, it uses an MLP to decode the signed distance field and the unsigned distance field for each octree node, and uses adaptive Dual Contouring \cite{ju2002dual} to extract the surface.

Dual OCNN \cite{Wang2022Dual} designed graph convolutions over the dual graph of octree nodes. The graph convolution operator is defined over a regular grid of features fused from irregular neighboring octree nodes at different levels, which reduces the computational and memory cost of the convolutions over irregular neighboring octree nodes, and improves the performance of feature learning. It takes a point cloud with or without normals as input, and then builds the octree and processes the input using the proposed convolutional kernels. It directly outputs a voxel grid of size up to $256^3$, which can be used to extract a mesh.

\begin{figure}[b!]
\begin{center}
\includegraphics[width=1.0\linewidth]{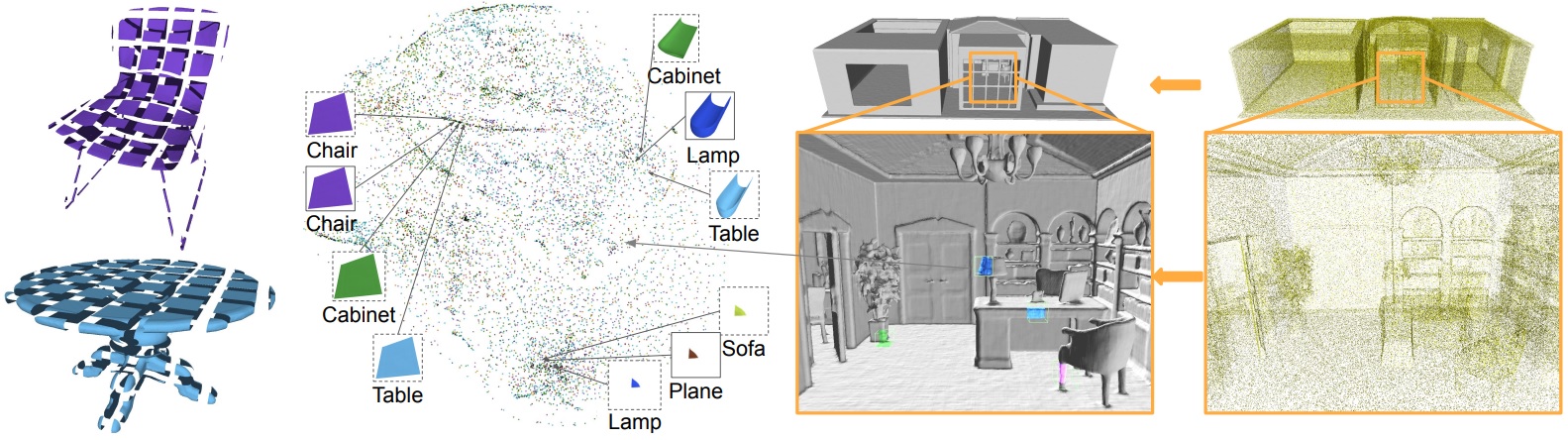}
\end{center}
\caption{
This figure is taken from LIG \cite{jiang2020local}, which divides the space into patches and fit each patch with a neural implicit. Unlike the results of works with explicit representations, the results of works with implicit representations all look alike, so one representative figure should be enough. The methods of these works are very different though.
}
\label{fig:pointcloud_2_implicit}
\end{figure}

\clearpage

\chapter{Reconstruction from Single Images}
\label{sec:singleimage}

Methods that reconstruct a shape from a single image can be divided into two categories based on the supervision they receive during training. One category is trained with ground truth 3D shapes as supervision. The methods in this category are typically trained on ShapeNet \cite{chang2015shapenet}. Another category is trained with only single-view images as supervision. Single-view images mean that there is only one image for each object for training, in contrast to multi-view images where each object has multiple images from different viewpoints. The methods in this category typically train on image datasets of birds, cars, horses, and faces, with shapes of sphere or disk topology.

\section{With 3D supervision}
\label{sec:singleimage_1_3Dsup}

\textbf{Global features.}
Similar to Section \ref{sec:pointcloud}, a significant amount of works in 3D deep learning use a global shape latent code to encode the shape, such as SurfNet \cite{sinha2017surfnet}, 3D-R2N2 \cite{choy20163d}, OGN \cite{tatarchenko2017octree}, HSP \cite{hane2017hierarchical}, ShapeHD \cite{wu2018learning}, AtlasNet \cite{groueix2018papier}, Im2Struct \cite{niu2018im2struct}, Matryoshka Networks \cite{richter2018matryoshka}, SkeletonNet \cite{tang2019skeleton}, IM-Net \cite{chen2019imnet}, OccNet \cite{mescheder2019occupancy}, Deep Level Sets \cite{michalkiewicz2019deep}, Deep Meta Functionals \cite{littwin2019deep}, topology-modifying AtlasNet \cite{pan2019deep}, Pix2Vox \cite{xie2019pix2vox}, PQ-NET \cite{wu2020pq}, BSP-Net \cite{chen2020bsp}, Cvxnet \cite{deng2020cvxnet}, LDIF \cite{genova2020local}, Neural Template \cite{hui2022neural}, AutoSDF \cite{mittal2022autosdf}.
Global shape latent codes cannot represent detailed shape geometries, and they often lead to learning shape recognition rather than shape reconstruction, as pointed out by ``What do single-view 3d reconstruction networks learn?'' \cite{tatarchenko2019single} in 2019. That is, an encoder-decoder structured neural network with a global shape latent code is likely to simply memorize the shapes in the training set during training, and ``retrieve'' a shape from the memory bank as output during testing. Therefore, in the following, we will only discuss works that take local features into account.

\textbf{Deforming a sphere mesh with local image features.}
Pixel2Mesh \cite{wang2018pixel2mesh} progressively deforms and subdivides a sphere mesh via graph convolutional networks. It extracts image features with a VGG-16 CNN, and then pools image features into the vertices of the mesh to enrich the vertex features, so that the graph convolutional networks can learn local-feature-aware deformations. Its method overview can be found in Figure~\ref{fig:3_deform_one_primitive} (a). 
Geometric Granularity Aware Pixel2Mesh \cite{shi2021geometric} is a follow-up of Pixel2Mesh \cite{wang2018pixel2mesh} and it can edit the topology of the mesh like \cite{pan2019deep} by utilizing an error estimator nerwork to identify faces to prone or repair. It applies a keypoint detector to detect keypoints from the ground truth 3D mesh, and then uses the keypoints to regularize the deformed mesh.

\begin{figure}[b!]
\begin{center}
\includegraphics[width=1.0\linewidth]{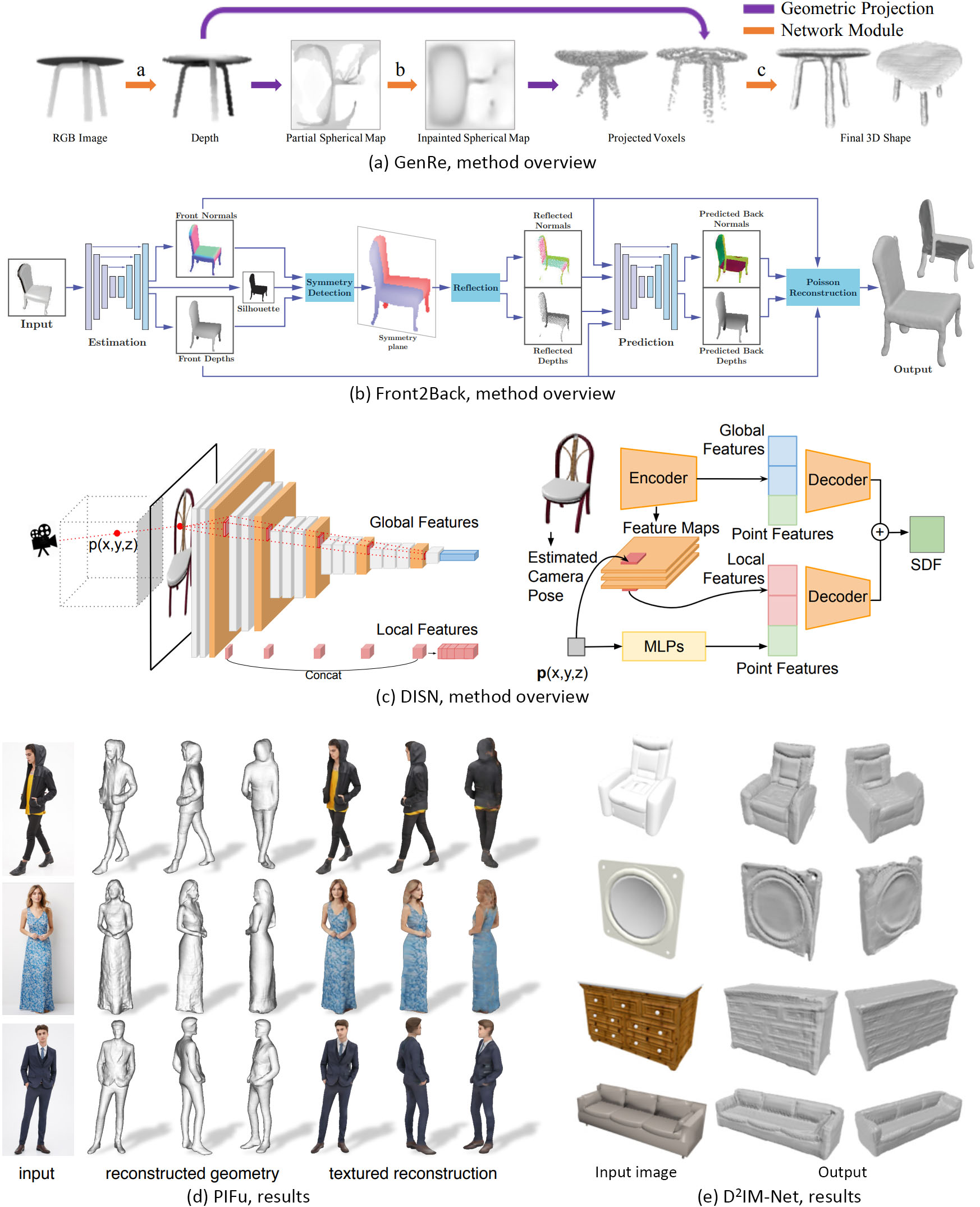}
\end{center}
\vspace{-8mm}
\caption{
Inputs and results of some single view 3D reconstruction methods trained with 3D supervision. These figures are taken from their corresponding publications (a)\cite{zhang2018learning}, (b)\cite{yao2020front2back}, (c)\cite{xu2019disn}, (d)\cite{saito2019pifu}, (e)\cite{li2021d2im}.
}
\label{fig:singleimage_1_3Dsup}
\end{figure}

\textbf{Depth estimation, shape completion in image and voxel space.}
``Learning to Reconstruct Shapes from Unseen Classes'' (GenRe) \cite{zhang2018learning} first uses a 2D CNN to predict a depth map in the original view from a single RGB image. Then it projects the depth map into a partial spherical map. A spherical inpainting network is designed to inpaint the partial spherical map into a complete partial spherical map. Then the complete partial spherical map and the previous depth map are projected into voxels to be processed by a voxel refinement network to produce the final output. Its method overview can be found in Figure~\ref{fig:singleimage_1_3Dsup} (a). The experiments show that the method can reconstruct objects from categories not seen during training.

\textbf{Front\&back Normal and Depth estimation, symmetry detection.}
Front2Back \cite{yao2020front2back} first predicts depth, normal, and silhouette maps from the input image. They represent the surface of the object in front of the camera. It then detects global reflective symmetries from these maps, and reflect the front depth and normal maps to create partial back depth and normal maps. The partial back depth and normal maps are fed into a network to predict the complete back depth and normal maps. Finally, a 3D mesh can be reconstructed from the Front\&back normal and depth maps using Screened Poisson \cite{kazhdan2013screened}. Its method overview can be found in Figure~\ref{fig:singleimage_1_3Dsup} (b).

\textbf{Neural implicit with local image features.} DISN \cite{xu2019disn} and PIFu \cite{saito2019pifu} are the two pioneers to first incorporate local features from the input image for single-view 3D reconstruction with neural implicit representation. DISN \cite{xu2019disn} learns image feature maps with a VGG-16 CNN, and uses the projected location for each 3D query point on the 2D image and extracts local feature from the image feature maps. The CNN also produces a global feature. The global and local features are being processed by two separate MLPs to produce two signed distances, and their sum is the final output signed distance of the query point. DISN is able to captures details such as holes and thin structures present in 3D shapes from single-view images. Its method overview can be found in Figure~\ref{fig:singleimage_1_3Dsup} (c). PIFu (``pixel-aligned implicit function'') \cite{saito2019pifu} is a concurrent work and has very similar idea with DISN. It is designed to reconstruct a high-resolution 3D textured surface of clothed humans from a single input image. It uses a ResNet \cite{he2016deep} image encoder to learn image feature maps. For a 3D query point, it projects the point on the 2D image and extracts local features from the image feature maps. The features are fed into an MLP to predict the inside-outside status of the query point. It also uses an almost identical network to predict the RGB color of each query point, forming the texture of the output mesh. PIFu can reconstruct details in clothing (e.g. wrinkles), complex hairstyles, and invisible regions such as the back of the subject. Its example results can be found in Figure~\ref{fig:singleimage_1_3Dsup} (d)

\textbf{Neural implicit with front\&back normal, coarse\&fine levels.}
PIFuHD \cite{saito2020pifuhd} is a follow-up of PIFu \cite{saito2019pifu}. It consists of a coarse PIFu nerwork and a fine PIFu network. The coarse PIFu is almost identical to PIFu, whose feature maps capture both global and local features. The fine PIFu's image encoder is designed with limited receptive field to focus only on local details. coarse PIFu also predicts the front\&back normal maps from the input image, and these maps are used as additional input to the fine PIFu network. For a 3D query point, PIFuHD extracts local feature from the fine PIFu and global+local features from the coarse PIFu, and use them to predict the inside-outside and RGB color of the query point.

\textbf{Neural implicit with feature fusion based on symmetry.}
Ladybird \cite{xu2020ladybird} does not depend on symmetry detection, but assumes the shapes are normalized to be symmetric about the xy plane crossing the origin. It also uses an image encoder to learn image feature maps. For a query 3D point, both the point and its mirrored point with respect to the symmetry plane are projected on the 2D image to extract local features. The two feature vectors are concatenated to be fed into an MLP to regress the signed distance.
 
\textbf{Neural implicit with front\&back displacement maps.}
D$^2$IM-Net \cite{li2021d2im} trains the network to learn a detail disentangled reconstruction consisting of a 3D implicit field representing the coarse 3D shape, and two 2D displacement maps capturing the details of the object. It encodes an input image into a global feature code and a 2D local feature grid. The global feature is fed into an MLP with 3D query point coordinates to produce the signed distance field of a coarse shape. The local feature grid is fed into a CNN to predict the displacement maps of the front\&back of the object with respect to the camera. The final 3D reconstruction is a fusion between the coarse shape and the displacement maps. In addition to the SDF reconstruction loss, the work proposed a novel Laplacian term on the displacement maps to enforce the recovery of surface details. Its example results can be found in Figure~\ref{fig:singleimage_1_3Dsup} (e)

\section{With 2D supervision}
\label{sec:singleimage_2_2Dsup}

Methods often employ a global shape latent code for this task since the task is very difficult and the methods must learn category-specific priors. Most methods adopt a spherical mesh as template and deform it into the target shape, often with textures. This representation and some related works are covered in Section \ref{sec:representation_3_deform_one_primitive} ``deform one primitive''. Methods that use neural implicit representation for this task often require ground-truth camera pose for each image and multi-view images of the same object. Table 1 in ``Share With Thy Neighbors: Single-View Reconstruction by Cross-Instance Consistency'' (UNICORN) \cite{monnier2022unicorn} and Table 1 in ``2D GANs Meet Unsupervised Single-view 3D Reconstruction'' (GANSVR) \cite{liu20222d} are great summaries of recent works on this topic. We recommend interested readers to take a look.

\chapter{Reconstruction from Multi-View Images}
\label{sec:multiview}

Only a few works follow the path of single image reconstruction methods in Section \ref{sec:singleimage} to learn priors from a collection of training shapes. They can aggregate the global shape latent codes from multiple input images using recurrent neural networks as in 3D-R2N2 \cite{choy20163d}, aggregate spatial features decoded from global shape latent codes as in Pix2Vox \cite{xie2019pix2vox}, aggregate image features from multiple input images as in Pixel2Mesh++ \cite{wen2019pixel2mesh++}, or aggregate image features from multiple input images and 3D embeddings of spatial locations using a Transformer as in EVolT \cite{wang2021multi}.

Most methods are overfitting a single shape or scene with respect to the multiple input images using methods based on the differentiable rendering algorithms on meshes or neural implicit, or based on the ray marching volume rendering formula of NeRF \cite{mildenhall2021nerf}.

\section{Differentiable rendering on explicit representation}
\label{sec:multiview_1_explicit}

Among the overfitting methods, only a few of them adopt explicit mesh representations. NeRS \cite{zhang2021ners}, DS \cite{goel2022differentiable}, and NDS \cite{worchel2022multi} use the ``deform one primitive'' representation to deform a sphere mesh into the target shape. DEFTET \cite{gao2020learning} and Nvdiffrec \cite{munkberg2022extracting} use the ``grid mesh'' representation. MobileNeRF \cite{chen2022mobilenerf} uses the ``grid polygon soup'' representation. NeRS, DS, NDS, and Nvdiffrec have a differentiable rendering formula that assumes the object segmentation mask is given for each input image. DEFTET \cite{gao2020learning} and MobileNeRF \cite{chen2022mobilenerf} do not need object segmentation masks. Following differentiable mesh rendering algorithms such as Soft Rasterizer \cite{liu2019soft}, for each pixel, the methods record all or first $k$ intersected points between the mesh and the ray from the pixel, and aggregate the colors via alpha-compositing during training.

NeRS \cite{zhang2021ners} is designed to learn the geometry of the target object as a deformable sphere mesh, and the (neural) bidirectional surface reflectance functions (BRDFs) that factorize view-dependent appearance into environmental illumination, diffuse color (albedo), and specular ``shininess''. It is supervised by a collection of multi-view images of the target object, and the collection may contain only a small number of views with unknown/rough camera estimates. Its example results can be found in Figure~\ref{fig:multiview} (a).

Differentiable Stereopsis (DS) \cite{goel2022differentiable} took inspiration from stereopsis that two images of the same scene which appear different become similar after projection onto an approximate 3D model of the scene. It proposes a differentiable texture transfer method by texture warping via 3D projections and unprojections. Therefore, unlike other methods in this section, it does not predict mesh textures, but the final mesh texture can be obtained by aggregating textures in the input images. It can also change the topology of the mesh template during training, by voxelizing the mesh, projecting voxels onto the view planes, checking for occupancy by comparing to the ground truth silhouettes, removing voxels that project to an unoccupied area in any mask, and re-meshing the remaining voxels using marching cubes. Its example results can be found in Figure~\ref{fig:multiview} (b).

Neural Deferred Shading (NDS) \cite{worchel2022multi} proposes a differentiable renderer to utilize a neural shader on the optimized mesh. Instead of predicting a 2D texture and relying on UV-mapping in differentiable rendering to produce colors, the method only rasterizes the mesh into per-pixel point location (from depth) and normals. And an MLP neural shader is designed to take point coordinates, normals, and a viewing direction as input and output RGB color for the corresponding pixel. The method's initial mesh is computed as a visual hull of the object segmentation masks, which is different from NeRS \cite{zhang2021ners} (cube initial mesh) and DS \cite{goel2022differentiable} (sphere initial mesh).

DEFTET \cite{gao2020learning} predicts a tetrahedron grid. To represent the geometry, the method predicts the occupancy for each tetrahedron, and the offset for each vertex relative to their initial positions in the regular tetrahedron grid. To represent the color and transparency, the method predicts an RGB attribute and a visibility attribute for each mesh vertex. Those attributes are used in alpha-compositing to produce the pixel color during differentiable rendering.

Nvdiffrec \cite{munkberg2022extracting} combines DMTET \cite{shen2021deep} and differentiable rendering to reconstruct meshes from multi-view images. DMTET predicts signed distance on each grid vertex in a deformable tetrahedron grid, and applies a differentiable marching tetrahedra layer to extract a triangle mesh. Nvdiffrec applies differentiable rendering on the extracted mesh of DMTET to perform mesh reconstruction from multi-view images. Nvdiffrec in addition recovers spatially-varying materials and environment map lighting from the input images, therefore the output mesh and materials can be deployed in traditional graphics engines unmodified. Its example results can be found in Figure~\ref{fig:multiview} (c).

MobileNeRF \cite{chen2022mobilenerf} can reconstruct a polygon soup from multi-view images. It uses a regular grid mesh as the initial mesh, and optimizes its vertex positions and face occupancies during training. To represent the color and transparency of the mesh, it uses MLPs to predict the view-dependent color and the view-independent opacity of each 3D point in space. The output mesh of MobileNeRF is coarse and has poor quality, as shown in Figure~\ref{fig:8_grid_polygon_soup} (c). Therefore, the output mesh must be coupled with the texture maps extracted from the MLPs to produce high-quality renderings, as shown in Figure~\ref{fig:8_grid_polygon_soup} (d).

\section{Surface rendering on implicit representation}
\label{sec:multiview_2_implicit}

The methods in this section all have a differentiable rendering formula that assumes the object segmentation mask is given for each input image, and each ray intersects the surface at most once (only one intersection point per ray for the gradient to propagate).

\begin{figure}[b!]
\begin{center}
\includegraphics[width=1.0\linewidth]{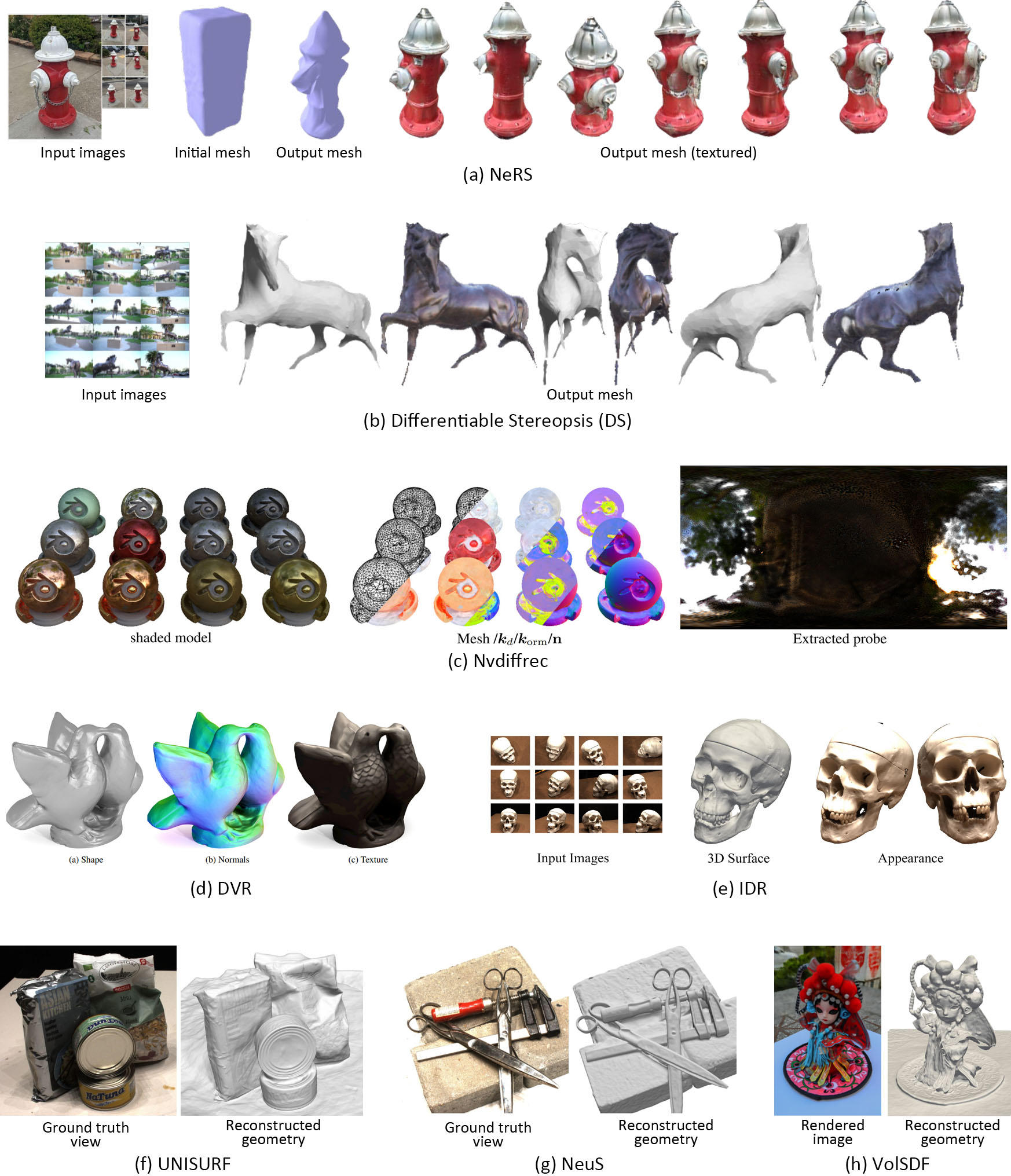}
\end{center}
\caption{
Inputs and results of some methods that reconstruct surfaces from multi-view images. These figures are taken from their corresponding publications (a)\cite{zhang2021ners}, (b)\cite{goel2022differentiable}, (c)\cite{munkberg2022extracting}, (d)\cite{niemeyer2020differentiable}, (e)\cite{yariv2020multiview}, (f)\cite{oechsle2021unisurf}, (g)\cite{wang2021neus}, (g)\cite{yariv2021volume}.
}
\label{fig:multiview}
\end{figure}

SDFDiff \cite{jiang2020sdfdiff}, DVR \cite{niemeyer2020differentiable}, and IDR \cite{yariv2020multiview} all propose different formulations of differentiable rendering on implicit surfaces, while SDFDiff \cite{jiang2020sdfdiff} uses regular grid SDF and others use neural implicit. To model colors, SDFDiff assumes the target shape does not have textures and it does not predict textures for the reconstructed shape; DVR adopts Texture fields \cite{oechsle2019texture}, using an MLP to predict the RGB color of each surface point, thus it cannot model view-dependent effects; IDR uses an MLP to approximate the bidirectional reflectance distribution function (BRDF) of each surface point. Some example results can be found in Figure~\ref{fig:multiview} (d)(e).

Neural Lumigraph Rendering (NLR) \cite{kellnhofer2021neural} improves IDR \cite{yariv2020multiview} by replacing its ReLU MLP backbone with SIREN \cite{sitzmann2020implicit}. It also shows that the extracted mesh from the learned implicit field can be combined with unstructured lumigraph rendering \cite{buehler2001unstructured} to achieve real-time rendering.

MVSDF \cite{zhang2021learning} leverages stereo matching and feature consistency to optimize the implicit surface representation. It applies a signed distance field and a surface light field to represent the scene geometry and appearance respectively. The SDF is directly supervised by geometry from stereo matching, and is refined by optimizing the multi-view feature consistency and the fidelity of rendered images. The method was shown to work without object segmentation masks.

RegSDF \cite{zhang2022critical} uses a pretrained network to reconstruct an oriented point cloud from the input images, and then use the oriented point cloud to supervise and regularize the learning of the neural SDF field within a differentiable rendering framework. The method was shown to work without object segmentation masks.

Reparameterization SDF renderer \cite{bangaru2022differentiable} presents a method to compute correct gradients with respect to network parameters in neural SDF renderers. It observes that prior differentiable renderers cannot handle the discontinuities created by occlusion or object silhouettes, therefore it proposes a continuous warping function for SDF to address the issue. The method was shown to work without object segmentation masks.

\section{Volume rendering on implicit representation}
\label{sec:multiview_3_nerf}

The methods in this section adopt a NeRF-style ray marching volume rendering algorithm. For each pixel, the camera shoots a ray crossing it. A number of points are sampled along the ray. Each sampled point carries density (``opacity'') and radiance (view-dependent RGB color), predicted by an MLP. The final pixel color is the accumulated radiance of all the sampled points with respect to their density, similar to alpha-compositing. Those methods usually do not need object segmentation masks, and they somehow represent the point density with well-defined neural implicit fields, so that the surface of the shape can be extracted via iso-surfacing.

UNISURF \cite{oechsle2021unisurf}, NeuS \cite{wang2021neus}, and VolSDF \cite{yariv2021volume} all propose different formulations of volume rendering on implicit surfaces. Their example results can be found in Figure~\ref{fig:multiview} (f)(g)(h).

UNISURF \cite{oechsle2021unisurf} formulates implicit occupancy field models and radiance fields in a unified way, enabling both surface and volume rendering using the same model. This unified perspective enables more efficient sampling procedures to sample point in an interval around the surface.

NeuS \cite{wang2021neus} proposes to represent a surface as the zero-level set of a signed distance function. It introduced a new volume rendering formulation on SDF to be free of bias (inherent geometric errors) in the first order of approximation, leading to more accurate surface reconstruction.

VolSDF \cite{yariv2021volume} represents density as transformed signed distance. In more detail, it defines the volume density function as Laplace’s cumulative distribution function (CDF) applied to a signed distance function (SDF). It also proposes a new adaptive sampling scheme to concentrate the sampled points near the shape surface.

``Neural RGB-D Surface Reconstruction'' \cite{azinovic2022neural} targets indoor scene reconstruction. It proposes an RGB-D based scene reconstruction method, to incorporate depth measurements into the optimization of a NeRF using a truncated signed distance function based surface representation to store the scene geometry.

NeuralWarp \cite{darmon2022improving} follows VolSDF \cite{yariv2021volume} and proposes to add a direct photo-consistency term across the different views during optimization, so that it warps views on each other in a consistent way to ensure the correctness of the implicit geometry.

ManhattanSDF \cite{guo2022neural} targets indoor scene reconstruction and follows VolSDF \cite{yariv2021volume}. It observed that prior works have difficulty in handling low-textured planar regions common in indoor scenes. Therefore to solve this issue, it incorporates planer constraints into implicit neural SDF based reconstruction methods. Based on the Manhattan-world assumption, planar constraints are employed to regularize the geometry in floor and wall regions predicted by a 2D semantic segmentation network.

Geo-Neus \cite{fugeo} follows NeuS \cite{wang2021neus}. It analyzed in theory that there exists a gap between the volume rendering integral and point-based signed distance function modeling. To bridge this gap, it directly locates the zero-level set of SDF networks and explicitly perform multi-view geometry optimization by leveraging the sparse geometry from structure from motion (SFM) and photometric consistency in multi-view stereo. This makes the SDF optimization unbiased and allows the multi-view geometry constraints to focus on the true surface optimization.

``Neural 3D Reconstruction in the Wild'' \cite{sun2022neural} follows NeuS \cite{wang2021neus} and ``NeRF in the Wild'' (NeRF-W) \cite{martin2021nerf} to reconstruct scenes from Internet photo collections in the presence of varying illumination. It proposes a hybrid voxel- and surface-guided adaptive sampling technique that allows for more efficient ray sampling around surfaces, which guides the network to explain the rendered color with near-surface
samples, leading to more accurate geometric fitting.

SNeS \cite{insafutdinov2022snes} follows NeuS \cite{wang2021neus}. It targets 3D reconstruction of partly-symmetric objects, by factoring the apparent color into material and lighting properties, and applying a soft symmetry constraint to the 3D geometry and material properties.

SparseNeuS \cite{long2022sparseneus} follows NeuS \cite{wang2021neus}, and it targets 3D reconstruction from sparse images. It learns generalizable priors across scenes by introducing geometry encoding volumes for generic surface prediction. It also proposes several strategies to effectively leverage sparse views for high-quality reconstruction.

MonoSDF \cite{yumonosdf} follows VolSDF \cite{yariv2021volume}, and it targets complex scene reconstruction and reconstruction from sparse images. it utilizes the depth and normal maps predicted by pretrained general-purpose monocular estimator networks for 2D images to improve reconstruction quality and optimization time.

HF-NeuS \cite{wanghf} follows NeuS \cite{wang2021neus} and improves NeuS's quality, especially on high-frequency details. It finds that attempting to jointly encode high-frequency and low-frequency components in a single SDF leads to unstable optimization, therefore it proposes to decompose the SDF into a base function and a displacement function with a coarse-to-fine strategy to gradually increase the high-frequency details.

\chapter{Conclusion and Future work}
\label{sec:conclusion}

In this survey, we reviewed different shape representations, and various deep learning 3D reconstruction methods that reconstruct surfaces from voxels, point clouds, single images, and multi-view images.

Although this survey is intended to cover methods that can produce explicit meshes, half of the cited works use implicit representations, and the state-of-the-art are always those who adopt implicit representations, except when the task is related to shape parsing. Therefore, it is worthwhile to investigate the fundamental issues in learning explicit representations, and/or bridge the gap between implicit representations and explicit representations.

Large image and language models have shown that generalizability can be achieved by simply having more training data. But training data is indeed very limited in 3D domain. For example, the majority of 3D-supervised methods use ShapeNet \cite{chang2015shapenet}, and there are a total of 6,778 chairs in ShapeNet. How can anyone expect a method trained on merely 6,778 chairs to generalize to real-life chairs which consist of millions of different structures, styles, and textures? Therefore, multiple new directions lie ahead, for example, effective data augmentation, using real 2D images to facilitate 3D learning, pre-training on synthetic datasets, synthetic-to-real domain adaptation, semi-automatic 3D data collection and labeling, and of course, someone needs to do the dirty work of creating large-scale 3D datasets.


\backmatter%
    \addtoToC{Bibliography}
    \bibliographystyle{plain}
    \bibliography{references}

\begin{thebibliography}{100}

\bibitem{sketchfab}
Sketchfab.
\newblock https://sketchfab.com/.

\bibitem{atzmon2020sal}
Matan Atzmon and Yaron Lipman.
\newblock Sal: Sign agnostic learning of shapes from raw data.
\newblock In {\em Proceedings of the IEEE/CVF Conference on Computer Vision and
  Pattern Recognition}, pages 2565--2574, 2020.

\bibitem{atzmon2020sald}
Matan Atzmon and Yaron Lipman.
\newblock Sald: Sign agnostic learning with derivatives.
\newblock In {\em International Conference on Learning Representations}, 2020.

\bibitem{azinovic2022neural}
Dejan Azinovi{\'c}, Ricardo Martin-Brualla, Dan~B Goldman, Matthias
  Nie{\ss}ner, and Justus Thies.
\newblock Neural rgb-d surface reconstruction.
\newblock In {\em Proceedings of the IEEE/CVF Conference on Computer Vision and
  Pattern Recognition}, pages 6290--6301, 2022.

\bibitem{badki2020meshlet}
Abhishek Badki, Orazio Gallo, Jan Kautz, and Pradeep Sen.
\newblock Meshlet priors for 3d mesh reconstruction.
\newblock In {\em Proceedings of the IEEE/CVF Conference on Computer Vision and
  Pattern Recognition}, pages 2849--2858, 2020.

\bibitem{bangaru2022differentiable}
Sai~Praveen Bangaru, Michael Gharbi, Fujun Luan, Tzu-Mao Li, Kalyan Sunkavalli,
  Milos Hasan, Sai Bi, Zexiang Xu, Gilbert Bernstein, and Fredo Durand.
\newblock Differentiable rendering of neural sdfs through reparameterization.
\newblock In {\em SIGGRAPH Asia 2022 Conference Papers}, pages 1--9, 2022.

\bibitem{barr1981superquadrics}
Alan~H Barr.
\newblock Superquadrics and angle-preserving transformations.
\newblock {\em IEEE Computer graphics and Applications}, 1(1):11--23, 1981.

\bibitem{bernardini1999ball}
Fausto Bernardini, Joshua Mittleman, Holly Rushmeier, Cl{\'a}udio Silva, and
  Gabriel Taubin.
\newblock The ball-pivoting algorithm for surface reconstruction.
\newblock {\em IEEE transactions on visualization and computer graphics},
  5(4):349--359, 1999.

\bibitem{bhatnagar2019multi}
Bharat~Lal Bhatnagar, Garvita Tiwari, Christian Theobalt, and Gerard Pons-Moll.
\newblock Multi-garment net: Learning to dress 3d people from images.
\newblock In {\em proceedings of the IEEE/CVF international conference on
  computer vision}, pages 5420--5430, 2019.

\bibitem{bogo2016keep}
Federica Bogo, Angjoo Kanazawa, Christoph Lassner, Peter Gehler, Javier Romero,
  and Michael~J Black.
\newblock Keep it smpl: Automatic estimation of 3d human pose and shape from a
  single image.
\newblock In {\em European conference on computer vision}, pages 561--578.
  Springer, 2016.

\bibitem{boulch2021needrop}
Alexandre Boulch, Pierre-Alain Langlois, Gilles Puy, and Renaud Marlet.
\newblock Needrop: Self-supervised shape representation from sparse point
  clouds using needle dropping.
\newblock In {\em 2021 International Conference on 3D Vision (3DV)}, pages
  940--950. IEEE, 2021.

\bibitem{boulch2022poco}
Alexandre Boulch and Renaud Marlet.
\newblock Poco: Point convolution for surface reconstruction.
\newblock In {\em Proceedings of the IEEE/CVF Conference on Computer Vision and
  Pattern Recognition}, pages 6302--6314, 2022.

\bibitem{buehler2001unstructured}
Chris Buehler, Michael Bosse, Leonard McMillan, Steven Gortler, and Michael
  Cohen.
\newblock Unstructured lumigraph rendering.
\newblock In {\em Proceedings of the 28th annual conference on Computer
  graphics and interactive techniques}, pages 425--432, 2001.

\bibitem{cao2018learning}
Yan-Pei Cao, Zheng-Ning Liu, Zheng-Fei Kuang, Leif Kobbelt, and Shi-Min Hu.
\newblock Learning to reconstruct high-quality 3d shapes with cascaded fully
  convolutional networks.
\newblock In {\em Proceedings of the European Conference on Computer Vision
  (ECCV)}, pages 616--633, 2018.

\bibitem{cazals2006delaunay}
Fr{\'e}d{\'e}ric Cazals and Joachim Giesen.
\newblock Delaunay triangulation based surface reconstruction.
\newblock In {\em Effective computational geometry for curves and surfaces},
  pages 231--276. Springer, 2006.

\bibitem{chabra2020deep}
Rohan Chabra, Jan~E Lenssen, Eddy Ilg, Tanner Schmidt, Julian Straub, Steven
  Lovegrove, and Richard Newcombe.
\newblock Deep local shapes: Learning local sdf priors for detailed 3d
  reconstruction.
\newblock In {\em European Conference on Computer Vision}, pages 608--625.
  Springer, 2020.

\bibitem{chang2015shapenet}
Angel~X. Chang, Thomas Funkhouser, Leonidas Guibas, Pat Hanrahan, Qixing Huang,
  Zimo Li, Silvio Savarese, Manolis Savva, Shuran Song, Hao Su, Jianxiong Xiao,
  Li~Yi, and Fisher Yu.
\newblock {ShapeNet}: An information-rich {3D} model repository.
\newblock {\em arXiv preprint arXiv:1512.03012}, 2015.

\bibitem{chen2019learning}
Wenzheng Chen, Huan Ling, Jun Gao, Edward Smith, Jaakko Lehtinen, Alec
  Jacobson, and Sanja Fidler.
\newblock Learning to predict 3d objects with an interpolation-based
  differentiable renderer.
\newblock {\em Advances in Neural Information Processing Systems}, 32, 2019.

\bibitem{chen2021dib}
Wenzheng Chen, Joey Litalien, Jun Gao, Zian Wang, Clement Fuji~Tsang, Sameh
  Khamis, Or~Litany, and Sanja Fidler.
\newblock Dib-r++: Learning to predict lighting and material with a hybrid
  differentiable renderer.
\newblock {\em Advances in Neural Information Processing Systems},
  34:22834--22848, 2021.

\bibitem{chen2022mobilenerf}
Zhiqin Chen, Thomas Funkhouser, Peter Hedman, and Andrea Tagliasacchi.
\newblock Mobilenerf: Exploiting the polygon rasterization pipeline for
  efficient neural field rendering on mobile architectures.
\newblock {\em arXiv preprint arXiv:2208.00277}, 2022.

\bibitem{chen2021decor}
Zhiqin Chen, Vladimir~G Kim, Matthew Fisher, Noam Aigerman, Hao Zhang, and
  Siddhartha Chaudhuri.
\newblock Decor-gan: 3d shape detailization by conditional refinement.
\newblock In {\em Proceedings of the IEEE/CVF Conference on Computer Vision and
  Pattern Recognition}, pages 15740--15749, 2021.

\bibitem{chen2022ndc}
Zhiqin Chen, Andrea Tagliasacchi, Thomas Funkhouser, and Hao Zhang.
\newblock Neural dual contouring.
\newblock {\em ACM Transactions on Graphics (Special Issue of SIGGRAPH)},
  41(4), 2022.

\bibitem{chen2020bsp}
Zhiqin Chen, Andrea Tagliasacchi, and Hao Zhang.
\newblock Bsp-net: Generating compact meshes via binary space partitioning.
\newblock In {\em Proceedings of the IEEE/CVF Conference on Computer Vision and
  Pattern Recognition}, pages 45--54, 2020.

\bibitem{chen2019imnet}
Zhiqin Chen and Hao Zhang.
\newblock Learning implicit fields for generative shape modeling.
\newblock In {\em Proceedings of the IEEE/CVF Conference on Computer Vision and
  Pattern Recognition}, pages 5939--5948, 2019.

\bibitem{chen2021nmc}
Zhiqin Chen and Hao Zhang.
\newblock Neural marching cubes.
\newblock {\em ACM Transactions on Graphics (Special Issue of SIGGRAPH Asia)},
  40(6), 2021.

\bibitem{chibane2020implicit}
Julian Chibane, Thiemo Alldieck, and Gerard Pons-Moll.
\newblock Implicit functions in feature space for 3d shape reconstruction and
  completion.
\newblock In {\em Proceedings of the IEEE/CVF Conference on Computer Vision and
  Pattern Recognition}, pages 6970--6981, 2020.

\bibitem{chibane2020neural}
Julian Chibane, Gerard Pons-Moll, et~al.
\newblock Neural unsigned distance fields for implicit function learning.
\newblock {\em Advances in Neural Information Processing Systems},
  33:21638--21652, 2020.

\bibitem{choy20163d}
Christopher~B Choy, Danfei Xu, JunYoung Gwak, Kevin Chen, and Silvio Savarese.
\newblock 3d-r2n2: A unified approach for single and multi-view 3d object
  reconstruction.
\newblock In {\em European conference on computer vision}, pages 628--644.
  Springer, 2016.

\bibitem{coons1967surfaces}
Steven~A Coons.
\newblock Surfaces for computer-aided design of space forms.
\newblock Technical report, MASSACHUSETTS INST OF TECH CAMBRIDGE PROJECT MAC,
  1967.

\bibitem{dai2019scan2mesh}
Angela Dai and Matthias Nie{\ss}ner.
\newblock Scan2mesh: From unstructured range scans to 3d meshes.
\newblock In {\em Proceedings of the IEEE/CVF Conference on Computer Vision and
  Pattern Recognition}, pages 5574--5583, 2019.

\bibitem{dai2017shape}
Angela Dai, Charles Ruizhongtai~Qi, and Matthias Nie{\ss}ner.
\newblock Shape completion using 3d-encoder-predictor cnns and shape synthesis.
\newblock In {\em Proceedings of the IEEE conference on computer vision and
  pattern recognition}, pages 5868--5877, 2017.

\bibitem{darmon2022improving}
Fran{\c{c}}ois Darmon, B{\'e}n{\'e}dicte Bascle, Jean-Cl{\'e}ment Devaux,
  Pascal Monasse, and Mathieu Aubry.
\newblock Improving neural implicit surfaces geometry with patch warping.
\newblock In {\em Proceedings of the IEEE/CVF Conference on Computer Vision and
  Pattern Recognition}, pages 6260--6269, 2022.

\bibitem{deng2020cvxnet}
Boyang Deng, Kyle Genova, Soroosh Yazdani, Sofien Bouaziz, Geoffrey Hinton, and
  Andrea Tagliasacchi.
\newblock Cvxnet: Learnable convex decomposition.
\newblock In {\em Proceedings of the IEEE/CVF Conference on Computer Vision and
  Pattern Recognition}, pages 31--44, 2020.

\bibitem{deng2019accurate}
Yu~Deng, Jiaolong Yang, Sicheng Xu, Dong Chen, Yunde Jia, and Xin Tong.
\newblock Accurate 3d face reconstruction with weakly-supervised learning: From
  single image to image set.
\newblock In {\em Proceedings of the IEEE/CVF Conference on Computer Vision and
  Pattern Recognition Workshops}, pages 0--0, 2019.

\bibitem{doi1991efficient}
Akio Doi and Akio Koide.
\newblock An efficient method of triangulating equi-valued surfaces by using
  tetrahedral cells.
\newblock {\em IEICE Transactions on Information and Systems}, 1991.

\bibitem{erler2020points2surf}
Philipp Erler, Paul Guerrero, Stefan Ohrhallinger, Niloy~J Mitra, and Michael
  Wimmer.
\newblock Points2surf learning implicit surfaces from point clouds.
\newblock In {\em European Conference on Computer Vision}, pages 108--124.
  Springer, 2020.

\bibitem{fugeo}
Qiancheng Fu, Qingshan Xu, Yew-Soon Ong, and Wenbing Tao.
\newblock Geo-neus: Geometry-consistent neural implicit surfaces learning for
  multi-view reconstruction.
\newblock {\em Advances in Neural Information Processing Systems (NeurIPS)},
  2022.

\bibitem{galyean1991sculpting}
Tinsley~A Galyean and John~F Hughes.
\newblock Sculpting: An interactive volumetric modeling technique.
\newblock {\em ACM SIGGRAPH Computer Graphics}, 25(4):267--274, 1991.

\bibitem{gao2020learning}
Jun Gao, Wenzheng Chen, Tommy Xiang, Alec Jacobson, Morgan McGuire, and Sanja
  Fidler.
\newblock Learning deformable tetrahedral meshes for 3d reconstruction.
\newblock {\em Advances In Neural Information Processing Systems},
  33:9936--9947, 2020.

\bibitem{gao2019sdm}
Lin Gao, Jie Yang, Tong Wu, Yu-Jie Yuan, Hongbo Fu, Yu-Kun Lai, and Hao Zhang.
\newblock Sdm-net: Deep generative network for structured deformable mesh.
\newblock {\em ACM Transactions on Graphics (TOG)}, 38(6):1--15, 2019.

\bibitem{genova2020local}
Kyle Genova, Forrester Cole, Avneesh Sud, Aaron Sarna, and Thomas Funkhouser.
\newblock Local deep implicit functions for 3d shape.
\newblock In {\em Proceedings of the IEEE/CVF Conference on Computer Vision and
  Pattern Recognition}, pages 4857--4866, 2020.

\bibitem{genova2019learning}
Kyle Genova, Forrester Cole, Daniel Vlasic, Aaron Sarna, William~T Freeman, and
  Thomas Funkhouser.
\newblock Learning shape templates with structured implicit functions.
\newblock In {\em Proceedings of the IEEE/CVF International Conference on
  Computer Vision}, pages 7154--7164, 2019.

\bibitem{gkioxari2019mesh}
Georgia Gkioxari, Jitendra Malik, and Justin Johnson.
\newblock Mesh r-cnn.
\newblock In {\em Proceedings of the IEEE/CVF International Conference on
  Computer Vision}, pages 9785--9795, 2019.

\bibitem{goel2022differentiable}
Shubham Goel, Georgia Gkioxari, and Jitendra Malik.
\newblock Differentiable stereopsis: Meshes from multiple views using
  differentiable rendering.
\newblock In {\em Proceedings of the IEEE/CVF Conference on Computer Vision and
  Pattern Recognition}, pages 8635--8644, 2022.

\bibitem{goel2020shape}
Shubham Goel, Angjoo Kanazawa, and Jitendra Malik.
\newblock Shape and viewpoint without keypoints.
\newblock In {\em European Conference on Computer Vision}, pages 88--104.
  Springer, 2020.

\bibitem{goodfellow2020generative}
Ian Goodfellow, Jean Pouget-Abadie, Mehdi Mirza, Bing Xu, David Warde-Farley,
  Sherjil Ozair, Aaron Courville, and Yoshua Bengio.
\newblock Generative adversarial networks.
\newblock {\em Communications of the ACM}, 63(11):139--144, 2020.

\bibitem{gropp2020implicit}
Amos Gropp, Lior Yariv, Niv Haim, Matan Atzmon, and Yaron Lipman.
\newblock Implicit geometric regularization for learning shapes.
\newblock In {\em Proceedings of the 37th International Conference on Machine
  Learning}, pages 3789--3799, 2020.

\bibitem{groueix2018papier}
Thibault Groueix, Matthew Fisher, Vladimir~G Kim, Bryan~C Russell, and Mathieu
  Aubry.
\newblock A papier-m{\^a}ch{\'e} approach to learning 3d surface generation.
\newblock In {\em Proceedings of the IEEE conference on computer vision and
  pattern recognition}, pages 216--224, 2018.

\bibitem{gu2002geometry}
Xianfeng Gu, Steven~J Gortler, and Hugues Hoppe.
\newblock Geometry images.
\newblock In {\em Proceedings of the 29th annual conference on Computer
  graphics and interactive techniques}, pages 355--361, 2002.

\bibitem{guillard2022meshudf}
Benoit Guillard, Federico Stella, and Pascal Fua.
\newblock Meshudf: Fast and differentiable meshing of unsigned distance field
  networks.
\newblock In {\em European Conference on Computer Vision}, pages 576--592.
  Springer, 2022.

\bibitem{guo2022complexgen}
Haoxiang Guo, Shilin Liu, Hao Pan, Yang Liu, Xin Tong, and Baining Guo.
\newblock Complexgen: Cad reconstruction by b-rep chain complex generation.
\newblock {\em ACM Transactions on Graphics (TOG)}, 41(4):1--18, 2022.

\bibitem{guo2022neural}
Haoyu Guo, Sida Peng, Haotong Lin, Qianqian Wang, Guofeng Zhang, Hujun Bao, and
  Xiaowei Zhou.
\newblock Neural 3d scene reconstruction with the manhattan-world assumption.
\newblock In {\em Proceedings of the IEEE/CVF Conference on Computer Vision and
  Pattern Recognition}, pages 5511--5520, 2022.

\bibitem{gupta2020neural}
Kunal Gupta and Manmohan Chandraker.
\newblock Neural mesh flow: 3d manifold mesh generation via diffeomorphic
  flows.
\newblock In {\em Proceedings of the 34th International Conference on Neural
  Information Processing Systems}, pages 1747--1758, 2020.

\bibitem{hane2017hierarchical}
Christian H{\"a}ne, Shubham Tulsiani, and Jitendra Malik.
\newblock Hierarchical surface prediction for 3d object reconstruction.
\newblock In {\em 2017 International Conference on 3D Vision (3DV)}, pages
  412--420. IEEE, 2017.

\bibitem{hanocka2020point2mesh}
Rana Hanocka, Gal Metzer, Raja Giryes, and Daniel Cohen-Or.
\newblock Point2mesh: a self-prior for deformable meshes.
\newblock {\em ACM Transactions on Graphics (TOG)}, 39(4):126--1, 2020.

\bibitem{he2016deep}
Kaiming He, Xiangyu Zhang, Shaoqing Ren, and Jian Sun.
\newblock Deep residual learning for image recognition.
\newblock In {\em Proceedings of the IEEE conference on computer vision and
  pattern recognition}, pages 770--778, 2016.

\bibitem{henderson2020leveraging}
Paul Henderson, Vagia Tsiminaki, and Christoph~H Lampert.
\newblock Leveraging 2d data to learn textured 3d mesh generation.
\newblock In {\em Proceedings of the IEEE/CVF Conference on Computer Vision and
  Pattern Recognition}, pages 7498--7507, 2020.

\bibitem{hochreiter1997long}
Sepp Hochreiter and J{\"u}rgen Schmidhuber.
\newblock Long short-term memory.
\newblock {\em Neural computation}, 9(8):1735--1780, 1997.

\bibitem{hui2022neural}
Ka-Hei Hui, Ruihui Li, Jingyu Hu, and Chi-Wing Fu.
\newblock Neural template: Topology-aware reconstruction and disentangled
  generation of 3d meshes.
\newblock In {\em Proceedings of the IEEE/CVF Conference on Computer Vision and
  Pattern Recognition}, pages 18572--18582, 2022.

\bibitem{insafutdinov2022snes}
Eldar Insafutdinov, Dylan Campbell, Jo{\~a}o~F Henriques, and Andrea Vedaldi.
\newblock Snes: Learning probably symmetric neural surfaces from incomplete
  data.
\newblock In {\em European Conference on Computer Vision (ECCV)}, pages
  367--383. Springer, 2022.

\bibitem{isola2017image}
Phillip Isola, Jun-Yan Zhu, Tinghui Zhou, and Alexei~A Efros.
\newblock Image-to-image translation with conditional adversarial networks.
\newblock In {\em Proceedings of the IEEE conference on computer vision and
  pattern recognition}, pages 1125--1134, 2017.

\bibitem{jiang2020bcnet}
Boyi Jiang, Juyong Zhang, Yang Hong, Jinhao Luo, Ligang Liu, and Hujun Bao.
\newblock Bcnet: Learning body and cloth shape from a single image.
\newblock In {\em European Conference on Computer Vision}, pages 18--35.
  Springer, 2020.

\bibitem{jiang2020shapeflow}
Chiyu Jiang, Jingwei Huang, Andrea Tagliasacchi, and Leonidas~J Guibas.
\newblock Shapeflow: Learnable deformation flows among 3d shapes.
\newblock {\em Advances in Neural Information Processing Systems},
  33:9745--9757, 2020.

\bibitem{jiang2020local}
Chiyu Jiang, Avneesh Sud, Ameesh Makadia, Jingwei Huang, Matthias Nie{\ss}ner,
  Thomas Funkhouser, et~al.
\newblock Local implicit grid representations for 3d scenes.
\newblock In {\em Proceedings of the IEEE/CVF Conference on Computer Vision and
  Pattern Recognition}, pages 6001--6010, 2020.

\bibitem{jiang2020sdfdiff}
Yue Jiang, Dantong Ji, Zhizhong Han, and Matthias Zwicker.
\newblock Sdfdiff: Differentiable rendering of signed distance fields for 3d
  shape optimization.
\newblock In {\em Proceedings of the IEEE/CVF conference on computer vision and
  pattern recognition}, pages 1251--1261, 2020.

\bibitem{ju2002dual}
Tao Ju, Frank Losasso, Scott Schaefer, and Joe Warren.
\newblock Dual contouring of {Hermite} data.
\newblock {\em ACM Transactions on graphics}, 21(3):339–346, 2002.

\bibitem{kanazawa2018learning}
Angjoo Kanazawa, Shubham Tulsiani, Alexei~A Efros, and Jitendra Malik.
\newblock Learning category-specific mesh reconstruction from image
  collections.
\newblock In {\em Proceedings of the European Conference on Computer Vision
  (ECCV)}, pages 371--386, 2018.

\bibitem{kania2020ucsg}
Kacper Kania, Maciej Zieba, and Tomasz Kajdanowicz.
\newblock Ucsg-net-unsupervised discovering of constructive solid geometry
  tree.
\newblock {\em Advances in Neural Information Processing Systems},
  33:8776--8786, 2020.

\bibitem{kar2015category}
Abhishek Kar, Shubham Tulsiani, Joao Carreira, and Jitendra Malik.
\newblock Category-specific object reconstruction from a single image.
\newblock In {\em Proceedings of the IEEE conference on computer vision and
  pattern recognition}, pages 1966--1974, 2015.

\bibitem{karras2019style}
Tero Karras, Samuli Laine, and Timo Aila.
\newblock A style-based generator architecture for generative adversarial
  networks.
\newblock In {\em Proceedings of the IEEE/CVF conference on computer vision and
  pattern recognition}, pages 4401--4410, 2019.

\bibitem{kato2018neural}
Hiroharu Kato, Yoshitaka Ushiku, and Tatsuya Harada.
\newblock Neural 3d mesh renderer.
\newblock In {\em Proceedings of the IEEE conference on computer vision and
  pattern recognition}, pages 3907--3916, 2018.

\bibitem{kawana2020neural}
Yuki Kawana, Yusuke Mukuta, and Tatsuya Harada.
\newblock Neural star domain as primitive representation.
\newblock {\em Advances in Neural Information Processing Systems},
  33:7875--7886, 2020.

\bibitem{kazhdan2006poisson}
Michael Kazhdan, Matthew Bolitho, and Hugues Hoppe.
\newblock Poisson surface reconstruction.
\newblock In {\em Proceedings of the fourth Eurographics symposium on Geometry
  processing}, volume~7, 2006.

\bibitem{kazhdan2013screened}
Michael Kazhdan and Hugues Hoppe.
\newblock Screened poisson surface reconstruction.
\newblock {\em ACM Transactions on Graphics (ToG)}, 32(3):1--13, 2013.

\bibitem{kellnhofer2021neural}
Petr Kellnhofer, Lars~C Jebe, Andrew Jones, Ryan Spicer, Kari Pulli, and Gordon
  Wetzstein.
\newblock Neural lumigraph rendering.
\newblock In {\em Proceedings of the IEEE/CVF Conference on Computer Vision and
  Pattern Recognition}, pages 4287--4297, 2021.

\bibitem{VAE}
Diederik~P Kingma and Max Welling.
\newblock Auto-encoding variational bayes.
\newblock {\em International Conference on Learning Representations (ICLR)},
  2014.

\bibitem{kolluri2008provably}
Ravikrishna Kolluri.
\newblock Provably good moving least squares.
\newblock {\em ACM Transactions on Algorithms (TALG)}, 4(2):1--25, 2008.

\bibitem{kolotouros2019convolutional}
Nikos Kolotouros, Georgios Pavlakos, and Kostas Daniilidis.
\newblock Convolutional mesh regression for single-image human shape
  reconstruction.
\newblock In {\em Proceedings of the IEEE/CVF Conference on Computer Vision and
  Pattern Recognition}, pages 4501--4510, 2019.

\bibitem{lambourne2022reconstructing}
Joseph~George Lambourne, Karl Willis, Pradeep~Kumar Jayaraman, Longfei Zhang,
  Aditya Sanghi, and Kamal~Rahimi Malekshan.
\newblock Reconstructing editable prismatic cad from rounded voxel models.
\newblock In {\em SIGGRAPH Asia 2022 Conference Papers}, pages 1--9, 2022.

\bibitem{le2021cpfn}
Eric-Tuan L{\^e}, Minhyuk Sung, Duygu Ceylan, Radomir Mech, Tamy Boubekeur, and
  Niloy~J Mitra.
\newblock Cpfn: Cascaded primitive fitting networks for high-resolution point
  clouds.
\newblock In {\em Proceedings of the IEEE/CVF International Conference on
  Computer Vision}, pages 7457--7466, 2021.

\bibitem{lei2020analytic}
Jiabao Lei and Kui Jia.
\newblock Analytic marching: An analytic meshing solution from deep implicit
  surface networks.
\newblock In {\em International Conference on Machine Learning}, pages
  5789--5798. PMLR, 2020.

\bibitem{li2019supervised}
Lingxiao Li, Minhyuk Sung, Anastasia Dubrovina, Li~Yi, and Leonidas~J Guibas.
\newblock Supervised fitting of geometric primitives to 3d point clouds.
\newblock In {\em Proceedings of the IEEE/CVF Conference on Computer Vision and
  Pattern Recognition}, pages 2652--2660, 2019.

\bibitem{li2021d2im}
Manyi Li and Hao Zhang.
\newblock D2im-net: Learning detail disentangled implicit fields from single
  images.
\newblock In {\em Proceedings of the IEEE/CVF Conference on Computer Vision and
  Pattern Recognition}, pages 10246--10255, 2021.

\bibitem{li2020self}
Xueting Li, Sifei Liu, Kihwan Kim, Shalini~De Mello, Varun Jampani, Ming-Hsuan
  Yang, and Jan Kautz.
\newblock Self-supervised single-view 3d reconstruction via semantic
  consistency.
\newblock In {\em European Conference on Computer Vision}, pages 677--693.
  Springer, 2020.

\bibitem{liao2018deep}
Yiyi Liao, Simon Donne, and Andreas Geiger.
\newblock Deep marching cubes: Learning explicit surface representations.
\newblock In {\em Proceedings of the IEEE Conference on Computer Vision and
  Pattern Recognition}, pages 2916--2925, 2018.

\bibitem{lin2019photometric}
Chen-Hsuan Lin, Oliver Wang, Bryan~C Russell, Eli Shechtman, Vladimir~G Kim,
  Matthew Fisher, and Simon Lucey.
\newblock Photometric mesh optimization for video-aligned 3d object
  reconstruction.
\newblock In {\em Proceedings of the IEEE/CVF Conference on Computer Vision and
  Pattern Recognition}, pages 969--978, 2019.

\bibitem{lin2020modeling}
Cheng Lin, Tingxiang Fan, Wenping Wang, and Matthias Nie{\ss}ner.
\newblock Modeling 3d shapes by reinforcement learning.
\newblock In {\em European Conference on Computer Vision}, pages 545--561.
  Springer, 2020.

\bibitem{littwin2019deep}
Gidi Littwin and Lior Wolf.
\newblock Deep meta functionals for shape representation.
\newblock In {\em Proceedings of the IEEE/CVF International Conference on
  Computer Vision}, pages 1824--1833, 2019.

\bibitem{liu20222d}
Feng Liu and Xiaoming Liu.
\newblock 2d gans meet unsupervised single-view 3d reconstruction.
\newblock In {\em European Conference on Computer Vision}, pages 497--514.
  Springer, 2022.

\bibitem{liu2020meshing}
Minghua Liu, Xiaoshuai Zhang, and Hao Su.
\newblock Meshing point clouds with predicted intrinsic-extrinsic ratio
  guidance.
\newblock In {\em European Conference on Computer Vision}, pages 68--84.
  Springer, 2020.

\bibitem{liu2020dist}
Shaohui Liu, Yinda Zhang, Songyou Peng, Boxin Shi, Marc Pollefeys, and Zhaopeng
  Cui.
\newblock Dist: Rendering deep implicit signed distance function with
  differentiable sphere tracing.
\newblock In {\em Proceedings of the IEEE/CVF Conference on Computer Vision and
  Pattern Recognition}, pages 2019--2028, 2020.

\bibitem{liu2021deep}
Shi-Lin Liu, Hao-Xiang Guo, Hao Pan, Peng-Shuai Wang, Xin Tong, and Yang Liu.
\newblock Deep implicit moving least-squares functions for 3d reconstruction.
\newblock In {\em Proceedings of the IEEE/CVF Conference on Computer Vision and
  Pattern Recognition}, pages 1788--1797, 2021.

\bibitem{liu2019soft}
Shichen Liu, Tianye Li, Weikai Chen, and Hao Li.
\newblock Soft rasterizer: A differentiable renderer for image-based 3d
  reasoning.
\newblock In {\em Proceedings of the IEEE/CVF International Conference on
  Computer Vision}, pages 7708--7717, 2019.

\bibitem{liu2019learning}
Shichen Liu, Shunsuke Saito, Weikai Chen, and Hao Li.
\newblock Learning to infer implicit surfaces without 3d supervision.
\newblock {\em Advances in Neural Information Processing Systems}, 32, 2019.

\bibitem{long2022sparseneus}
Xiaoxiao Long, Cheng Lin, Peng Wang, Taku Komura, and Wenping Wang.
\newblock Sparseneus: Fast generalizable neural surface reconstruction from
  sparse views.
\newblock In {\em Computer Vision--ECCV 2022: 17th European Conference, Tel
  Aviv, Israel, October 23--27, 2022, Proceedings, Part XXXII}, pages 210--227.
  Springer, 2022.

\bibitem{lorensen1987marching}
William~E Lorensen and Harvey~E Cline.
\newblock Marching cubes: A high resolution 3d surface construction algorithm.
\newblock {\em ACM siggraph computer graphics}, 21(4):163--169, 1987.

\bibitem{luo2021deepdt}
Yiming Luo, Zhenxing Mi, and Wenbing Tao.
\newblock Deepdt: Learning geometry from delaunay triangulation for surface
  reconstruction.
\newblock In {\em Proceedings of the AAAI Conference on Artificial
  Intelligence}, volume~35, pages 2277--2285, 2021.

\bibitem{martin2021nerf}
Ricardo Martin-Brualla, Noha Radwan, Mehdi~SM Sajjadi, Jonathan~T Barron,
  Alexey Dosovitskiy, and Daniel Duckworth.
\newblock Nerf in the wild: Neural radiance fields for unconstrained photo
  collections.
\newblock In {\em Proceedings of the IEEE/CVF Conference on Computer Vision and
  Pattern Recognition}, pages 7210--7219, 2021.

\bibitem{mescheder2019occupancy}
Lars Mescheder, Michael Oechsle, Michael Niemeyer, Sebastian Nowozin, and
  Andreas Geiger.
\newblock Occupancy networks: Learning 3d reconstruction in function space.
\newblock In {\em Proceedings of the IEEE/CVF conference on computer vision and
  pattern recognition}, pages 4460--4470, 2019.

\bibitem{michalkiewicz2019deep}
Mateusz Michalkiewicz, Jhony~K Pontes, Dominic Jack, Mahsa Baktashmotlagh, and
  Anders Eriksson.
\newblock Implicit surface representations as layers in neural networks.
\newblock In {\em Proceedings of the IEEE/CVF International Conference on
  Computer Vision}, pages 4743--4752, 2019.

\bibitem{mildenhall2021nerf}
Ben Mildenhall, Pratul~P Srinivasan, Matthew Tancik, Jonathan~T Barron, Ravi
  Ramamoorthi, and Ren Ng.
\newblock Nerf: Representing scenes as neural radiance fields for view
  synthesis.
\newblock In {\em ECCV}, pages 405--421. Springer, 2020.

\bibitem{mittal2022autosdf}
Paritosh Mittal, Yen-Chi Cheng, Maneesh Singh, and Shubham Tulsiani.
\newblock Autosdf: Shape priors for 3d completion, reconstruction and
  generation.
\newblock In {\em Proceedings of the IEEE/CVF Conference on Computer Vision and
  Pattern Recognition}, pages 306--315, 2022.

\bibitem{monnier2022unicorn}
Tom Monnier, Matthew Fisher, Alexei~A. Efros, and Mathieu Aubry.
\newblock {Share With Thy Neighbors: Single-View Reconstruction by
  Cross-Instance Consistency}.
\newblock In {\em {ECCV}}, 2022.

\bibitem{mueller2022instant}
Thomas M\"uller, Alex Evans, Christoph Schied, and Alexander Keller.
\newblock Instant neural graphics primitives with a multiresolution hash
  encoding.
\newblock {\em ACM Trans. Graph.}, 41(4):102:1--102:15, July 2022.

\bibitem{munkberg2022extracting}
Jacob Munkberg, Jon Hasselgren, Tianchang Shen, Jun Gao, Wenzheng Chen, Alex
  Evans, Thomas M{\"u}ller, and Sanja Fidler.
\newblock Extracting triangular 3d models, materials, and lighting from images.
\newblock In {\em Proceedings of the IEEE/CVF Conference on Computer Vision and
  Pattern Recognition}, pages 8280--8290, 2022.

\bibitem{nash2020polygen}
Charlie Nash, Yaroslav Ganin, SM~Ali Eslami, and Peter Battaglia.
\newblock Polygen: An autoregressive generative model of 3d meshes.
\newblock In {\em International conference on machine learning}, pages
  7220--7229. PMLR, 2020.

\bibitem{niemeyer2020differentiable}
Michael Niemeyer, Lars Mescheder, Michael Oechsle, and Andreas Geiger.
\newblock Differentiable volumetric rendering: Learning implicit 3d
  representations without 3d supervision.
\newblock In {\em Proceedings of the IEEE/CVF Conference on Computer Vision and
  Pattern Recognition}, pages 3504--3515, 2020.

\bibitem{niu2018im2struct}
Chengjie Niu, Jun Li, and Kai Xu.
\newblock Im2struct: Recovering 3d shape structure from a single rgb image.
\newblock In {\em Proceedings of the IEEE conference on computer vision and
  pattern recognition}, pages 4521--4529, 2018.

\bibitem{oechsle2019texture}
Michael Oechsle, Lars Mescheder, Michael Niemeyer, Thilo Strauss, and Andreas
  Geiger.
\newblock Texture fields: Learning texture representations in function space.
\newblock In {\em Proceedings of the IEEE/CVF International Conference on
  Computer Vision}, pages 4531--4540, 2019.

\bibitem{oechsle2021unisurf}
Michael Oechsle, Songyou Peng, and Andreas Geiger.
\newblock Unisurf: Unifying neural implicit surfaces and radiance fields for
  multi-view reconstruction.
\newblock In {\em Proceedings of the IEEE/CVF International Conference on
  Computer Vision}, pages 5589--5599, 2021.

\bibitem{pan2019deep}
Junyi Pan, Xiaoguang Han, Weikai Chen, Jiapeng Tang, and Kui Jia.
\newblock Deep mesh reconstruction from single rgb images via topology
  modification networks.
\newblock In {\em Proceedings of the IEEE/CVF International Conference on
  Computer Vision}, pages 9964--9973, 2019.

\bibitem{park2019deepsdf}
Jeong~Joon Park, Peter Florence, Julian Straub, Richard Newcombe, and Steven
  Lovegrove.
\newblock Deepsdf: Learning continuous signed distance functions for shape
  representation.
\newblock In {\em Proceedings of the IEEE/CVF conference on computer vision and
  pattern recognition}, pages 165--174, 2019.

\bibitem{paschalidou2020learning}
Despoina Paschalidou, Luc~Van Gool, and Andreas Geiger.
\newblock Learning unsupervised hierarchical part decomposition of 3d objects
  from a single rgb image.
\newblock In {\em Proceedings of the IEEE/CVF Conference on Computer Vision and
  Pattern Recognition}, pages 1060--1070, 2020.

\bibitem{paschalidou2019superquadrics}
Despoina Paschalidou, Ali~Osman Ulusoy, and Andreas Geiger.
\newblock Superquadrics revisited: Learning 3d shape parsing beyond cuboids.
\newblock In {\em Proceedings of the IEEE/CVF Conference on Computer Vision and
  Pattern Recognition}, pages 10344--10353, 2019.

\bibitem{pavlakos2018learning}
Georgios Pavlakos, Luyang Zhu, Xiaowei Zhou, and Kostas Daniilidis.
\newblock Learning to estimate 3d human pose and shape from a single color
  image.
\newblock In {\em Proceedings of the IEEE conference on computer vision and
  pattern recognition}, pages 459--468, 2018.

\bibitem{pavllo2020convolutional}
Dario Pavllo, Graham Spinks, Thomas Hofmann, Marie-Francine Moens, and Aurelien
  Lucchi.
\newblock Convolutional generation of textured 3d meshes.
\newblock {\em Advances in Neural Information Processing Systems}, 33:870--882,
  2020.

\bibitem{peng2021shape}
Songyou Peng, Chiyu Jiang, Yiyi Liao, Michael Niemeyer, Marc Pollefeys, and
  Andreas Geiger.
\newblock Shape as points: A differentiable poisson solver.
\newblock {\em Advances in Neural Information Processing Systems},
  34:13032--13044, 2021.

\bibitem{peng2020convolutional}
Songyou Peng, Michael Niemeyer, Lars Mescheder, Marc Pollefeys, and Andreas
  Geiger.
\newblock Convolutional occupancy networks.
\newblock In {\em European Conference on Computer Vision}, pages 523--540.
  Springer, 2020.

\bibitem{qi2017pointnet}
Charles~R Qi, Hao Su, Kaichun Mo, and Leonidas~J Guibas.
\newblock Pointnet: Deep learning on point sets for 3d classification and
  segmentation.
\newblock In {\em Proceedings of the IEEE conference on computer vision and
  pattern recognition}, pages 652--660, 2017.

\bibitem{rakotosaona2021learning}
Marie-Julie Rakotosaona, Paul Guerrero, Noam Aigerman, Niloy~J Mitra, and Maks
  Ovsjanikov.
\newblock Learning delaunay surface elements for mesh reconstruction.
\newblock In {\em Proceedings of the IEEE/CVF Conference on Computer Vision and
  Pattern Recognition}, pages 22--31, 2021.

\bibitem{remelli2020meshsdf}
Edoardo Remelli, Artem Lukoianov, Stephan Richter, Beno{\^\i}t Guillard, Timur
  Bagautdinov, Pierre Baque, and Pascal Fua.
\newblock Meshsdf: Differentiable iso-surface extraction.
\newblock {\em Advances in Neural Information Processing Systems},
  33:22468--22478, 2020.

\bibitem{ren2021csg}
Daxuan Ren, Jianmin Zheng, Jianfei Cai, Jiatong Li, Haiyong Jiang, Zhongang
  Cai, Junzhe Zhang, Liang Pan, Mingyuan Zhang, Haiyu Zhao, et~al.
\newblock Csg-stump: A learning friendly csg-like representation for
  interpretable shape parsing.
\newblock In {\em Proceedings of the IEEE/CVF International Conference on
  Computer Vision}, pages 12478--12487, 2021.

\bibitem{ren2022extrudenet}
Daxuan Ren, Jianmin Zheng, Jianfei Cai, Jiatong Li, and Junzhe Zhang.
\newblock Extrudenet: Unsupervised inverse sketch-and-extrude for shape
  parsing.
\newblock In {\em European Conference on Computer Vision}, pages 482--498.
  Springer, 2022.

\bibitem{richter2018matryoshka}
Stephan~R Richter and Stefan Roth.
\newblock Matryoshka networks: Predicting 3d geometry via nested shape layers.
\newblock In {\em Proceedings of the IEEE conference on computer vision and
  pattern recognition}, pages 1936--1944, 2018.

\bibitem{riegler2017octnetfusion}
Gernot Riegler, Ali~Osman Ulusoy, Horst Bischof, and Andreas Geiger.
\newblock Octnetfusion: Learning depth fusion from data.
\newblock In {\em 2017 International Conference on 3D Vision (3DV)}, pages
  57--66. IEEE, 2017.

\bibitem{ronneberger2015u}
Olaf Ronneberger, Philipp Fischer, and Thomas Brox.
\newblock U-net: Convolutional networks for biomedical image segmentation.
\newblock In {\em International Conference on Medical image computing and
  computer-assisted intervention}, pages 234--241. Springer, 2015.

\bibitem{saito2019pifu}
Shunsuke Saito, Zeng Huang, Ryota Natsume, Shigeo Morishima, Angjoo Kanazawa,
  and Hao Li.
\newblock Pifu: Pixel-aligned implicit function for high-resolution clothed
  human digitization.
\newblock In {\em Proceedings of the IEEE/CVF International Conference on
  Computer Vision}, pages 2304--2314, 2019.

\bibitem{saito2020pifuhd}
Shunsuke Saito, Tomas Simon, Jason Saragih, and Hanbyul Joo.
\newblock Pifuhd: Multi-level pixel-aligned implicit function for
  high-resolution 3d human digitization.
\newblock In {\em Proceedings of the IEEE/CVF Conference on Computer Vision and
  Pattern Recognition}, pages 84--93, 2020.

\bibitem{schoenberger2016sfm}
Johannes~Lutz Sch\"{o}nberger and Jan-Michael Frahm.
\newblock Structure-from-motion revisited.
\newblock In {\em Conference on Computer Vision and Pattern Recognition
  (CVPR)}, 2016.

\bibitem{schoenberger2016mvs}
Johannes~Lutz Sch\"{o}nberger, Enliang Zheng, Marc Pollefeys, and Jan-Michael
  Frahm.
\newblock Pixelwise view selection for unstructured multi-view stereo.
\newblock In {\em European Conference on Computer Vision (ECCV)}, 2016.

\bibitem{sharma2018csgnet}
Gopal Sharma, Rishabh Goyal, Difan Liu, Evangelos Kalogerakis, and Subhransu
  Maji.
\newblock Csgnet: Neural shape parser for constructive solid geometry.
\newblock In {\em Proceedings of the IEEE Conference on Computer Vision and
  Pattern Recognition}, pages 5515--5523, 2018.

\bibitem{sharma2020parsenet}
Gopal Sharma, Difan Liu, Subhransu Maji, Evangelos Kalogerakis, Siddhartha
  Chaudhuri, and Radom{\'\i}r M{\v{e}}ch.
\newblock Parsenet: A parametric surface fitting network for 3d point clouds.
\newblock In {\em European Conference on Computer Vision}, pages 261--276.
  Springer, 2020.

\bibitem{sharp2020pointtrinet}
Nicholas Sharp and Maks Ovsjanikov.
\newblock Pointtrinet: Learned triangulation of 3d point sets.
\newblock In {\em European Conference on Computer Vision}, pages 762--778.
  Springer, 2020.

\bibitem{shen2021deep}
Tianchang Shen, Jun Gao, Kangxue Yin, Ming-Yu Liu, and Sanja Fidler.
\newblock Deep marching tetrahedra: a hybrid representation for high-resolution
  3d shape synthesis.
\newblock {\em Advances in Neural Information Processing Systems},
  34:6087--6101, 2021.

\bibitem{shi2021geometric}
Yue Shi, Bingbing Ni, Jinxian Liu, Dingyi Rong, Ye~Qian, and Wenjun Zhang.
\newblock Geometric granularity aware pixel-to-mesh.
\newblock In {\em Proceedings of the IEEE/CVF International Conference on
  Computer Vision}, pages 13097--13106, 2021.

\bibitem{siddiqui2021retrievalfuse}
Yawar Siddiqui, Justus Thies, Fangchang Ma, Qi~Shan, Matthias Nie{\ss}ner, and
  Angela Dai.
\newblock Retrievalfuse: Neural 3d scene reconstruction with a database.
\newblock In {\em Proceedings of the IEEE/CVF International Conference on
  Computer Vision}, pages 12568--12577, 2021.

\bibitem{sinha2017surfnet}
Ayan Sinha, Asim Unmesh, Qixing Huang, and Karthik Ramani.
\newblock Surfnet: Generating 3d shape surfaces using deep residual networks.
\newblock In {\em Proceedings of the IEEE conference on computer vision and
  pattern recognition}, pages 6040--6049, 2017.

\bibitem{sitzmann2020implicit}
Vincent Sitzmann, Julien Martel, Alexander Bergman, David Lindell, and Gordon
  Wetzstein.
\newblock Implicit neural representations with periodic activation functions.
\newblock {\em Advances in Neural Information Processing Systems},
  33:7462--7473, 2020.

\bibitem{smirnov2020learning}
Dmitriy Smirnov, Mikhail Bessmeltsev, and Justin Solomon.
\newblock Learning manifold patch-based representations of man-made shapes.
\newblock In {\em International Conference on Learning Representations}, 2020.

\bibitem{sudhakaran2021growing}
Shyam Sudhakaran, Djordje Grbic, Siyan Li, Adam Katona, Elias Najarro, Claire
  Glanois, and Sebastian Risi.
\newblock Growing 3d artefacts and functional machines with neural cellular
  automata.
\newblock {\em arXiv preprint arXiv:2103.08737}, 2021.

\bibitem{sun2022neural}
Jiaming Sun, Xi~Chen, Qianqian Wang, Zhengqi Li, Hadar Averbuch-Elor, Xiaowei
  Zhou, and Noah Snavely.
\newblock Neural 3d reconstruction in the wild.
\newblock In {\em ACM SIGGRAPH 2022 Conference Proceedings}, pages 1--9, 2022.

\bibitem{tancik2020fourier}
Matthew Tancik, Pratul Srinivasan, Ben Mildenhall, Sara Fridovich-Keil, Nithin
  Raghavan, Utkarsh Singhal, Ravi Ramamoorthi, Jonathan Barron, and Ren Ng.
\newblock Fourier features let networks learn high frequency functions in low
  dimensional domains.
\newblock {\em Advances in Neural Information Processing Systems},
  33:7537--7547, 2020.

\bibitem{tang2021octfield}
Jia-Heng Tang, Weikai Chen, Bo~Wang, Songrun Liu, Bo~Yang, Lin Gao, et~al.
\newblock Octfield: Hierarchical implicit functions for 3d modeling.
\newblock {\em Advances in Neural Information Processing Systems},
  34:12648--12660, 2021.

\bibitem{tang2019skeleton}
Jiapeng Tang, Xiaoguang Han, Junyi Pan, Kui Jia, and Xin Tong.
\newblock A skeleton-bridged deep learning approach for generating meshes of
  complex topologies from single rgb images.
\newblock In {\em Proceedings of the ieee/cvf conference on computer vision and
  pattern recognition}, pages 4541--4550, 2019.

\bibitem{tang2021sa}
Jiapeng Tang, Jiabao Lei, Dan Xu, Feiying Ma, Kui Jia, and Lei Zhang.
\newblock Sa-convonet: Sign-agnostic optimization of convolutional occupancy
  networks.
\newblock In {\em Proceedings of the IEEE/CVF International Conference on
  Computer Vision}, pages 6504--6513, 2021.

\bibitem{tatarchenko2017octree}
Maxim Tatarchenko, Alexey Dosovitskiy, and Thomas Brox.
\newblock Octree generating networks: Efficient convolutional architectures for
  high-resolution 3d outputs.
\newblock In {\em Proceedings of the IEEE international conference on computer
  vision}, pages 2088--2096, 2017.

\bibitem{tatarchenko2019single}
Maxim Tatarchenko, Stephan~R Richter, Ren{\'e} Ranftl, Zhuwen Li, Vladlen
  Koltun, and Thomas Brox.
\newblock What do single-view 3d reconstruction networks learn?
\newblock In {\em Proceedings of the IEEE/CVF conference on computer vision and
  pattern recognition}, pages 3405--3414, 2019.

\bibitem{tulsiani2017learning}
Shubham Tulsiani, Hao Su, Leonidas~J Guibas, Alexei~A Efros, and Jitendra
  Malik.
\newblock Learning shape abstractions by assembling volumetric primitives.
\newblock In {\em Proceedings of the IEEE Conference on Computer Vision and
  Pattern Recognition}, pages 2635--2643, 2017.

\bibitem{ulyanov2018deep}
Dmitry Ulyanov, Andrea Vedaldi, and Victor Lempitsky.
\newblock Deep image prior.
\newblock In {\em Proceedings of the IEEE conference on computer vision and
  pattern recognition}, pages 9446--9454, 2018.

\bibitem{ummenhofer2021adaptive}
Benjamin Ummenhofer and Vladlen Koltun.
\newblock Adaptive surface reconstruction with multiscale convolutional
  kernels.
\newblock In {\em Proceedings of the IEEE/CVF International Conference on
  Computer Vision}, pages 5651--5660, 2021.

\bibitem{uy2022point2cyl}
Mikaela~Angelina Uy, Yen-Yu Chang, Minhyuk Sung, Purvi Goel, Joseph~G
  Lambourne, Tolga Birdal, and Leonidas~J Guibas.
\newblock Point2cyl: Reverse engineering 3d objects from point clouds to
  extrusion cylinders.
\newblock In {\em Proceedings of the IEEE/CVF Conference on Computer Vision and
  Pattern Recognition}, pages 11850--11860, 2022.

\bibitem{uy2020deformation}
Mikaela~Angelina Uy, Jingwei Huang, Minhyuk Sung, Tolga Birdal, and Leonidas
  Guibas.
\newblock Deformation-aware 3d model embedding and retrieval.
\newblock In {\em European Conference on Computer Vision}, pages 397--413.
  Springer, 2020.

\bibitem{vaswani2017attention}
Ashish Vaswani, Noam Shazeer, Niki Parmar, Jakob Uszkoreit, Llion Jones,
  Aidan~N Gomez, {\L}ukasz Kaiser, and Illia Polosukhin.
\newblock Attention is all you need.
\newblock {\em Advances in neural information processing systems}, 30, 2017.

\bibitem{wang2021multi}
Dan Wang, Xinrui Cui, Xun Chen, Zhengxia Zou, Tianyang Shi, Septimiu Salcudean,
  Z~Jane Wang, and Rabab Ward.
\newblock Multi-view 3d reconstruction with transformers.
\newblock In {\em Proceedings of the IEEE/CVF International Conference on
  Computer Vision}, pages 5722--5731, 2021.

\bibitem{wang2018pixel2mesh}
Nanyang Wang, Yinda Zhang, Zhuwen Li, Yanwei Fu, Wei Liu, and Yu-Gang Jiang.
\newblock Pixel2mesh: Generating 3d mesh models from single rgb images.
\newblock In {\em Proceedings of the European conference on computer vision
  (ECCV)}, pages 52--67, 2018.

\bibitem{wang2021neus}
Peng Wang, Lingjie Liu, Yuan Liu, Christian Theobalt, Taku Komura, and Wenping
  Wang.
\newblock Neus: Learning neural implicit surfaces by volume rendering for
  multi-view reconstruction.
\newblock {\em Advances in Neural Information Processing Systems},
  34:27171--27183, 2021.

\bibitem{wang2017cnn}
Peng-Shuai Wang, Yang Liu, Yu-Xiao Guo, Chun-Yu Sun, and Xin Tong.
\newblock O-cnn: Octree-based convolutional neural networks for 3d shape
  analysis.
\newblock {\em ACM Transactions On Graphics (TOG)}, 36(4):1--11, 2017.

\bibitem{Wang2022Dual}
Peng-Shuai Wang, Yang Liu, and Xin Tong.
\newblock Dual octree graph networks for learning adaptive volumetric shape
  representations.
\newblock {\em ACM Trans. Graph.}, 41(4), jul 2022.

\bibitem{wang2018adaptive}
Peng-Shuai Wang, Chun-Yu Sun, Yang Liu, and Xin Tong.
\newblock Adaptive o-cnn: A patch-based deep representation of 3d shapes.
\newblock {\em ACM Transactions on Graphics (TOG)}, 37(6):1--11, 2018.

\bibitem{wang20193dn}
Weiyue Wang, Duygu Ceylan, Radomir Mech, and Ulrich Neumann.
\newblock 3dn: 3d deformation network.
\newblock In {\em Proceedings of the IEEE/CVF Conference on Computer Vision and
  Pattern Recognition}, pages 1038--1046, 2019.

\bibitem{wanghf}
Yiqun Wang, Ivan Skorokhodov, and Peter Wonka.
\newblock Hf-neus: Improved surface reconstruction using high-frequency
  details.
\newblock In {\em Advances in Neural Information Processing Systems}, 2022.

\bibitem{wen2019pixel2mesh++}
Chao Wen, Yinda Zhang, Zhuwen Li, and Yanwei Fu.
\newblock Pixel2mesh++: Multi-view 3d mesh generation via deformation.
\newblock In {\em Proceedings of the IEEE/CVF international conference on
  computer vision}, pages 1042--1051, 2019.

\bibitem{williams2022neural}
Francis Williams, Zan Gojcic, Sameh Khamis, Denis Zorin, Joan Bruna, Sanja
  Fidler, and Or~Litany.
\newblock Neural fields as learnable kernels for 3d reconstruction.
\newblock In {\em Proceedings of the IEEE/CVF Conference on Computer Vision and
  Pattern Recognition}, pages 18500--18510, 2022.

\bibitem{williams2019deep}
Francis Williams, Teseo Schneider, Claudio Silva, Denis Zorin, Joan Bruna, and
  Daniele Panozzo.
\newblock Deep geometric prior for surface reconstruction.
\newblock In {\em Proceedings of the IEEE/CVF Conference on Computer Vision and
  Pattern Recognition}, pages 10130--10139, 2019.

\bibitem{williams2021neural}
Francis Williams, Matthew Trager, Joan Bruna, and Denis Zorin.
\newblock Neural splines: Fitting 3d surfaces with infinitely-wide neural
  networks.
\newblock In {\em Proceedings of the IEEE/CVF Conference on Computer Vision and
  Pattern Recognition}, pages 9949--9958, 2021.

\bibitem{worchel2022multi}
Markus Worchel, Rodrigo Diaz, Weiwen Hu, Oliver Schreer, Ingo Feldmann, and
  Peter Eisert.
\newblock Multi-view mesh reconstruction with neural deferred shading.
\newblock In {\em Proceedings of the IEEE/CVF Conference on Computer Vision and
  Pattern Recognition}, pages 6187--6197, 2022.

\bibitem{wu2018learning}
Jiajun Wu, Chengkai Zhang, Xiuming Zhang, Zhoutong Zhang, William~T Freeman,
  and Joshua~B Tenenbaum.
\newblock Learning shape priors for single-view 3d completion and
  reconstruction.
\newblock In {\em Proceedings of the European Conference on Computer Vision
  (ECCV)}, pages 646--662, 2018.

\bibitem{wu2021deepcad}
Rundi Wu, Chang Xiao, and Changxi Zheng.
\newblock Deepcad: A deep generative network for computer-aided design models.
\newblock In {\em Proceedings of the IEEE/CVF International Conference on
  Computer Vision}, pages 6772--6782, 2021.

\bibitem{wu2020pq}
Rundi Wu, Yixin Zhuang, Kai Xu, Hao Zhang, and Baoquan Chen.
\newblock Pq-net: A generative part seq2seq network for 3d shapes.
\newblock In {\em Proceedings of the IEEE/CVF Conference on Computer Vision and
  Pattern Recognition}, pages 829--838, 2020.

\bibitem{wu2020unsupervised}
Shangzhe Wu, Christian Rupprecht, and Andrea Vedaldi.
\newblock Unsupervised learning of probably symmetric deformable 3d objects
  from images in the wild.
\newblock In {\em Proceedings of the IEEE/CVF Conference on Computer Vision and
  Pattern Recognition}, pages 1--10, 2020.

\bibitem{xie2019pix2vox}
Haozhe Xie, Hongxun Yao, Xiaoshuai Sun, Shangchen Zhou, and Shengping Zhang.
\newblock Pix2vox: Context-aware 3d reconstruction from single and multi-view
  images.
\newblock In {\em Proceedings of the IEEE/CVF international conference on
  computer vision}, pages 2690--2698, 2019.

\bibitem{xie2022neural}
Yiheng Xie, Towaki Takikawa, Shunsuke Saito, Or~Litany, Shiqin Yan, Numair
  Khan, Federico Tombari, James Tompkin, Vincent Sitzmann, and Srinath Sridhar.
\newblock Neural fields in visual computing and beyond.
\newblock In {\em Computer Graphics Forum}, volume~41, pages 641--676. Wiley
  Online Library, 2022.

\bibitem{xu2019disn}
Qiangeng Xu, Weiyue Wang, Duygu Ceylan, Radomir Mech, and Ulrich Neumann.
\newblock Disn: Deep implicit surface network for high-quality single-view 3d
  reconstruction.
\newblock {\em Advances in Neural Information Processing Systems}, 32, 2019.

\bibitem{xu2020ladybird}
Yifan Xu, Tianqi Fan, Yi~Yuan, and Gurprit Singh.
\newblock Ladybird: Quasi-monte carlo sampling for deep implicit field based 3d
  reconstruction with symmetry.
\newblock In {\em European Conference on Computer Vision}, pages 248--263.
  Springer, 2020.

\bibitem{yan2021hpnet}
Siming Yan, Zhenpei Yang, Chongyang Ma, Haibin Huang, Etienne Vouga, and Qixing
  Huang.
\newblock Hpnet: Deep primitive segmentation using hybrid representations.
\newblock In {\em Proceedings of the IEEE/CVF International Conference on
  Computer Vision}, pages 2753--2762, 2021.

\bibitem{yao2020front2back}
Yuan Yao, Nico Schertler, Enrique Rosales, Helge Rhodin, Leonid Sigal, and Alla
  Sheffer.
\newblock Front2back: Single view 3d shape reconstruction via front to back
  prediction.
\newblock In {\em Proceedings of the IEEE/CVF Conference on Computer Vision and
  Pattern Recognition}, pages 531--540, 2020.

\bibitem{yariv2021volume}
Lior Yariv, Jiatao Gu, Yoni Kasten, and Yaron Lipman.
\newblock Volume rendering of neural implicit surfaces.
\newblock {\em Advances in Neural Information Processing Systems},
  34:4805--4815, 2021.

\bibitem{yariv2020multiview}
Lior Yariv, Yoni Kasten, Dror Moran, Meirav Galun, Matan Atzmon, Basri Ronen,
  and Yaron Lipman.
\newblock Multiview neural surface reconstruction by disentangling geometry and
  appearance.
\newblock {\em Advances in Neural Information Processing Systems},
  33:2492--2502, 2020.

\bibitem{ye2022gifs}
Jianglong Ye, Yuntao Chen, Naiyan Wang, and Xiaolong Wang.
\newblock Gifs: Neural implicit function for general shape representation.
\newblock In {\em Proceedings of the IEEE/CVF Conference on Computer Vision and
  Pattern Recognition}, pages 12829--12839, 2022.

\bibitem{yin2018p2p}
Kangxue Yin, Hui Huang, Daniel Cohen-Or, and Hao Zhang.
\newblock P2p-net: Bidirectional point displacement net for shape transform.
\newblock {\em ACM Transactions on Graphics (TOG)}, 37(4):1--13, 2018.

\bibitem{yu2022capri}
Fenggen Yu, Zhiqin Chen, Manyi Li, Aditya Sanghi, Hooman Shayani, Ali
  Mahdavi-Amiri, and Hao Zhang.
\newblock Capri-net: Learning compact cad shapes with adaptive primitive
  assembly.
\newblock In {\em Proceedings of the IEEE/CVF Conference on Computer Vision and
  Pattern Recognition}, pages 11768--11778, 2022.

\bibitem{yumonosdf}
Zehao Yu, Songyou Peng, Michael Niemeyer, Torsten Sattler, and Andreas Geiger.
\newblock Monosdf: Exploring monocular geometric cues for neural implicit
  surface reconstruction.
\newblock {\em Advances in Neural Information Processing Systems (NeurIPS)},
  2022.

\bibitem{zhang2021ners}
Jason Zhang, Gengshan Yang, Shubham Tulsiani, and Deva Ramanan.
\newblock Ners: Neural reflectance surfaces for sparse-view 3d reconstruction
  in the wild.
\newblock {\em Advances in Neural Information Processing Systems},
  34:29835--29847, 2021.

\bibitem{zhang2022critical}
Jingyang Zhang, Yao Yao, Shiwei Li, Tian Fang, David McKinnon, Yanghai Tsin,
  and Long Quan.
\newblock Critical regularizations for neural surface reconstruction in the
  wild.
\newblock In {\em Proceedings of the IEEE/CVF Conference on Computer Vision and
  Pattern Recognition}, pages 6270--6279, 2022.

\bibitem{zhang2021learning}
Jingyang Zhang, Yao Yao, and Long Quan.
\newblock Learning signed distance field for multi-view surface reconstruction.
\newblock In {\em Proceedings of the IEEE/CVF International Conference on
  Computer Vision}, pages 6525--6534, 2021.

\bibitem{zhang2018learning}
Xiuming Zhang, Zhoutong Zhang, Chengkai Zhang, Josh Tenenbaum, Bill Freeman,
  and Jiajun Wu.
\newblock Learning to reconstruct shapes from unseen classes.
\newblock {\em Advances in neural information processing systems}, 31, 2018.

\bibitem{zhang2020image}
Yuxuan Zhang, Wenzheng Chen, Huan Ling, Jun Gao, Yinan Zhang, Antonio Torralba,
  and Sanja Fidler.
\newblock Image gans meet differentiable rendering for inverse graphics and
  interpretable 3d neural rendering.
\newblock In {\em International Conference on Learning Representations}, 2020.

\bibitem{zhao2021sign}
Wenbin Zhao, Jiabao Lei, Yuxin Wen, Jianguo Zhang, and Kui Jia.
\newblock Sign-agnostic implicit learning of surface self-similarities for
  shape modeling and reconstruction from raw point clouds.
\newblock In {\em Proceedings of the IEEE/CVF Conference on Computer Vision and
  Pattern Recognition}, pages 10256--10265, 2021.

\end{thebibliography}

\end{document}